\documentclass[final]{cvpr}

\usepackage{times}
\usepackage{epsfig}
\usepackage{graphicx}
\usepackage{amsmath}
\usepackage{array}
\usepackage{amssymb}
\usepackage{appendix}
\usepackage{dsfont}
\usepackage{booktabs}
\usepackage{multirow}
\usepackage{lipsum} %
\usepackage{enumitem}
\usepackage{arydshln}
\usepackage[ruled,vlined,linesnumbered]{algorithm2e}
\usepackage[section]{placeins}

\newcommand*\rot[1]{\raisebox{2\normalbaselineskip}[0pt][0pt]{\rotatebox[origin=c]{90}{#1}}}
\newcolumntype{x}[1]{%
>{\centering\hspace{0pt}}p{#1}}%

\newenvironment{myitemize}
{ \begin{itemize}
    \setlength{\itemsep}{0pt}
    \setlength{\parskip}{0pt}
    \setlength{\parsep}{0pt}
    \setlength{\topsep}{0pt}}
{ \end{itemize}                  } 

\usepackage[pagebackref=true,breaklinks=true,colorlinks,bookmarks=false]{hyperref}

\def\cvprPaperID{7214} %
\def\confYear{CVPR 2021}

\begin{document}

\title{Hijack-GAN: Unintended-Use of Pretrained, Black-Box GANs}

\author{Hui-Po Wang$^1$ \quad Ning Yu$^{2,3}$ \quad Mario Fritz$^1$\\ \\
$^1$CISPA Helmholtz Center for Information Security, Germany \\
$^2$Max Planck Institute for Informatics, Saarland Informatics Campus, Germany \\
$^3$University of Maryland, College Park 
}

\maketitle

\begin{abstract}

   While Generative Adversarial Networks (GANs) show increasing performance and the level of realism is becoming indistinguishable from natural images, this also comes with high demands on data and computation. We show that state-of-the-art GAN models -- such as they are being publicly released by researchers and industry -- can be used for a range of applications beyond unconditional image generation. We achieve this by an iterative scheme that also allows gaining control over the image generation process despite the highly non-linear latent spaces of the latest GAN models. We demonstrate that this opens up the possibility to re-use state-of-the-art, difficult to train, pre-trained GANs with a high level of control even if only black-box access is granted. Our work also raises concerns and awareness that the use cases of a published GAN model may well reach beyond the creators' intention, which needs to be taken into account before a full public release. Code is available at \url{https://github.com/hui-po-wang/hijackgan}.

\end{abstract}

\section{Introduction}
Generative Adversarial Nets (GANs)~\cite{goodfellow2014generative} have achieved remarkable success in many applications, such as image synthesis~\cite{karras2019style, karras2018progressive, karras2020analyzing} and image translation~\cite{chang2019all, choi2020stargan, choi2018stargan, zhou2020generate}. By learning a mapping between noise and images, the models are skilled to produce photo-realistic images from random noise. However, as the architectures become sophisticated~\cite{karras2019style, karras2018progressive, brock2018large}, training modern GANs often requires massive data and computation resources. For example, it takes one high-quality face dataset, 8 V100 GPUs, and one week to train a single StyleGAN~\cite{abdal2019image2stylegan} model. In light of this trend, it is crucial to reuse existing pre-trained GANs, such as those being released by researchers or industry, for building green AI systems, in which the critical factor is to achieve other tasks beyond the original intention of GANs.

To reuse GANs for other tasks, prior works have shown that semantic manipulation can be realized by vector arithmetic~\cite{radford2015unsupervised} or moving along constant attribute vectors~\cite{goetschalckx2019ganalyze, upchurch2017deep, shen2020interpreting, denton2019detecting} in latent spaces. For instance, InterfaceGAN~\cite{shen2020interpreting} demonstrates that facial attribute manipulation can be achieved by moving noise close to or away from the linear decision boundary of the desired attribute. Although these methods reveal the potential that pre-trained GANs could go beyond unconditional generation, they highly rely on the assumption of linear manifolds, thereby ignoring the nature of highly non-linear latent spaces (e.g., $\mathcal{Z}$-space of StyleGAN~\cite{shen2020interpreting, zhu2020domain}). This strong assumption could be harmful, especially for rare attributes, and lead to ineffective manipulation.

Inspired by this observation, we propose a novel framework, which gains high-level control over unconditional image generation by iteratively traversing the non-linear latent spaces. Specifically, we first train a proxy model that bypasses the gradients from one pretrained GAN and other fixed task models, and then dynamically decides the moving direction in each step, thus producing smoother and more effective attribute transition. Next, we propose an orthogonal constraint to solely edit the attribute of interest while retaining others in images. Despite only black-box access, we show that our method can achieve various unintended tasks, including manipulation over facial attributes, head poses, and landmarks.

As a result of our experiments, we find that our framework with pre-trained GANs not only facilitates other vision tasks but raises concerns regarding further usage. Even without access to model parameters, the models can still be applied to unintended tasks potentially for malicious purposes. The owners of GANs should be aware and cautious about the potential risks before releasing their models. Overall, our contributions are summarized below.
\begin{myitemize}
\item We propose a framework which leverages off-the-shelf GANs to approach unintended vision tasks in a black-box setting.
\item  We propose a constraint that helps our framework solely edit the attribute of interest while retaining others in images.
\item Extensive results show that our method can produce smoother and more effective manipulation while preserving non-target attributes better, as compared to prior work. We also shed light on the potential risks of unintended usage by gaining control over facial attributes, head poses, and landmarks.
\end{myitemize}
\section{Related Work}
\label{sec:related_work}
\noindent \textbf{Generative Adversarial Nets (GANs).} Since GAN~\cite{goodfellow2014generative} was proposed, it has advanced many applications such as image synthesis~\cite{abdal2019image2stylegan, karras2018progressive, karras2020analyzing} and image translation~\cite{chang2019all, choi2020stargan, choi2018stargan, zhou2020generate}. The rationale behind it is to map noise drawn from a simple distribution (e.g., Gaussian) to a real data distribution (e.g., images) by non-linear networks. Many works~\cite{karras2019style, brock2018large, karras2018progressive, karras2020analyzing, zhang2019self, arjovsky2017wasserstein} have been proposed to improve image quality while the demands on resources also significantly increase. In this work, we consider two state-of-the-art GANs, PGGAN~\cite{karras2018progressive} and StyleGAN~\cite{karras2019style}, as our generator backbones. The former one takes a vector of noise as input and progressively upsamples features to generate images. The later one adopts a similar strategy but first embed the noise vector by neural networks and treat them as style representations. As observed by Shen~\etal~\cite{shen2020interpreting}, the $z$-space of StyleGAN is expected to be more entangled than PGGAN. In general, GAN models for new tasks are often trained from scratch, making them inefficient and difficult to scale up. With our framework, we show that pre-trained GANs can well approach a range of applications beyond their original purposes.

\vspace{0.1cm}\noindent \textbf{Study on Latent Spaces of GANs.} A considerable amount of works has attempted to understand the latent space of GANs. In particular, \cite{radford2015unsupervised, upchurch2017deep} show that semantic manipulation is achievable by vector arithmetic on latent code; \cite{karras2019style} show that style transfer can be achieved by mixing two latent represents. %
Some studies have attempted to identity semantically meaningful directions by self-supervised learning~\cite{voynov2020unsupervised} or PCA on latent spaces~\cite{harkonen2020ganspace}. They are able to perform transformations by moving noise toward the direction. However, since they approach it in an unsupervised fashion, they may be limited to generalize to new tasks beyond simple transformations.

Recent works further indicate that various semantic meanings may be implicitly encoded in the latent space of pre-trained GANs even though the concepts are not specified in the training set, ranging from memorability~\cite{goetschalckx2019ganalyze}, transformation~\cite{jahanian2019steerability}, to facial attributes~\cite{shen2020interpreting, denton2019detecting, tewari2020stylerig}. For instance, InterfaceGAN~\cite{shen2020interpreting} uses the normal vector of the SVM decision boundary to edit facial attributes by moving noise along the vector. Notably, most of the works mentioned above assume that the manifolds expand linearly in the latent space, ignoring that the underlying manifold could be extremely non-linear. In contrast to the prior works, we propose a general framework that can be applied to various vision applications by traversing the non-linear manifolds in an iterative scheme.

\vspace{0.1cm}\noindent \textbf{Study on Leveraging Pre-trained GANs.}
Recently, several studies have investigated how to leverage existing GANs to achieve new tasks. One way is to treat pre-trained GANs as strong image priors, which can be achieved by either exploiting GANs as a component of their networks~\cite{pan2020exploiting, bau2020rewriting, bau2019semantic, hussein2020image, abdal2020styleflow} or directly editing the latent code to find feasible solutions~\cite{abdal2019image2stylegan, abdal2020image2stylegan++, collins2020editing}. These methods often require access to model parameters or additional training data. In contrast, this work aims to study the potential usage of black-box pretrained GANs, advising the owners to be careful about the risk before releasing models.

\section{Hijack-GAN}
\begin{figure*}[htbp]
    \centering
    \includegraphics[trim=0cm 0cm 0cm 19cm,clip,width=0.85\linewidth]{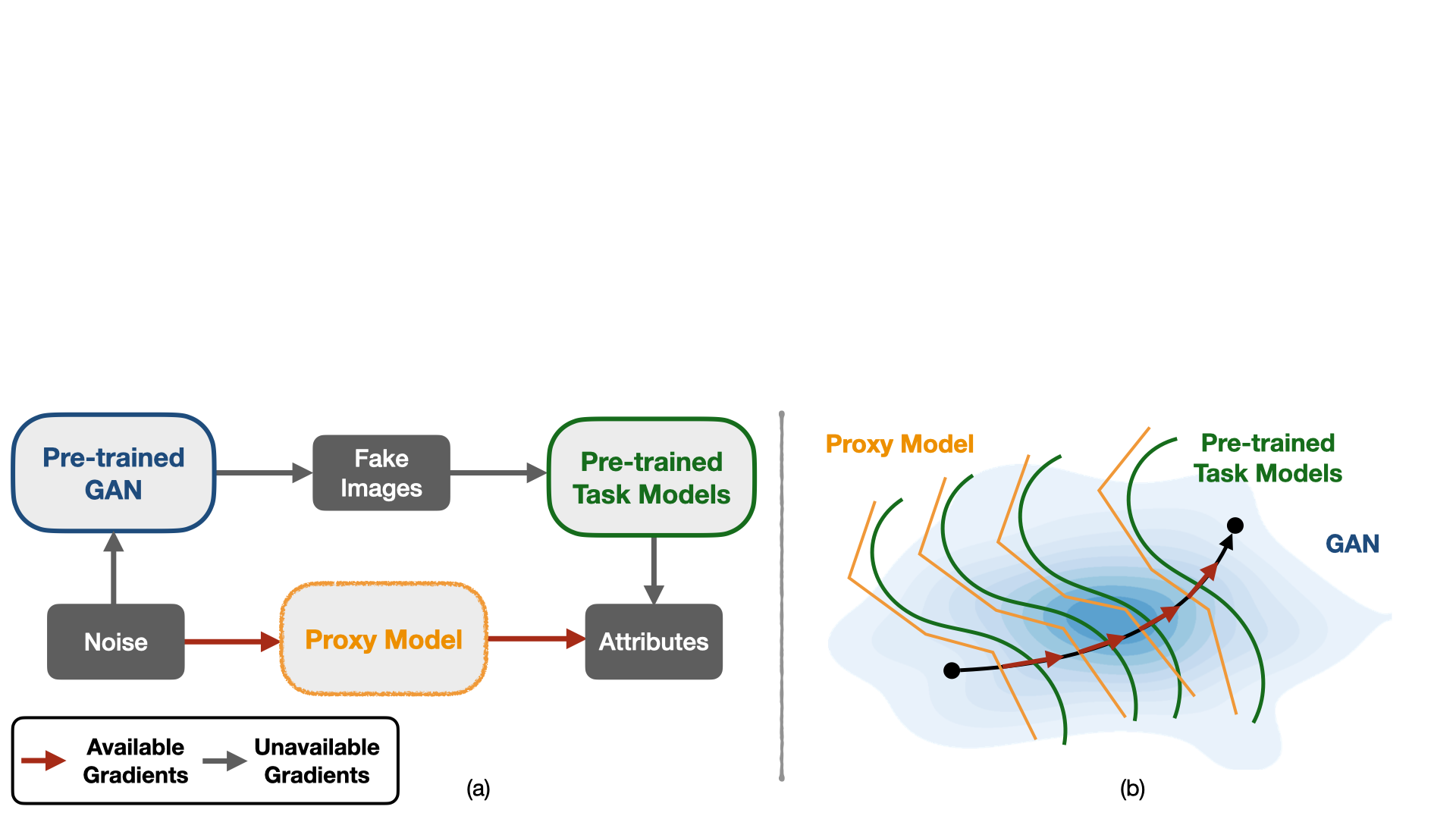}
    \caption{An overview of the proposed framework. The framework takes two steps to reuse GANs: (a) train a proxy model to distill information from pre-trained models, circumventing accessing the gradients of pre-trained models; (b) identify and iteratively traverse a non-linear trajectory under the guidance of gradients.}
    \label{fig:architecture}
\end{figure*}

In this section, we first formalize the problem and then describe how to reuse models by our framework, followed by a proposed orthogonal constraint to improve disentanglement ability. Lastly, we discuss the practical considerations. The overall architecture is shown in Figure~\ref{fig:architecture}.

\subsection{Problem Statement}
\label{ssec:problem_statement}
We aim to study the possibility that pre-trained GANs can be used for unintended applications beyond unconditional generation.
Specifically, we consider a victim generator $G:z \rightarrow I$ that maps noise $z\sim p(z)$ to realistic images $I$, and one or multiple victim task models $\mathcal{M}:I \rightarrow A$ that map images $I$ to the attributes space $A$. We aim to find a trajectory $\mathcal{T}$ in the latent space $p(z)$ such that, as traversing along the path, the desired task can be gradually achieved, which can be further expressed as follows:

\begin{equation}
    \label{eq:objective}
    \mathcal{L} \circ \mathcal{M} \circ G(z^{(i+1)}) \leq \mathcal{L} \circ \mathcal{M} \circ G(z^{(i)}), \forall z^{(i)} \in \mathcal{T},
\end{equation}
where $\mathcal{L}$ is the loss function of unintended tasks. Note that the parameters and training data for the task are inaccessible, making it impossible to be directly solved by optimization methods since they require the gradients from the models.

\subsection{Non-linear Traversal}
\label{ssec:method_traversal}
To identify a meaningful trajectory in the black-box setting, we have to address the following two issues: (1) the gradients from the GAN and the task models are impassable. (2) we need to identify a trajectory that precisely describes the highly non-linear manifolds. %

Since the gradients are impassable, we first train a proxy model $\mathcal{P}$ to distill the information from the models. We assume that only $z$-space is available and synthesize data pairs $(z, \mathcal{M} \circ G(z))$ to train the proxy model such that it can map input noise to the attribute space; therefore, we circumvent directly accessing the gradients while remaining informed about the relation between attributes and noise. Next, we compute the Jacobian matrix of the proxy model $\mathcal{P}$ with respect to the input noise $z \in \mathbb{R}^n$.

\begin{equation}
    \label{eq:jacobian}
    \renewcommand\arraystretch{2}
    \mathbf{J} = \begin{bmatrix}
        \frac{\partial \mathcal{P}_1}{\partial z_1} & \cdots & \frac{\partial \mathcal{P}_1}{\partial z_n} \\
        \vdots & \ddots & \vdots \\
        \frac{\partial \mathcal{P}_m}{\partial z_1} & \cdots & \frac{\partial \mathcal{P}_m}{\partial z_n} \\
    \end{bmatrix},
\end{equation}
where $\mathcal{P}_j$ denotes j-th attributes predicted by the proxy. Each row vector of the matrix $\mathbf{J}$, termed $\mathbf{J}_j$ for simplicity, can be interpreted as the direction in which the corresponding attribute changes most quickly; meanwhile, this vector also gives us a hint to identify a meaningful non-linear trajectory.

In light of this, we design an algorithm that iteratively updates the position of noise under the guidance of the vector $\mathbf{J}_j$:
\begin{equation}
    \label{eq:iterative_traversal}
    z^{(i+1)} = z^{(i)} - \lambda \mathbf{J}^{(i)}_j,
\end{equation}
where $\lambda$ is a hyper-parameter deciding moving speed and $\mathbf{J}^{(i)}_j$ is associated with the attribute of interest at step i. Note that we normalize the $\mathbf{J}^{(i)}_j$ to better control the level of changes. By repeatedly computing Eq.~\ref{eq:iterative_traversal} in each step, the target attribute in the generated image $G(z)$ would be gradually modified, granting us high-level control over image generation despite the black-box access.

\subsection{Constraint for Disentanglement}
\label{ssec:method_constraint}
In many cases, attributes may be entangled with each other, meaning that other non-target attributes would be changed if we solely follow the steepest direction in Eq.~\ref{eq:iterative_traversal}. To alleviate the problem, we additionally propose a constraint to encourage disentanglement. Since each row vector $\mathbf{J}_j$ in Eq.~\ref{eq:jacobian} represents one direction that affects certain attributes most, we aim to find a vector which have the maximum inner product with the target direction $\mathbf{J}_j$ while being orthogonal to other non-target directions $\mathbf{J}_{k \neq j}$. We formulate the constraint as a linear program as follows.

\begin{equation}
\label{eq:constraint}
\begin{aligned}
 & \underset{n}{\text{maximize}} && \mathbf{J}_{j}^Tn \\
& \text{subject to} && An = 0,
\end{aligned}
\end{equation}
where $n$ is the direction vector of interest, and each row of $A$ consists of the attribute vector $\mathbf{J}_{k \neq j}$ on which we want to condition. By substituting $n$ for $\mathbf{J}^{(i)}_j$ in Eq.~\ref{eq:iterative_traversal}, we can solely edit the attribute of interest while retaining other non-target attributes in the image. Note that since Eq.~\ref{eq:constraint} is evaluated at each iteration, we are still capable of capturing the non-linear manifolds.

\subsection{Implementation}
\label{ssec:implementation}
We consider two state-of-the-art GANs, PGGAN~\cite{karras2018progressive} and StyleGAN~\cite{karras2019style}, as the victim generators. The input of both models is 512-D noise drawn from a standard normal distribution. Following Shen~\etal~\cite{shen2020interpreting}, we do not normalize the input noise for both models. The proxy models are implemented by a stack of fully-connected layers. Except for the last layer, each layer is followed by a ReLU activation function and a Dropout~\cite{srivastava2014dropout} layer with a rate of 0.2. We empirically find that a proxy model with 3 layers works sufficiently well for PGGAN, while it takes 8 layers to work on StyleGAN. It is also observed that the Dropout function plays a critical role to prevent over-fitting on StyleGAN. These observations imply that the $z$-space of StyleGAN may be highly entangled and non-linear. 

For attribute manipulation, we find it important to train the proxy models with balanced datasets. We synthesize datasets for every attribute, each of which involves 100k positive and 100k negative samples.
Since the annotations are unavailable, we adopt a ResNet-50~\cite{he2016deep} classifier pre-trained on CelebAHQ~\cite{karras2018progressive} to annotate the generated images. We discard data with confidence lower than 0.9 to reduce the ambiguity. For other regression tasks, we generate random samples over the latent space and discard those with lower confidence. It is also observed that the proxy model on StyleGAN benefits from more training data, while the effect is marginal on PGGAN.
\section{Experiments}
In this section, we provide analysis of our non-linear iterative scheme by applying it to attribute manipulation, head pose manipulation, and landmark editing, showing that our method benefits from the non-linearity and demonstrates the possibility of unintended usage.

\subsection{Experiment Setup}
\label{ssec:exp_setup}
\noindent\textbf{Settings.} We take two state-of-the-art GANs, StyleGAN~\cite{karras2019style} and PGGAN~\cite{karras2018progressive}, as our backbone generators. Both models are pre-trained on CelebAHQ~\cite{karras2018progressive}. We adopt different task models according to the applications. We use ResNet-50~\cite{he2016deep} pre-trained on CelebAHQ for attribute manipulation, HopeNet~\cite{ruiz2018fine} pre-trained on 300W-LP~\cite{zhu2016face} for head pose manipulation, and  MTCNN~\cite{zhang2016joint} to crop faces and detect landmarks for landmark editing. 
Note that in all experiments, the model parameters and additional training data are unavailable. 

\vspace{0.1cm}\noindent\textbf{Baselines.} We consider two baselines here.
First, following Denton~\etal~\cite{denton2019detecting}, we employ the same classifier as our framework to compute the gradient with respect to the initial point and take it as a constant attribute vector $v$. The attribute manipulation is realized as below, 
\begin{equation}
    \label{eq:linear}
    z^{(i+1)} = z^{(i)} + \lambda v.
\end{equation}
We refer to this baseline as \textit{Linear}. The key difference is that the proposed framework recomputes gradients in every step. Similarly, the second baseline, InterfaceGAN~\cite{shen2020interpreting}, adopts the same linear strategy but derive the vector $v$ from the normal vector of the decision boundary of a linear SVM, which can be viewed as a common direction that changes the attribute most. InterfaceGAN achieves conditional manipulation by pairwisely computing an orthogonal vector for the normal vectors. Note that, instead of using a common direction for all data, we derive moving directions for specific input.

\subsection{Attribute Manipulation}
\label{ssec:attr_manipulation}
We show that our framework can be applied to attribute manipulation. We consider unconditional and conditional settings, respectively. The former one is achieved by solely applying Eq.~\ref{eq:iterative_traversal}; The latter one adopts Eq.~\ref{eq:constraint} to ensure the preservation of non-target attributes.

\begin{figure*}[htbp]
\setlength\tabcolsep{0.1em}
\centering

\begin{tabular}{@{}ccccc:ccccc@{}}
&\multicolumn{4}{l:}{\hspace{1\normalbaselineskip}\raisebox{2\normalbaselineskip}[0pt][0pt]{PGGAN}\hspace{2.5\normalbaselineskip}\rot{Input} \includegraphics[width=0.115\linewidth]{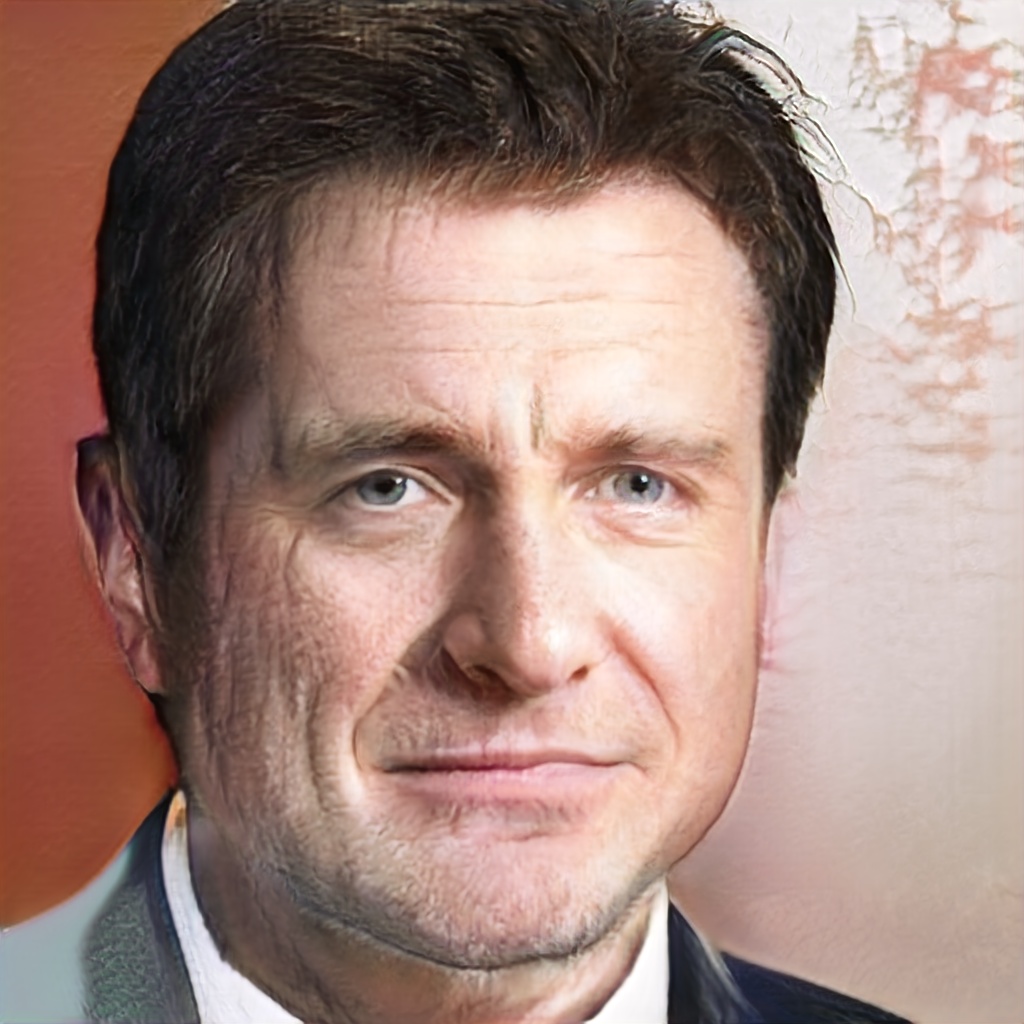}} & \multicolumn{4}{l}{\hspace{1\normalbaselineskip}\raisebox{2\normalbaselineskip}[0pt][0pt]{StyleGAN}\hspace{2\normalbaselineskip}\rot{Input} \includegraphics[width=0.115\linewidth]{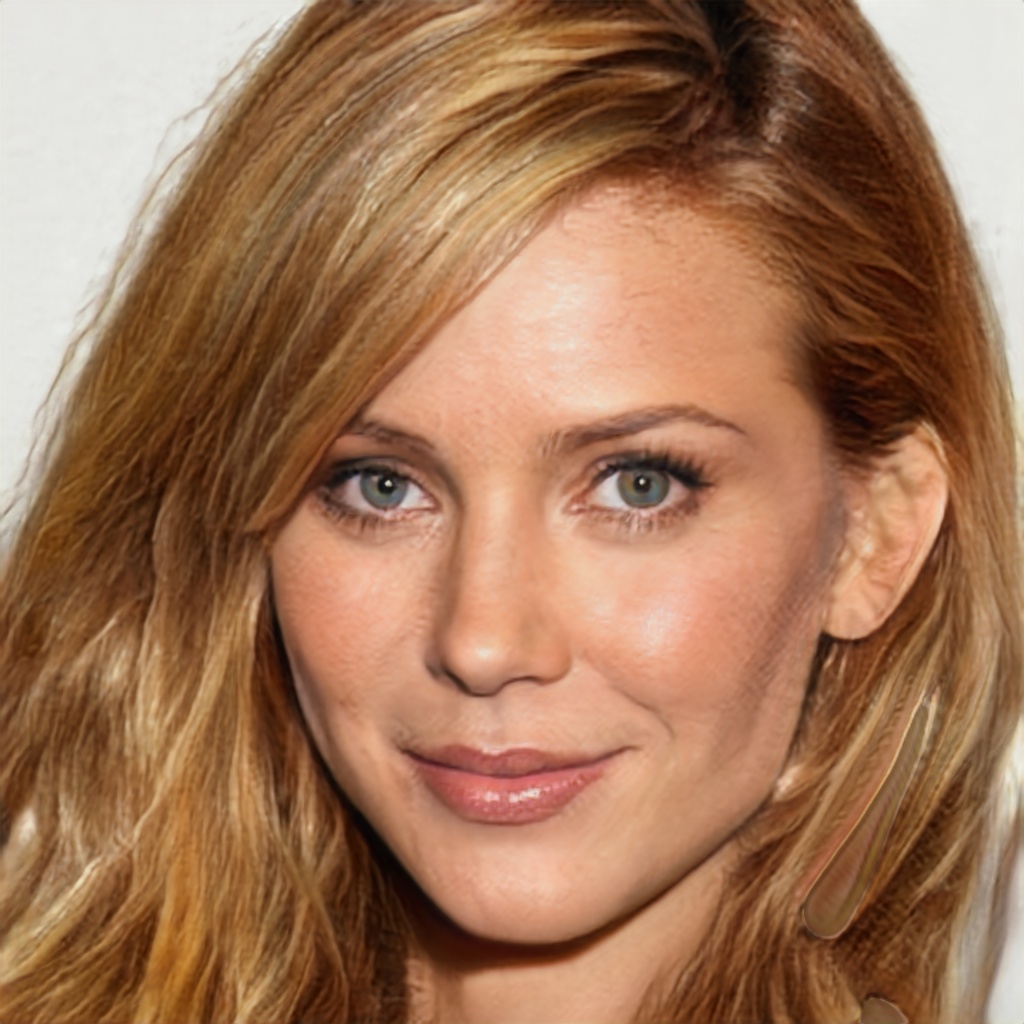}} & \\
 \rot{Linear}& \includegraphics[width=0.115\linewidth]{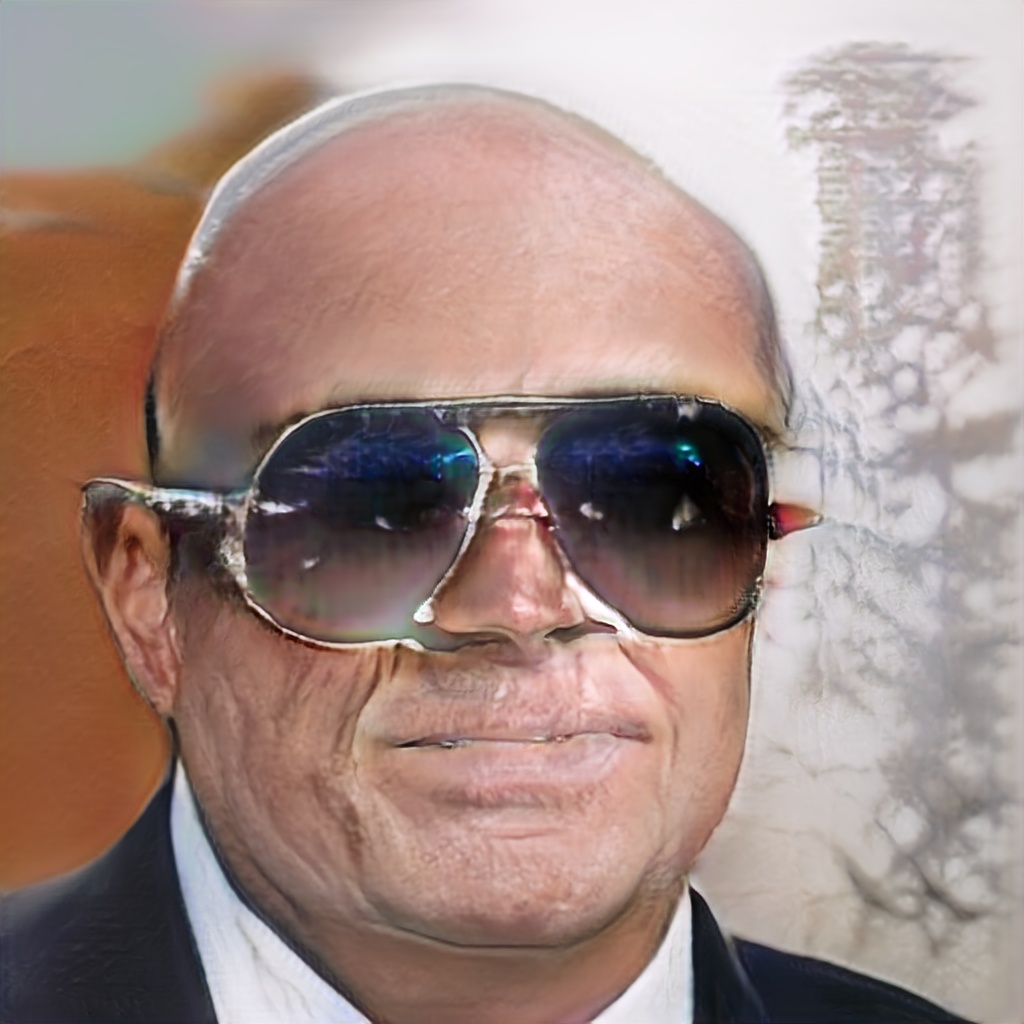} & \includegraphics[width=0.115\linewidth]{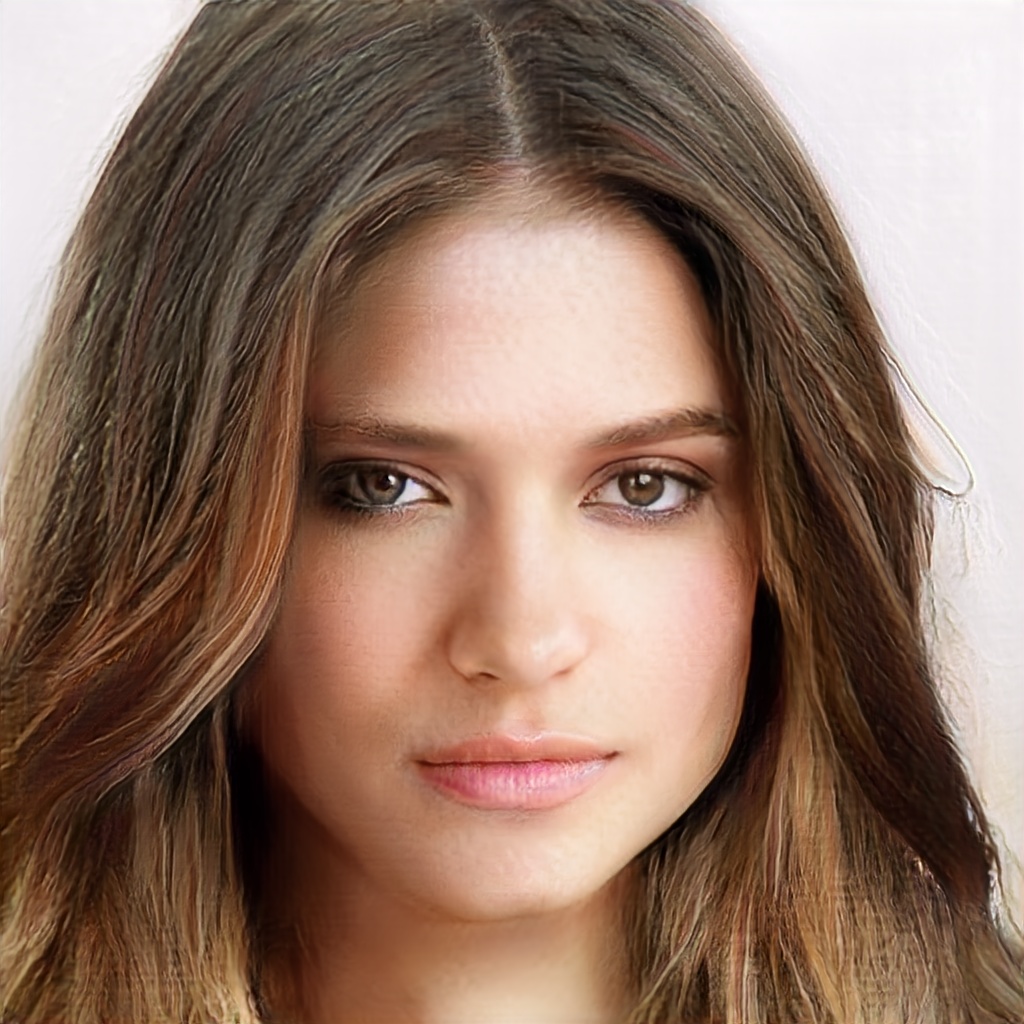} & \includegraphics[width=0.115\linewidth]{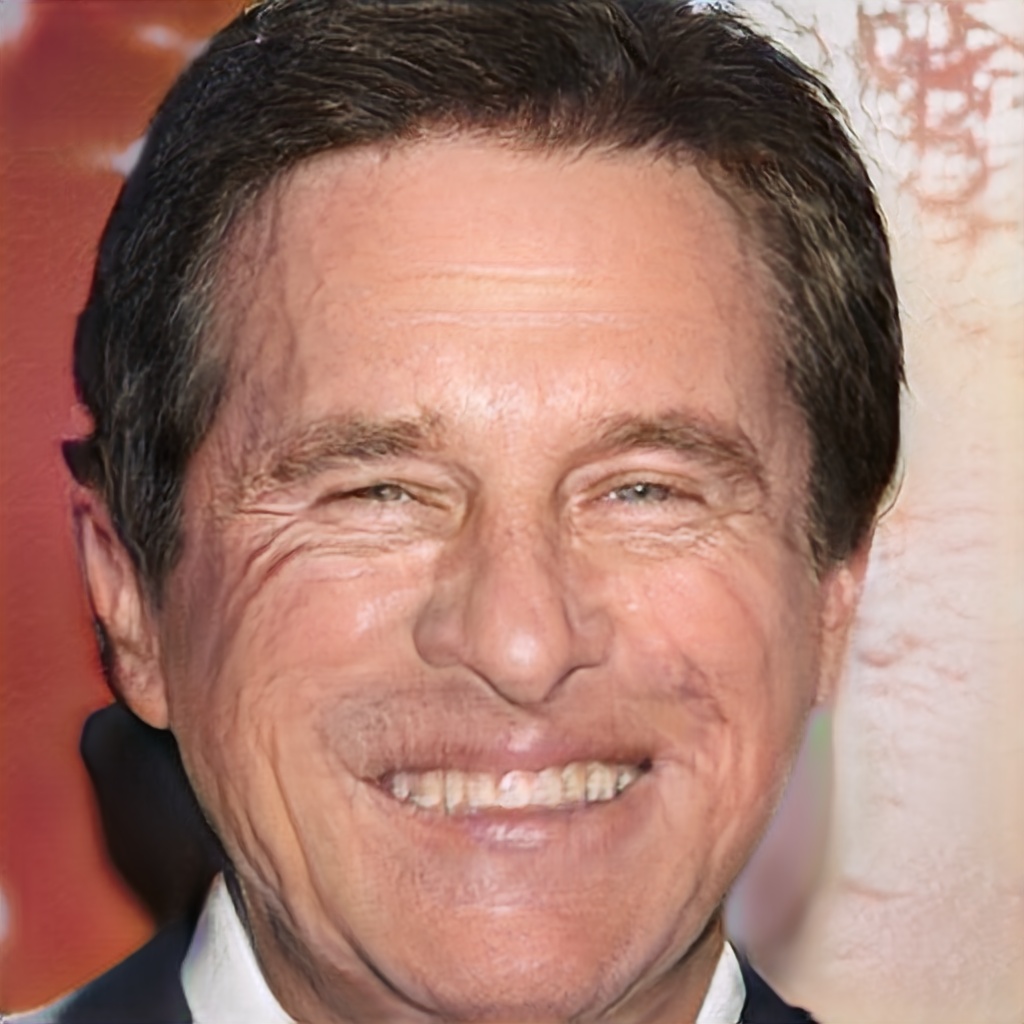} & \includegraphics[width=0.115\linewidth]{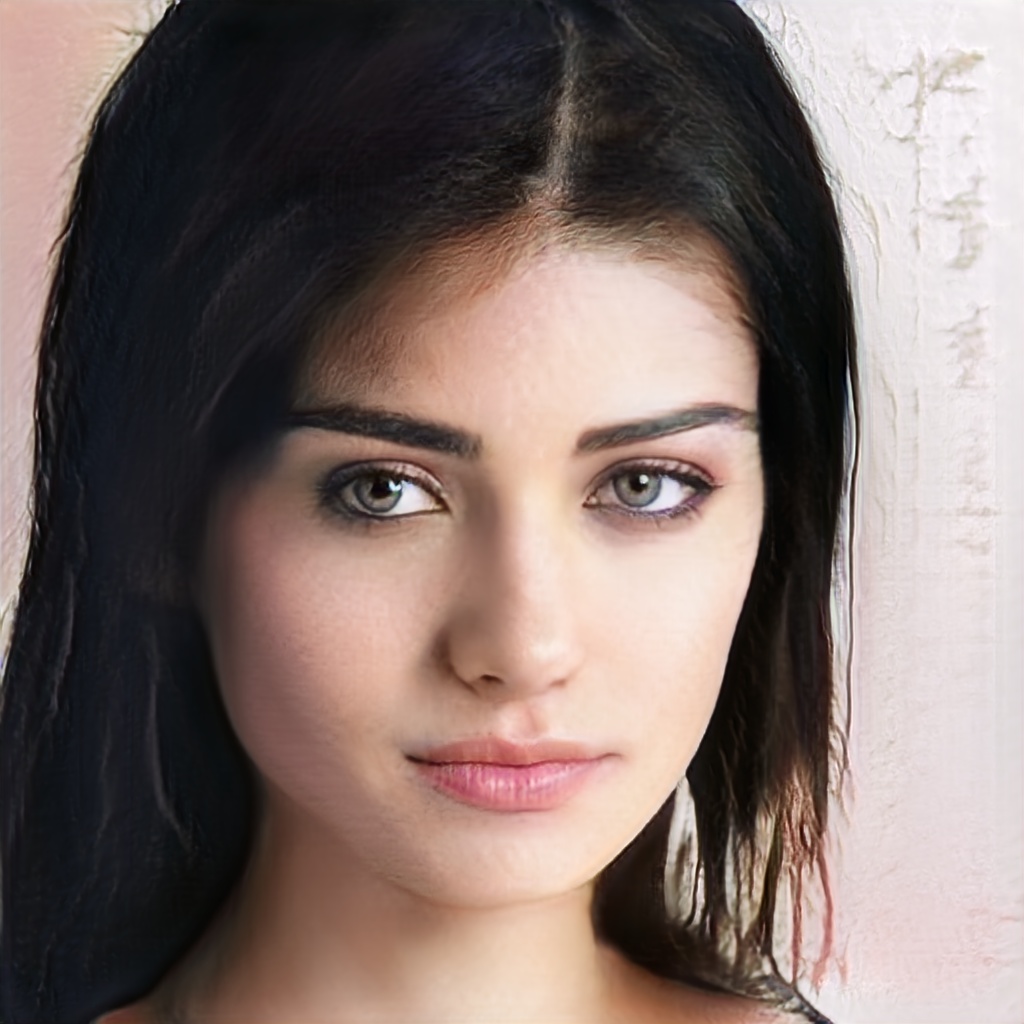} & \includegraphics[width=0.115\linewidth]{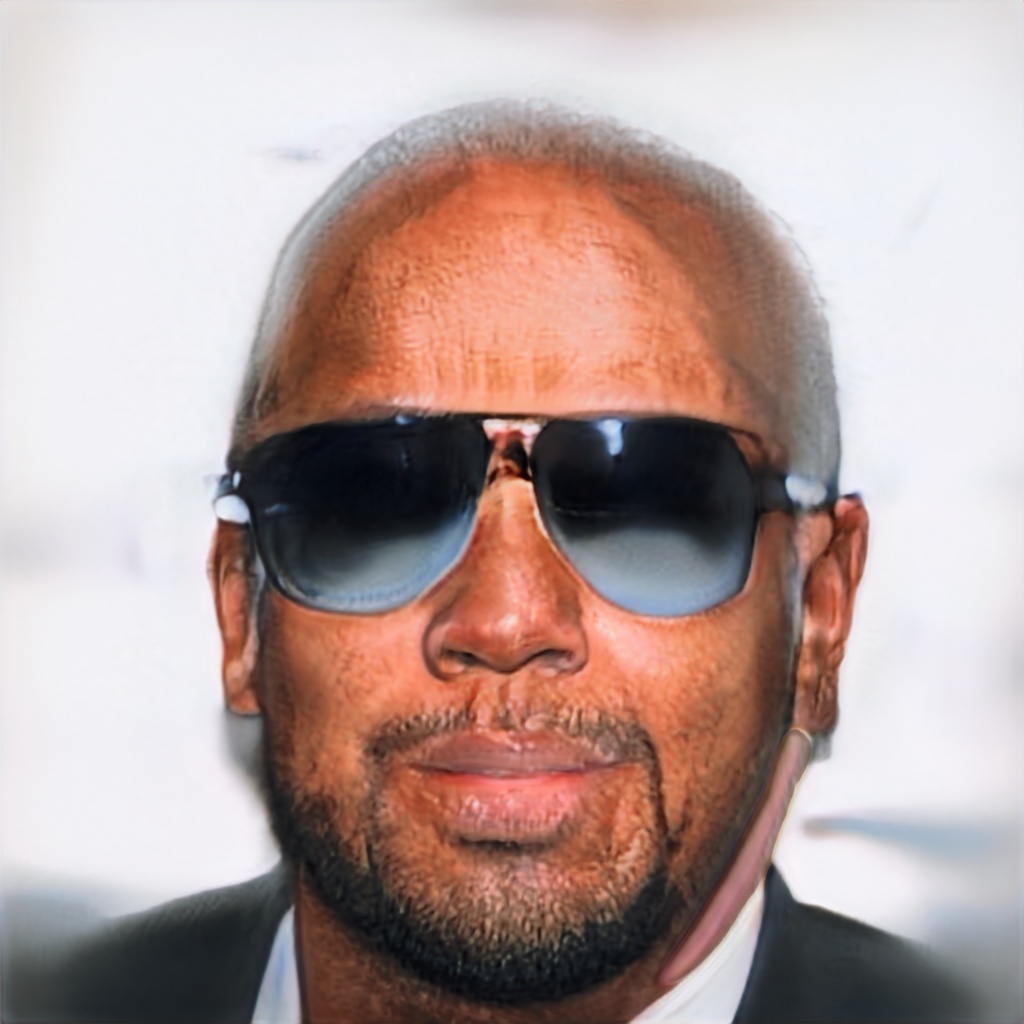} & \includegraphics[width=0.115\linewidth]{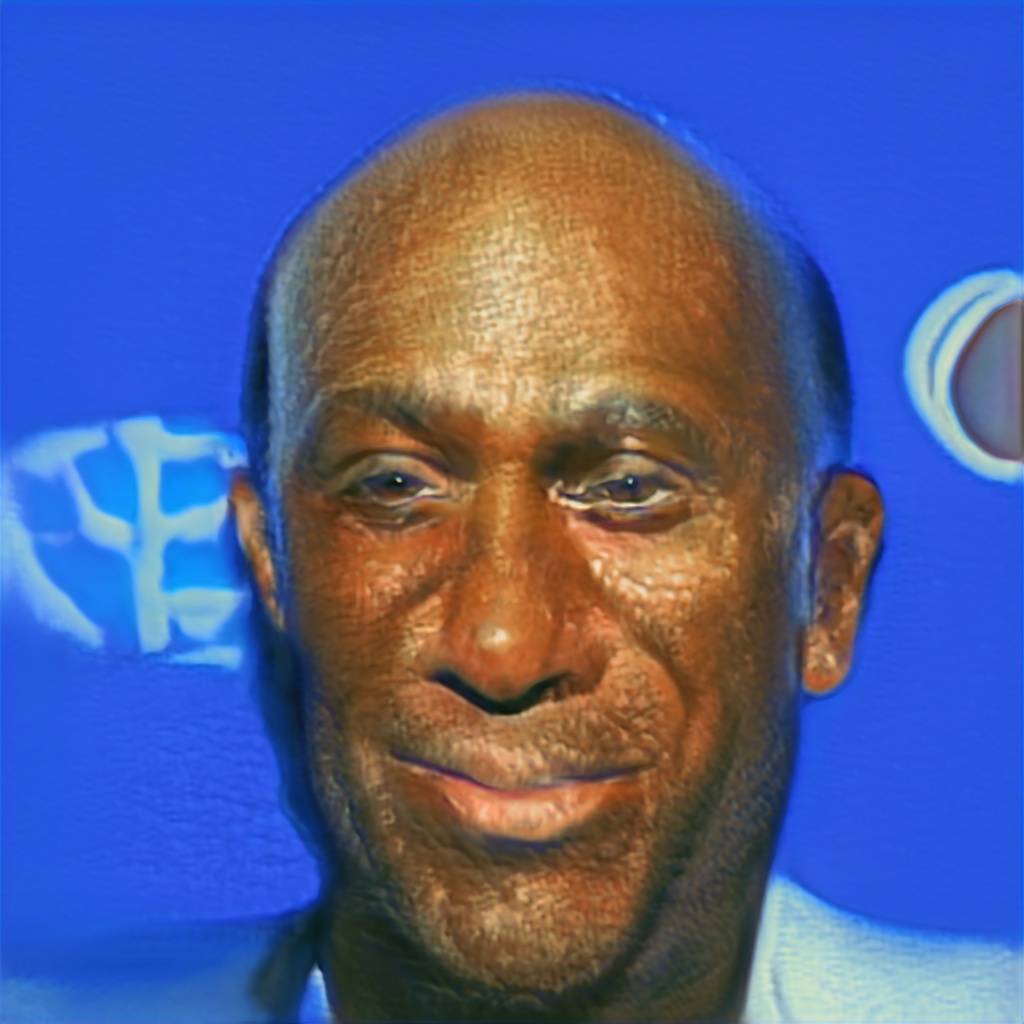} & \includegraphics[width=0.115\linewidth]{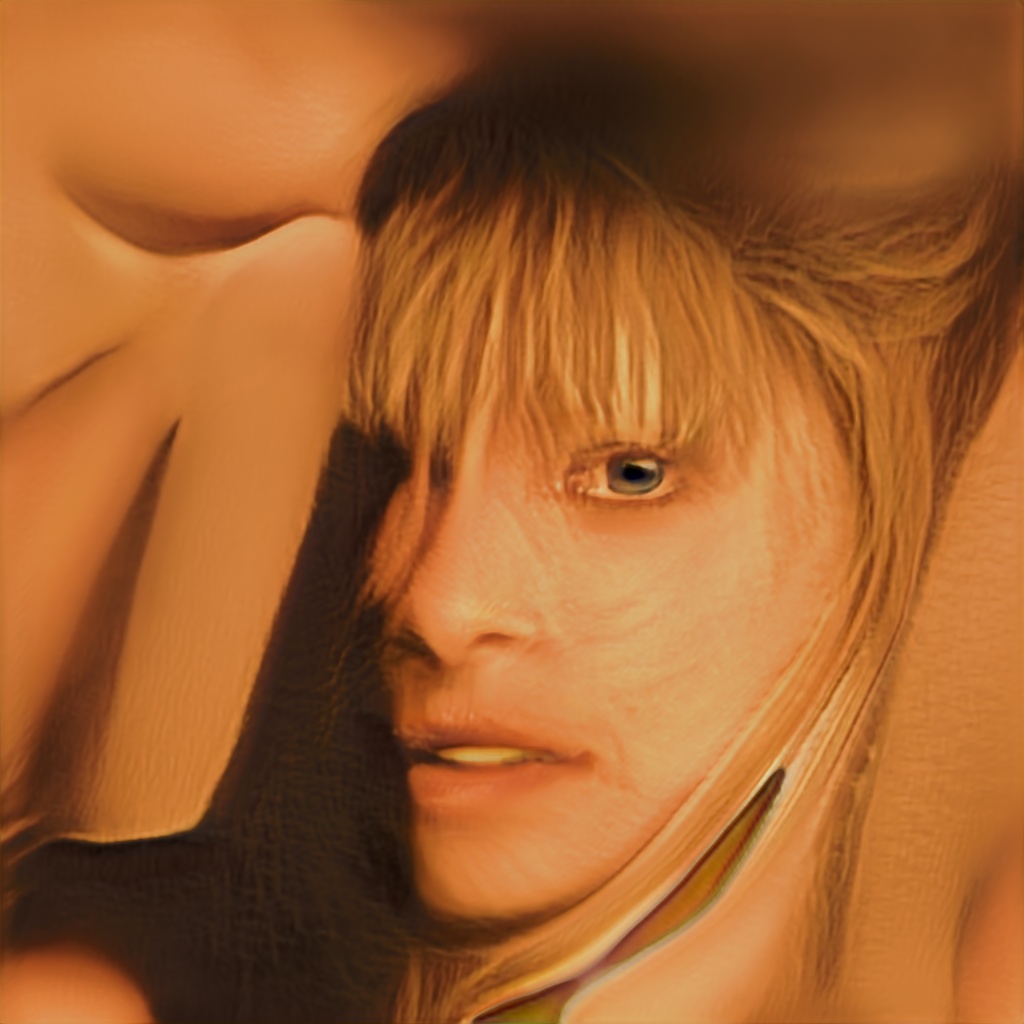} & \includegraphics[width=0.115\linewidth]{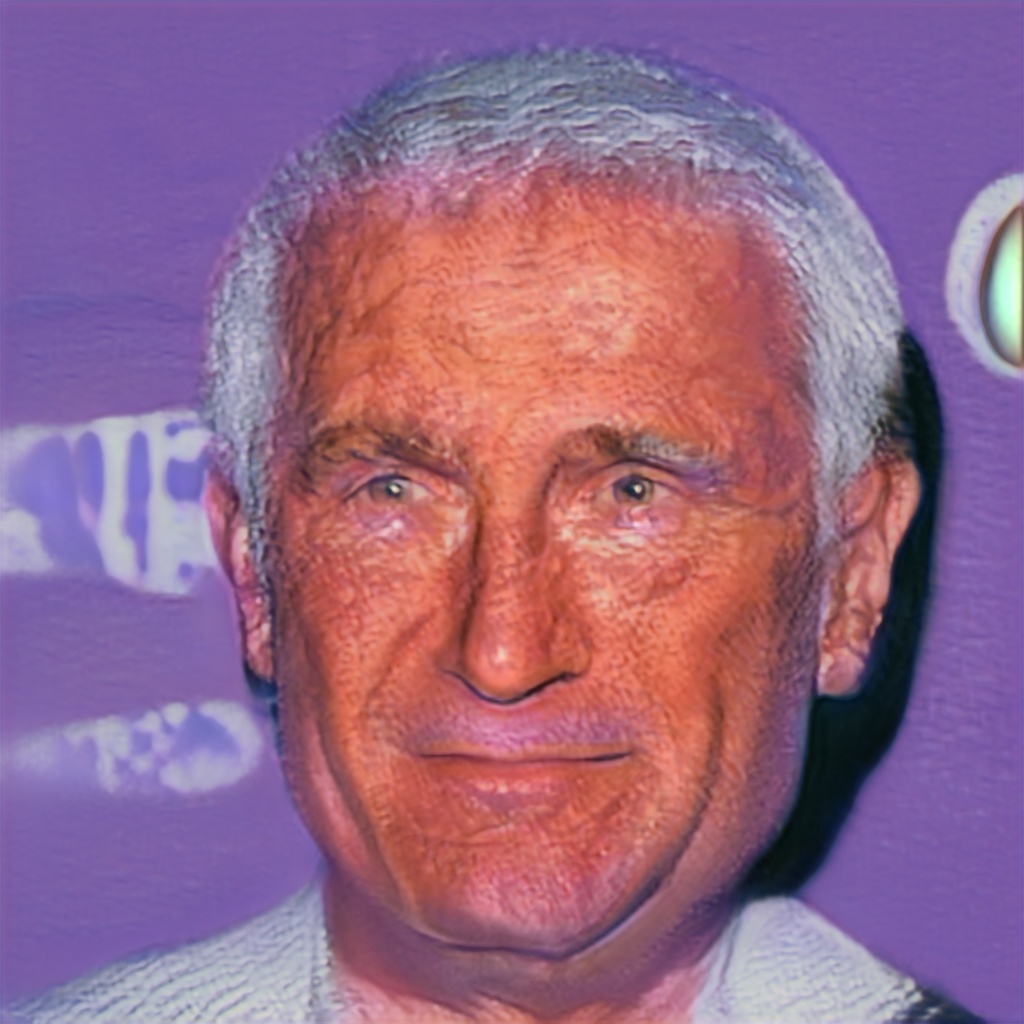} & \rot{Linear} \\
 \rot{InterfaceGAN}& \includegraphics[width=0.115\linewidth]{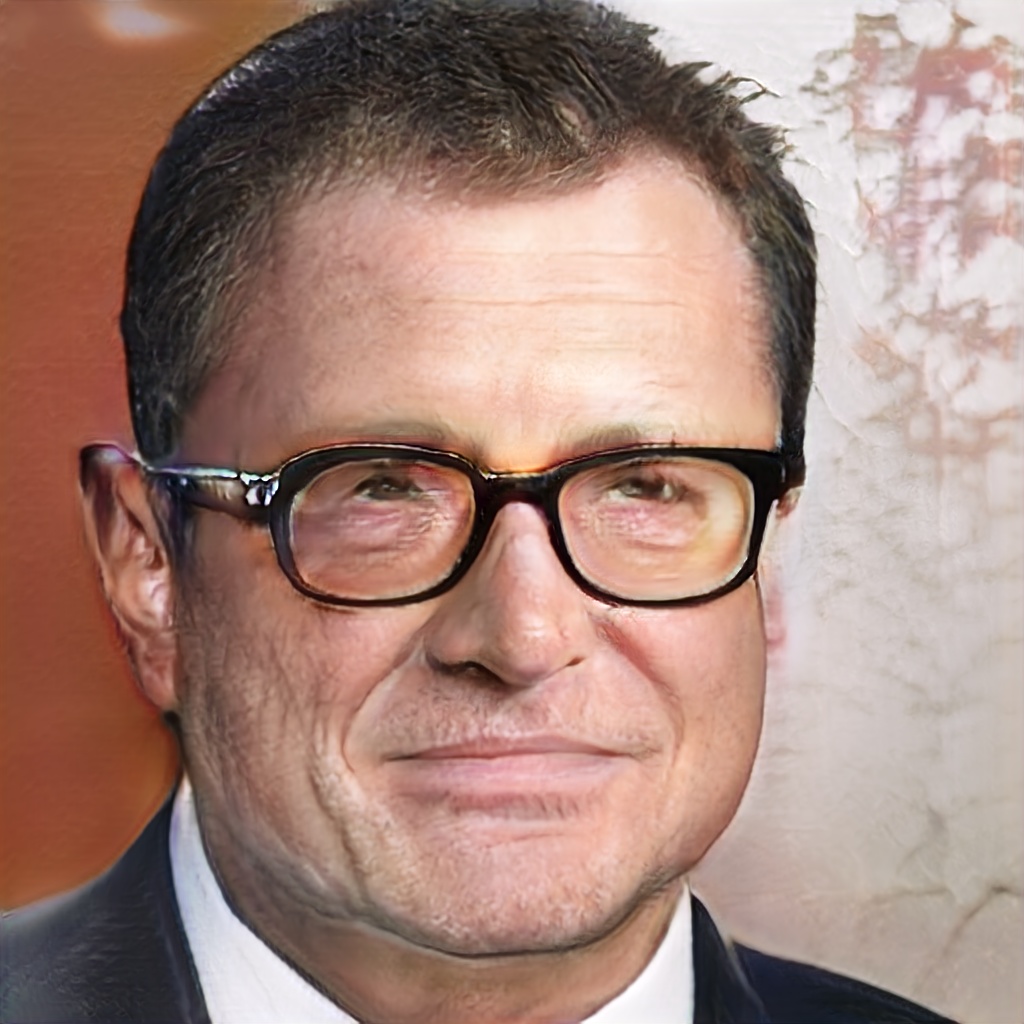} & \includegraphics[width=0.115\linewidth]{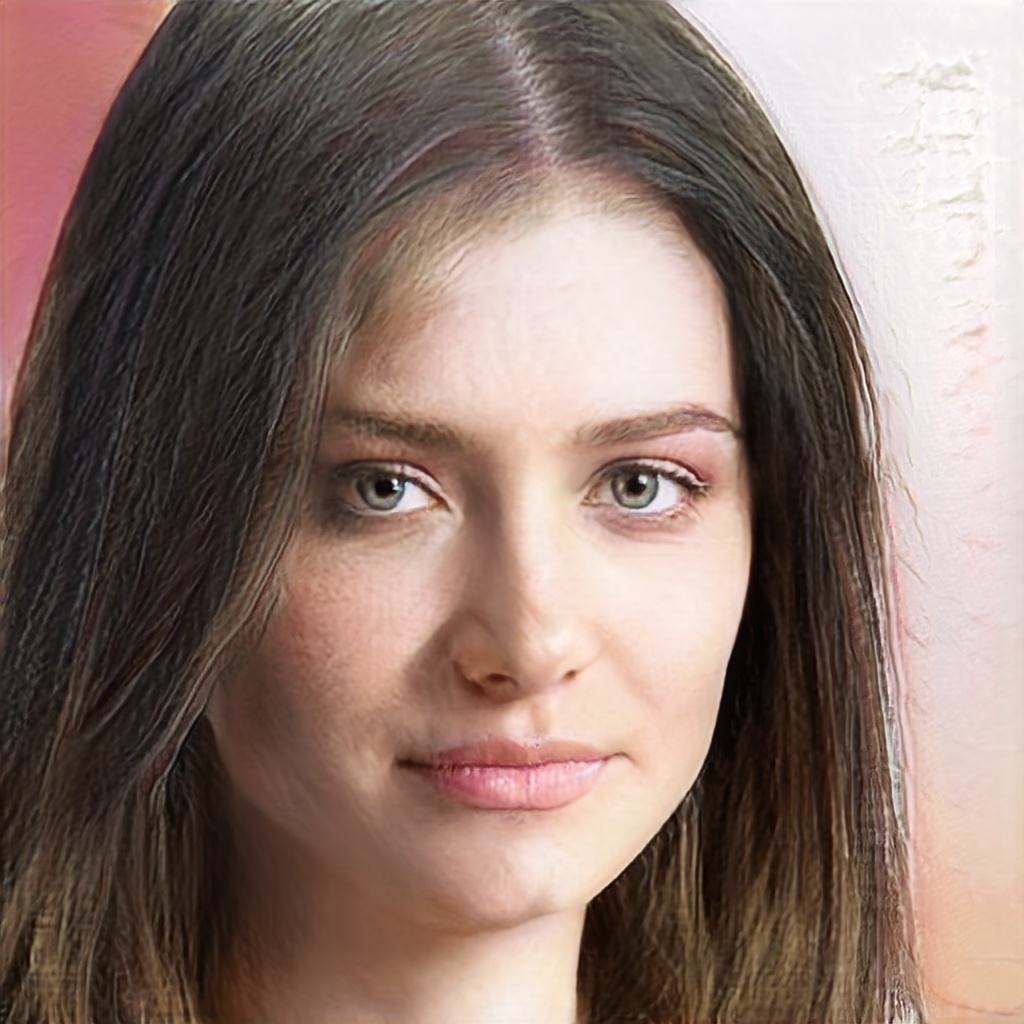} & \includegraphics[width=0.115\linewidth]{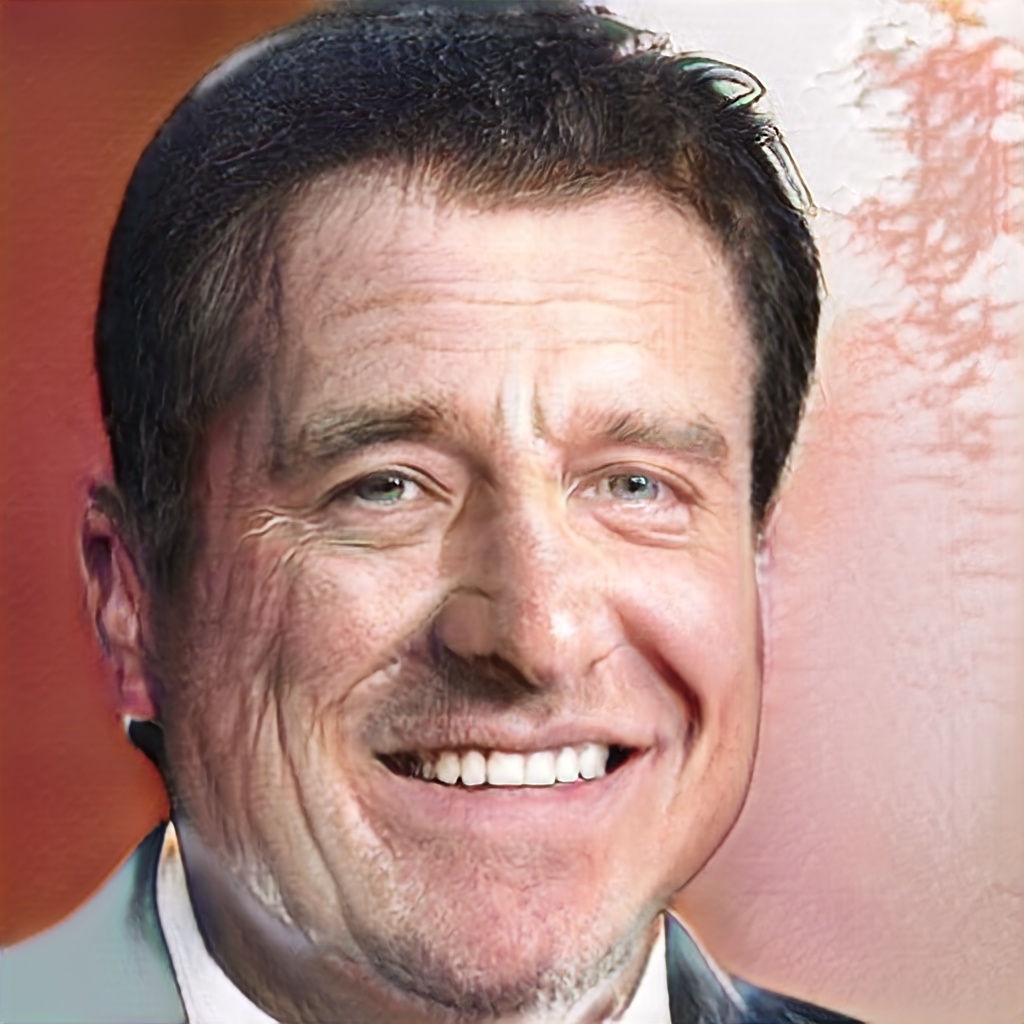} & \includegraphics[width=0.115\linewidth]{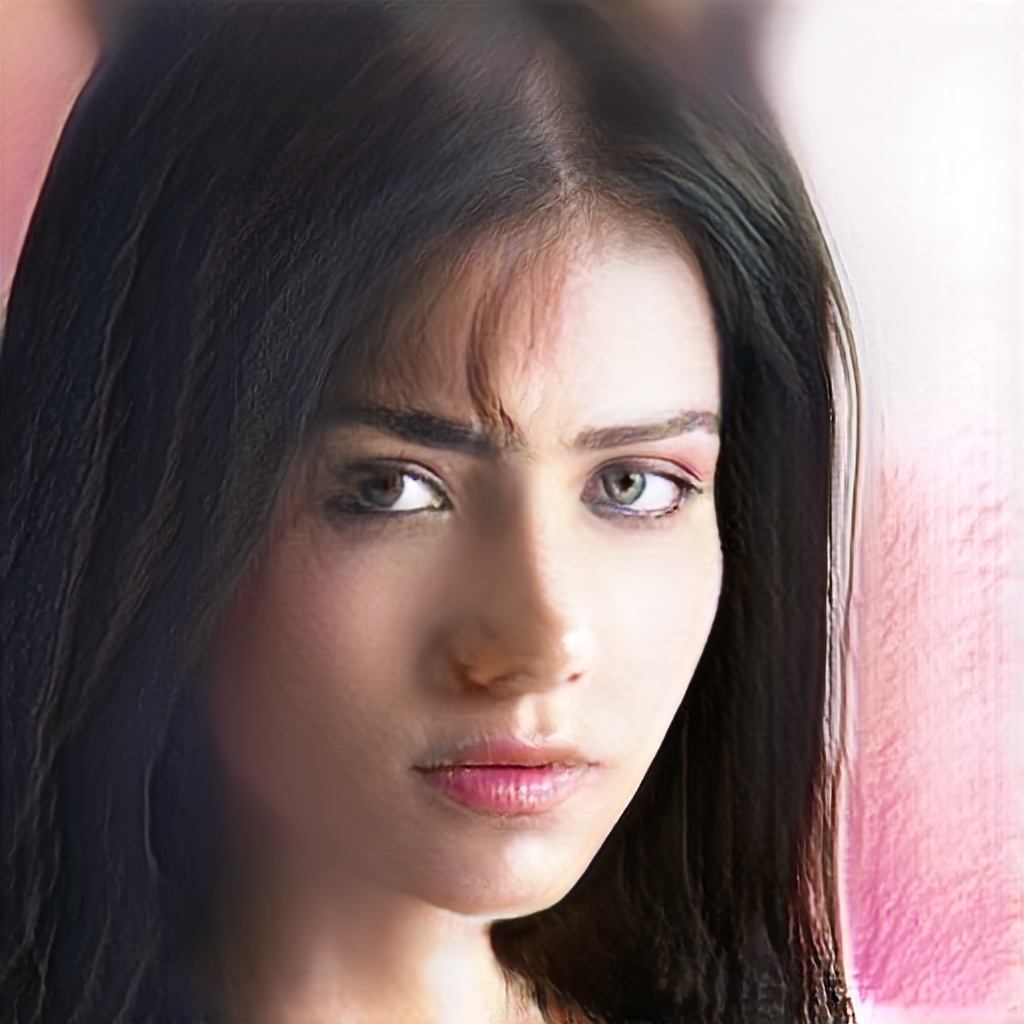} & \includegraphics[width=0.115\linewidth]{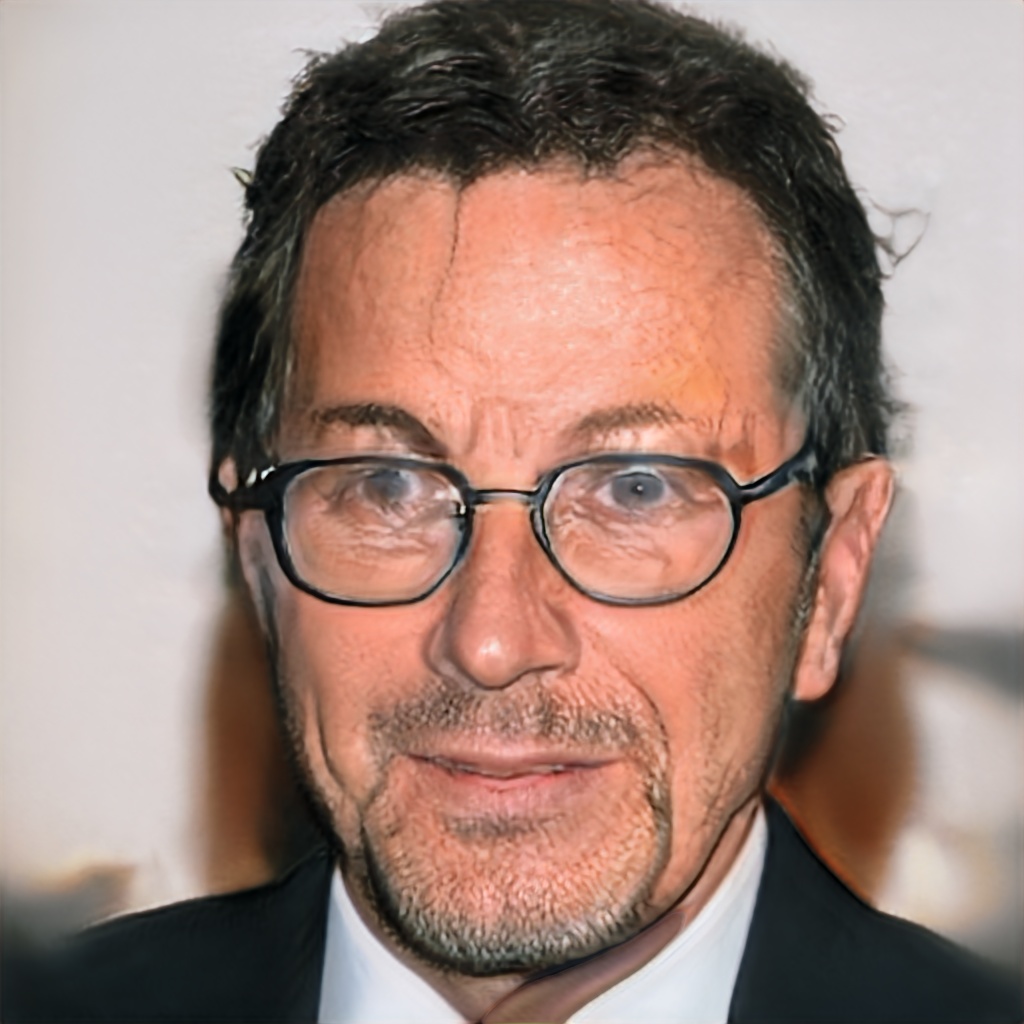} & \includegraphics[width=0.115\linewidth]{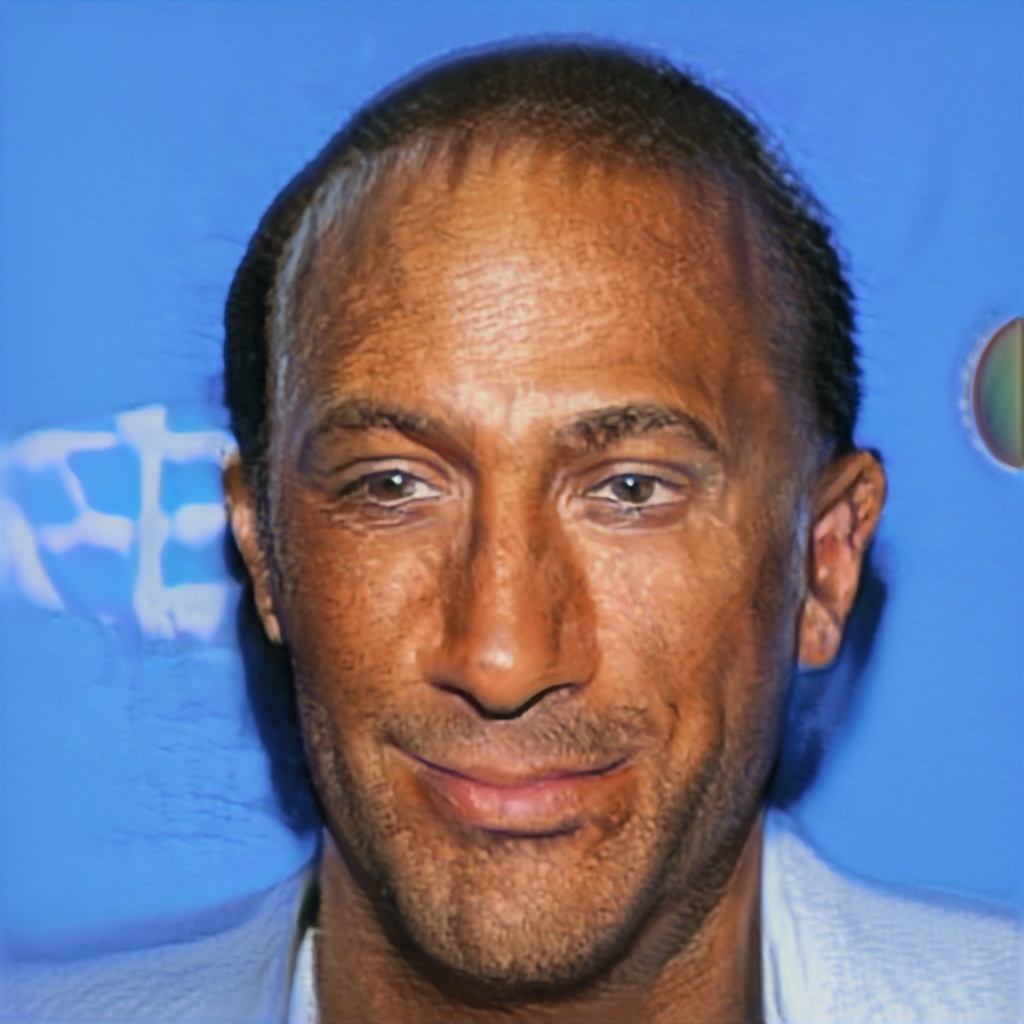} & \includegraphics[width=0.115\linewidth]{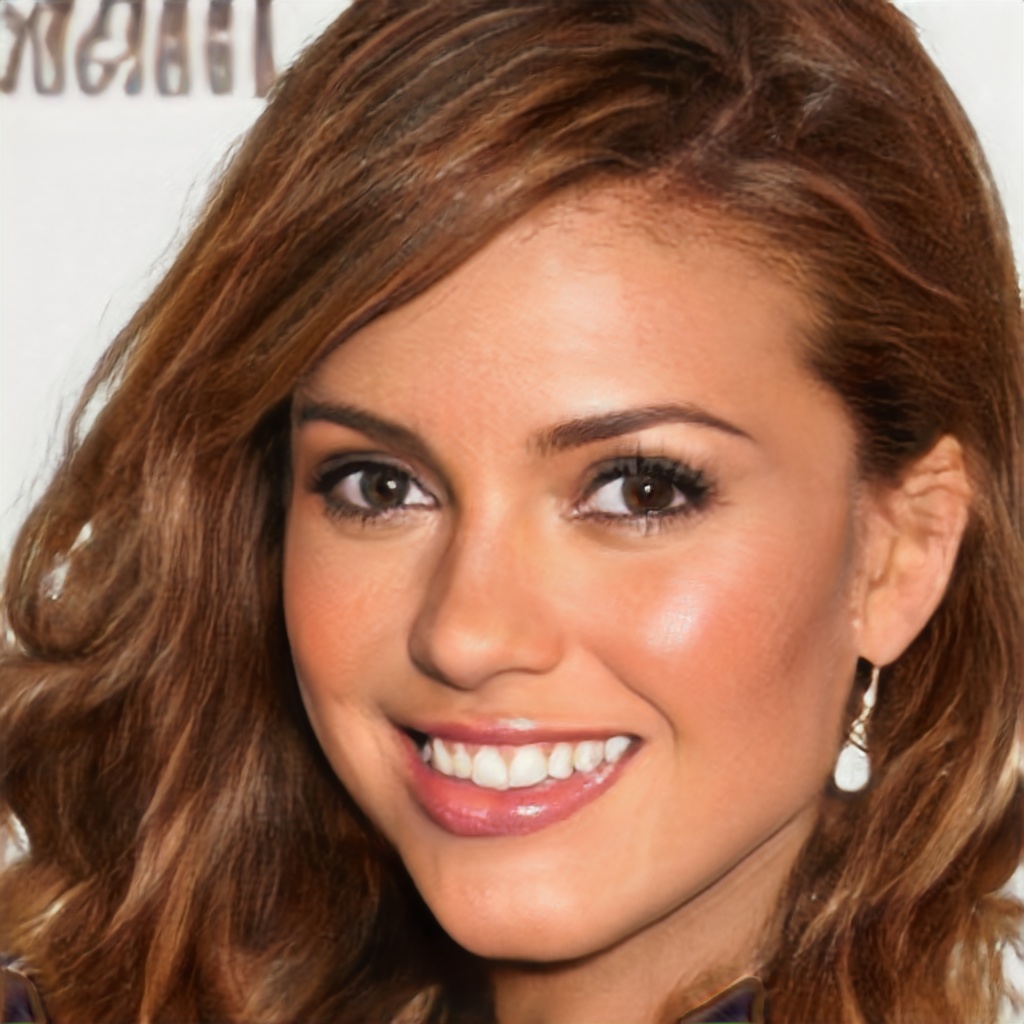} & \includegraphics[width=0.115\linewidth]{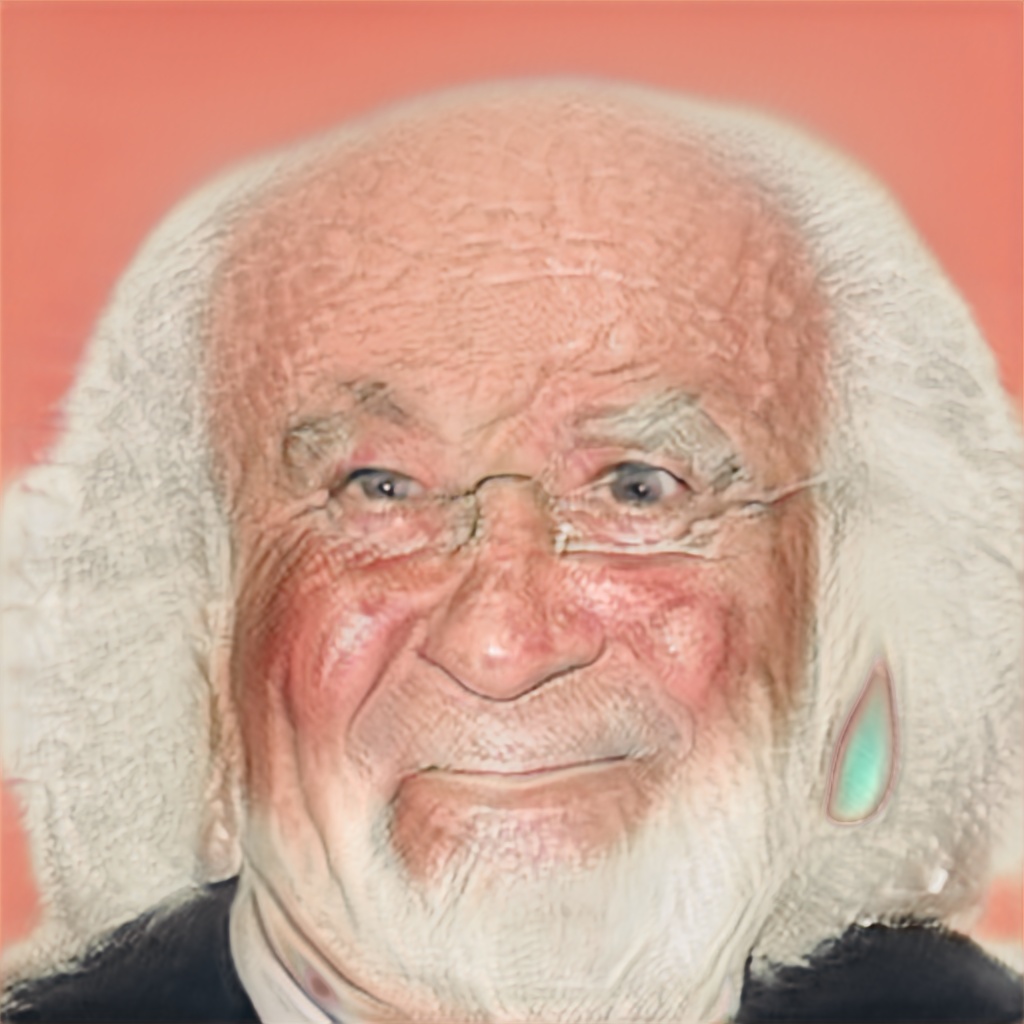} & \rot{InterfaceGAN} \\
 \rot{Ours}& \includegraphics[width=0.115\linewidth]{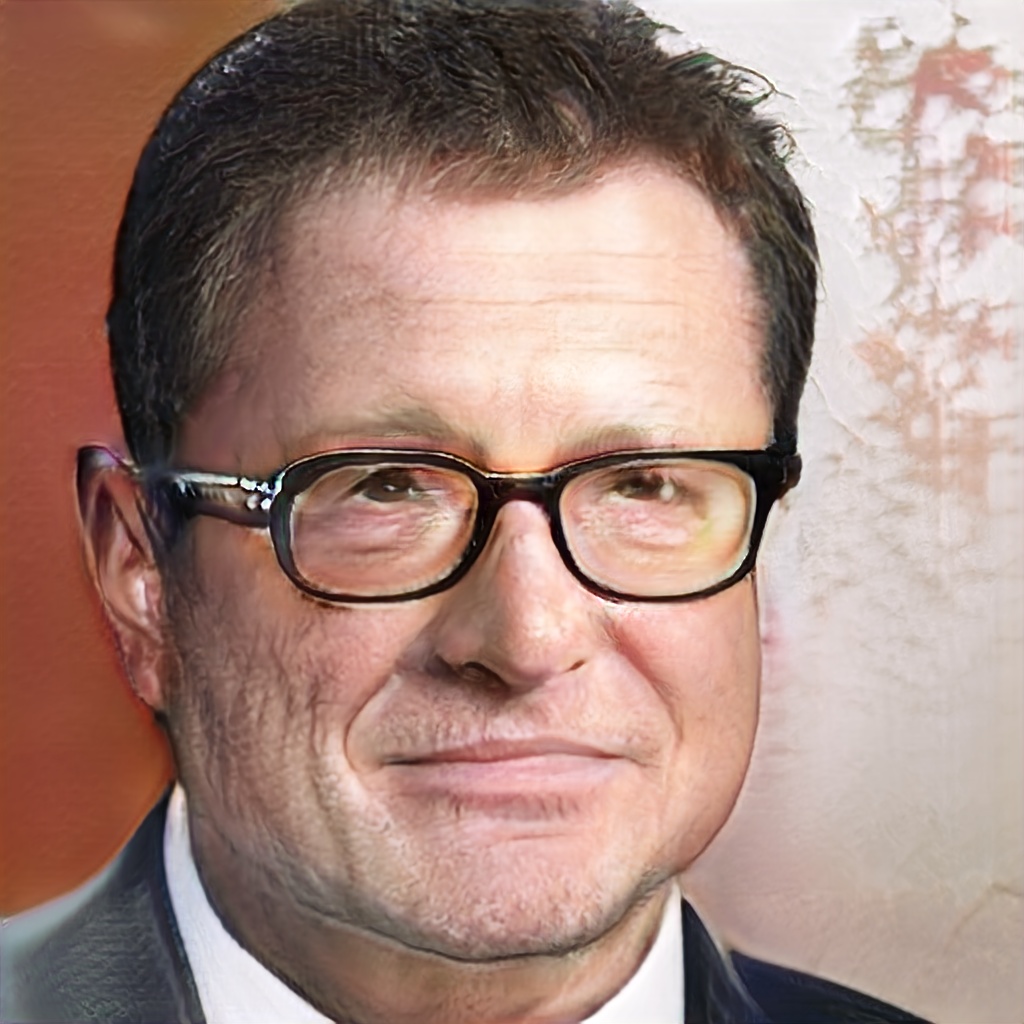}  & \includegraphics[width=0.115\linewidth]{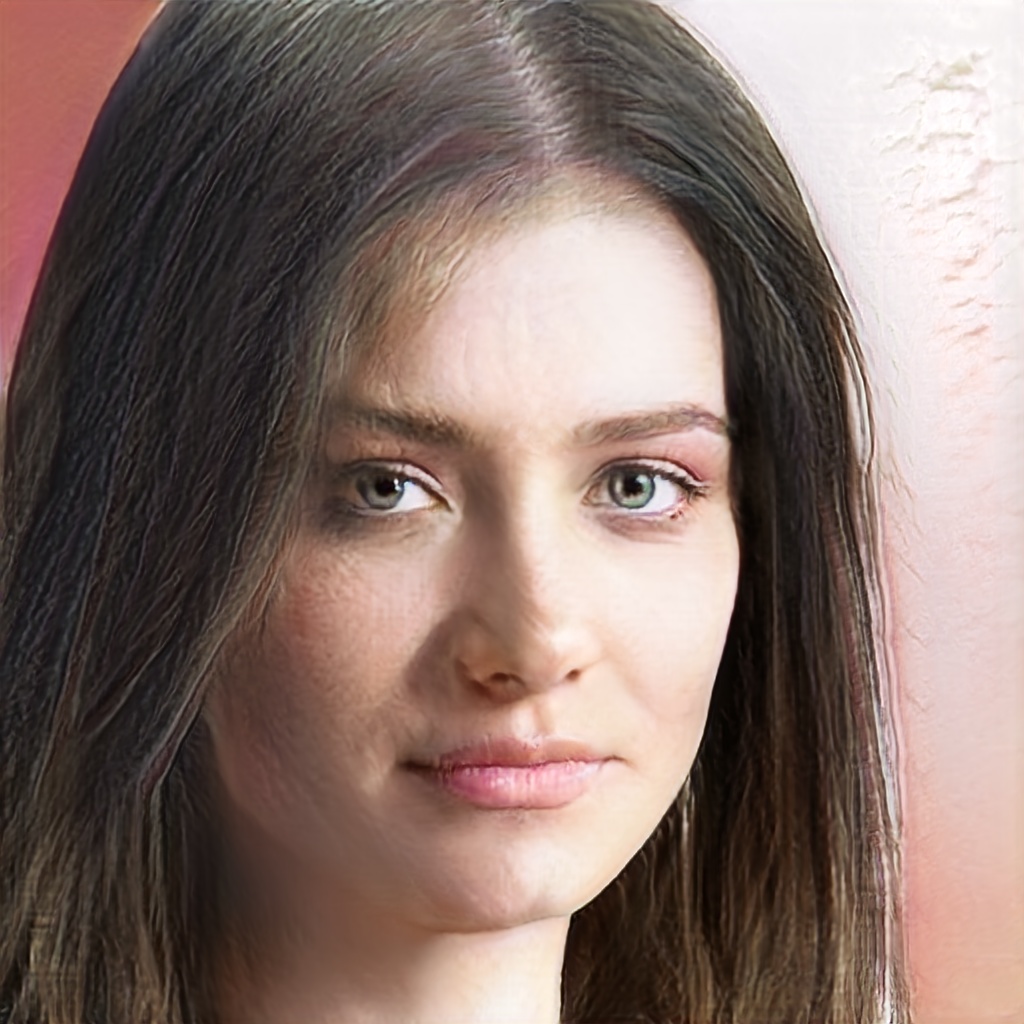} & \includegraphics[width=0.115\linewidth]{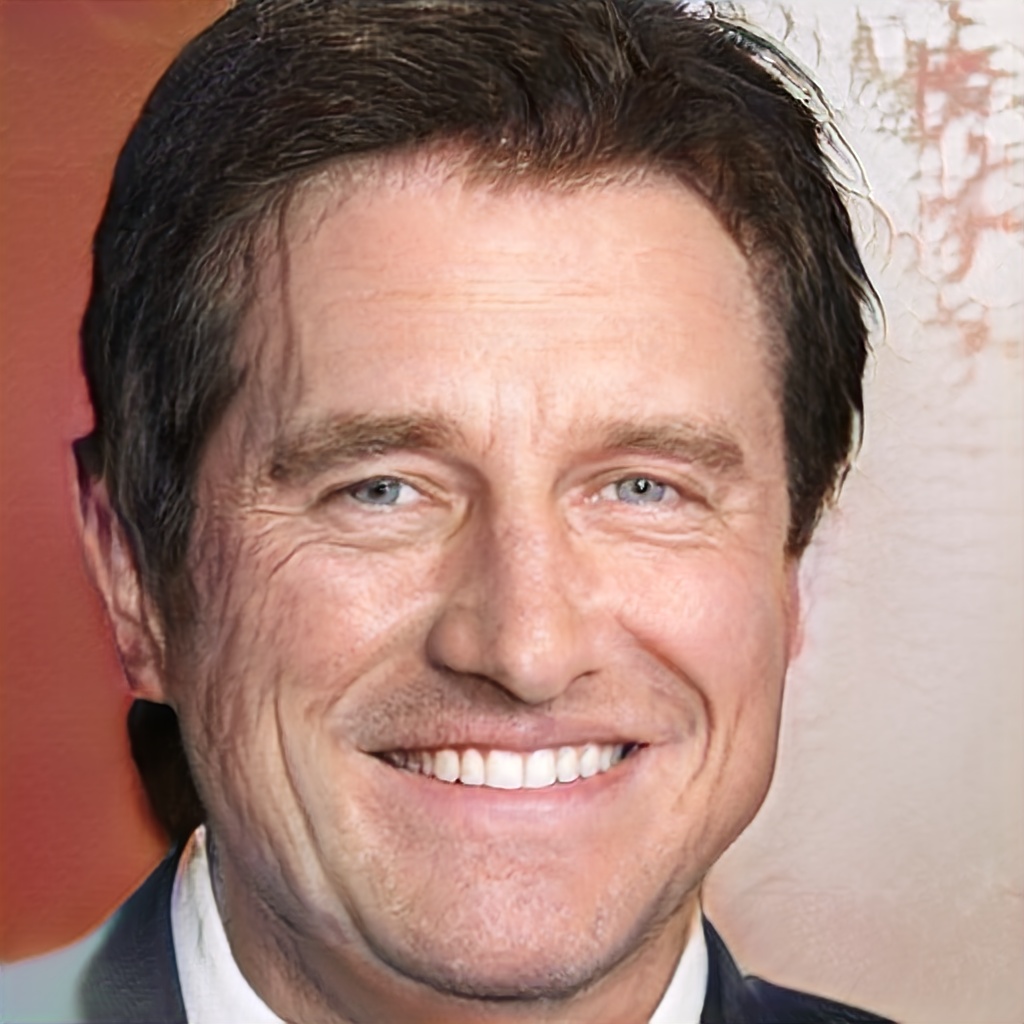} & \includegraphics[width=0.115\linewidth]{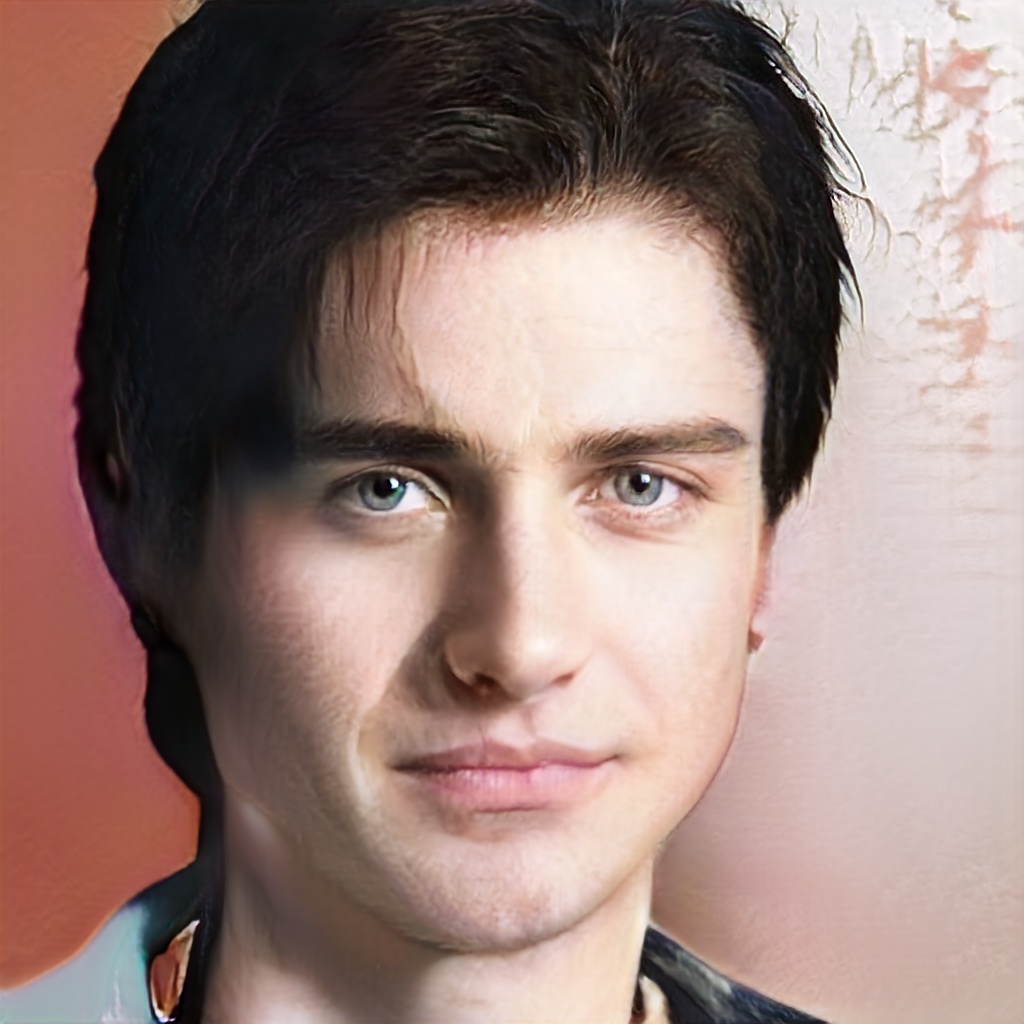} & \includegraphics[width=0.115\linewidth]{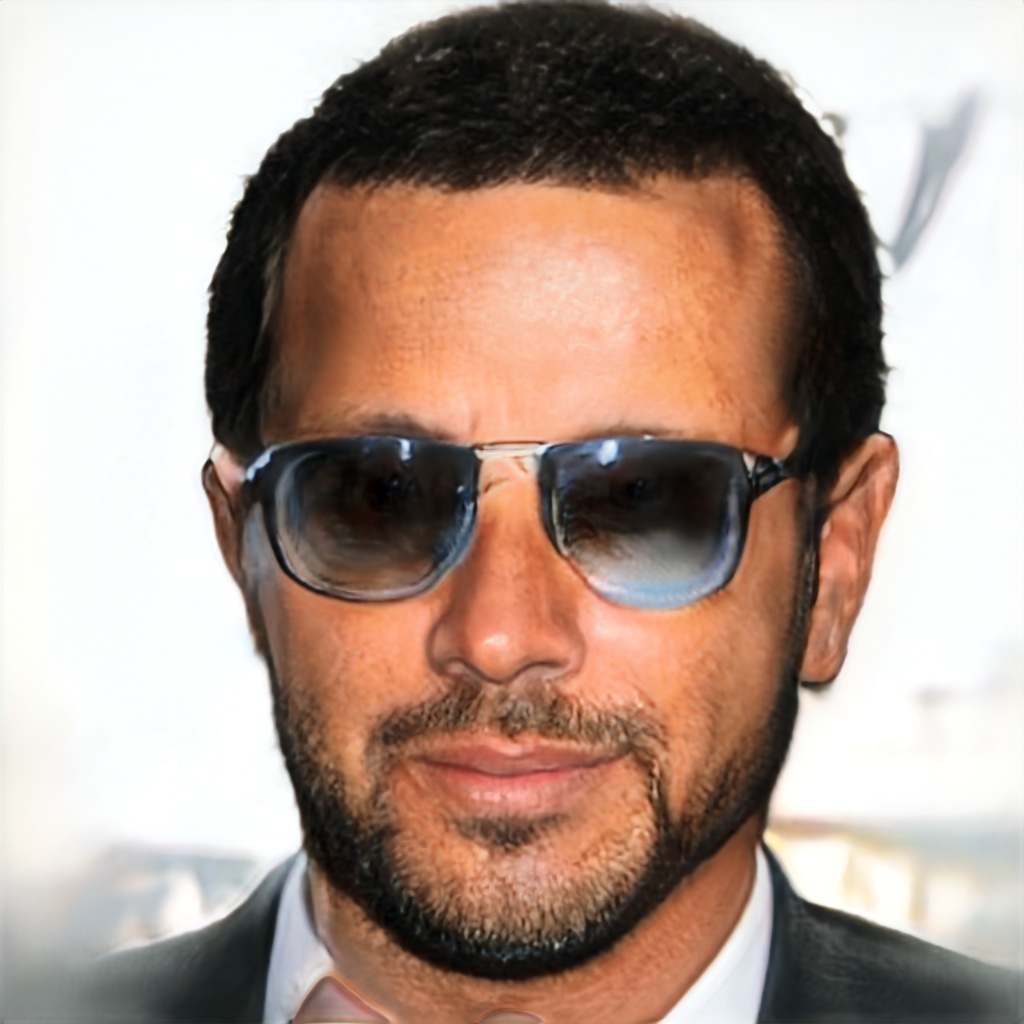}  & \includegraphics[width=0.115\linewidth]{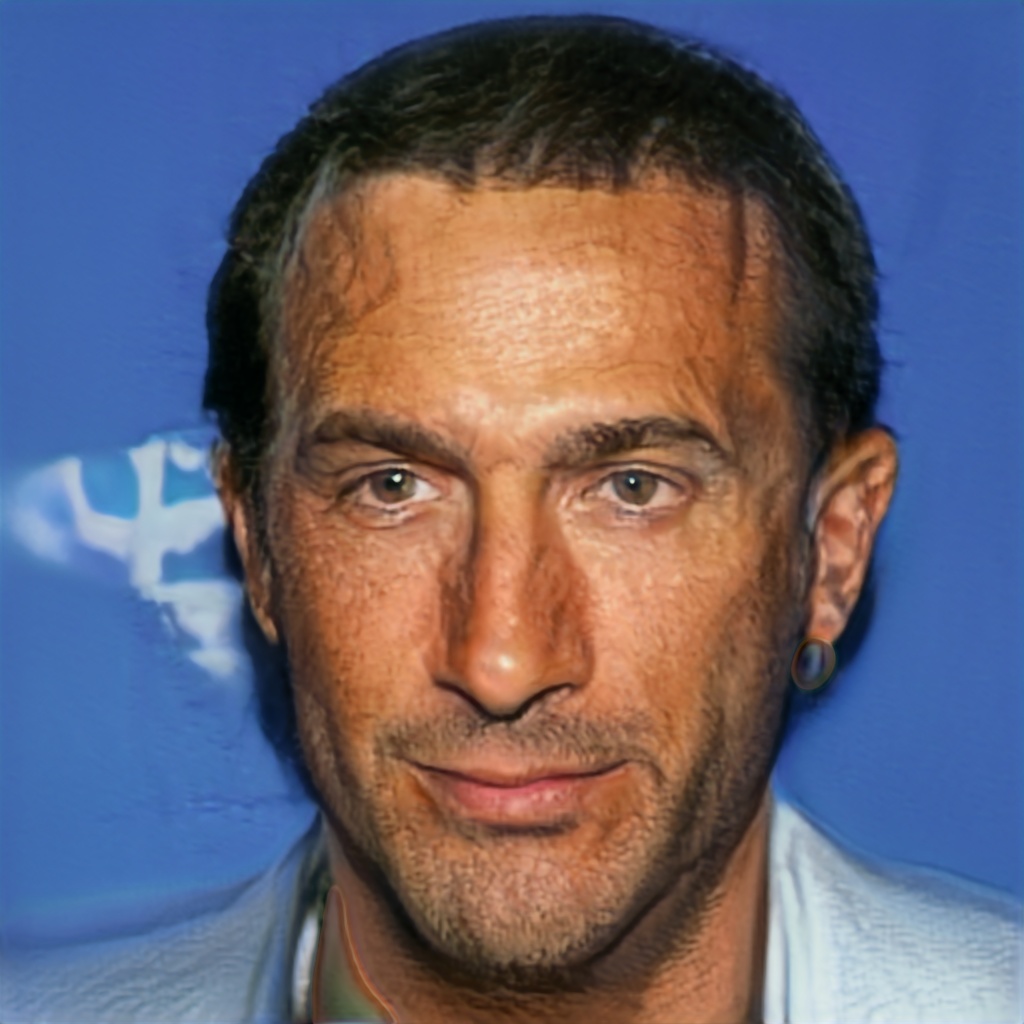} & \includegraphics[width=0.115\linewidth]{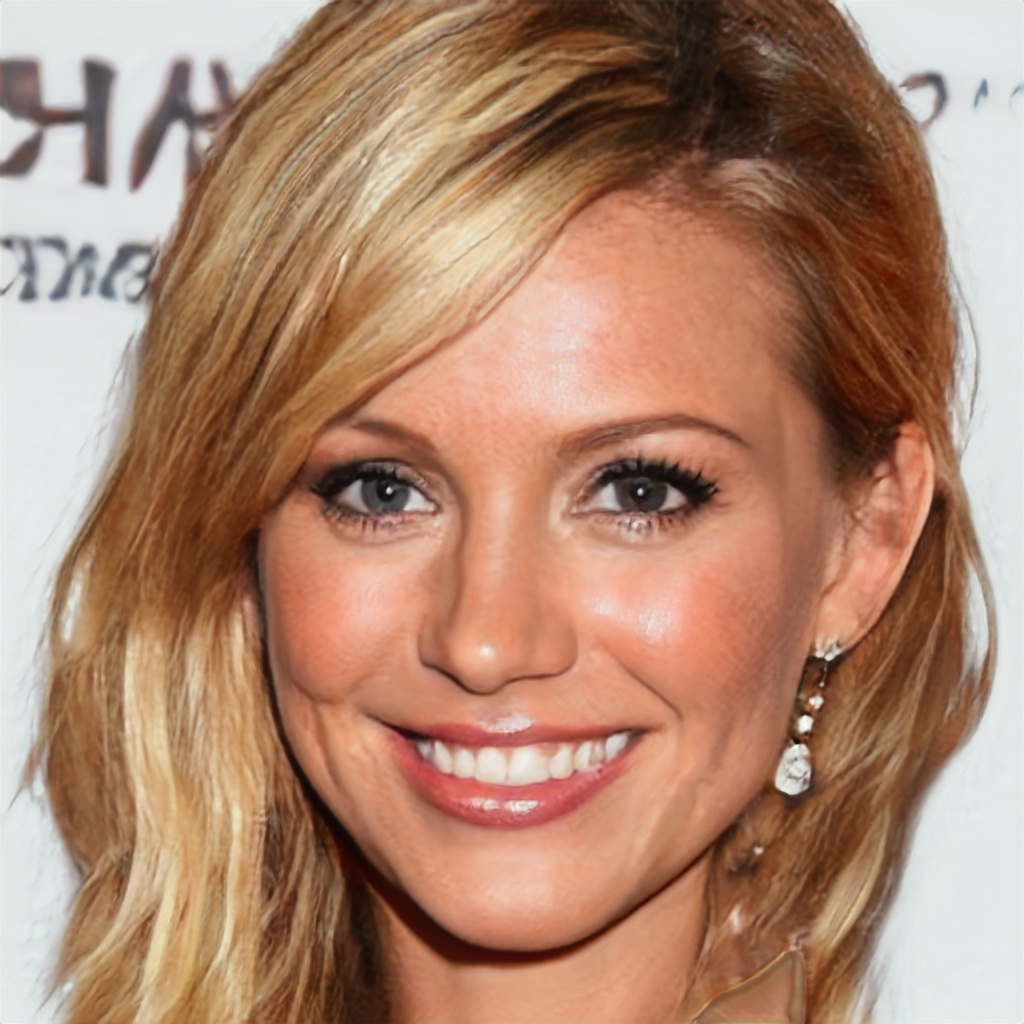} & \includegraphics[width=0.115\linewidth]{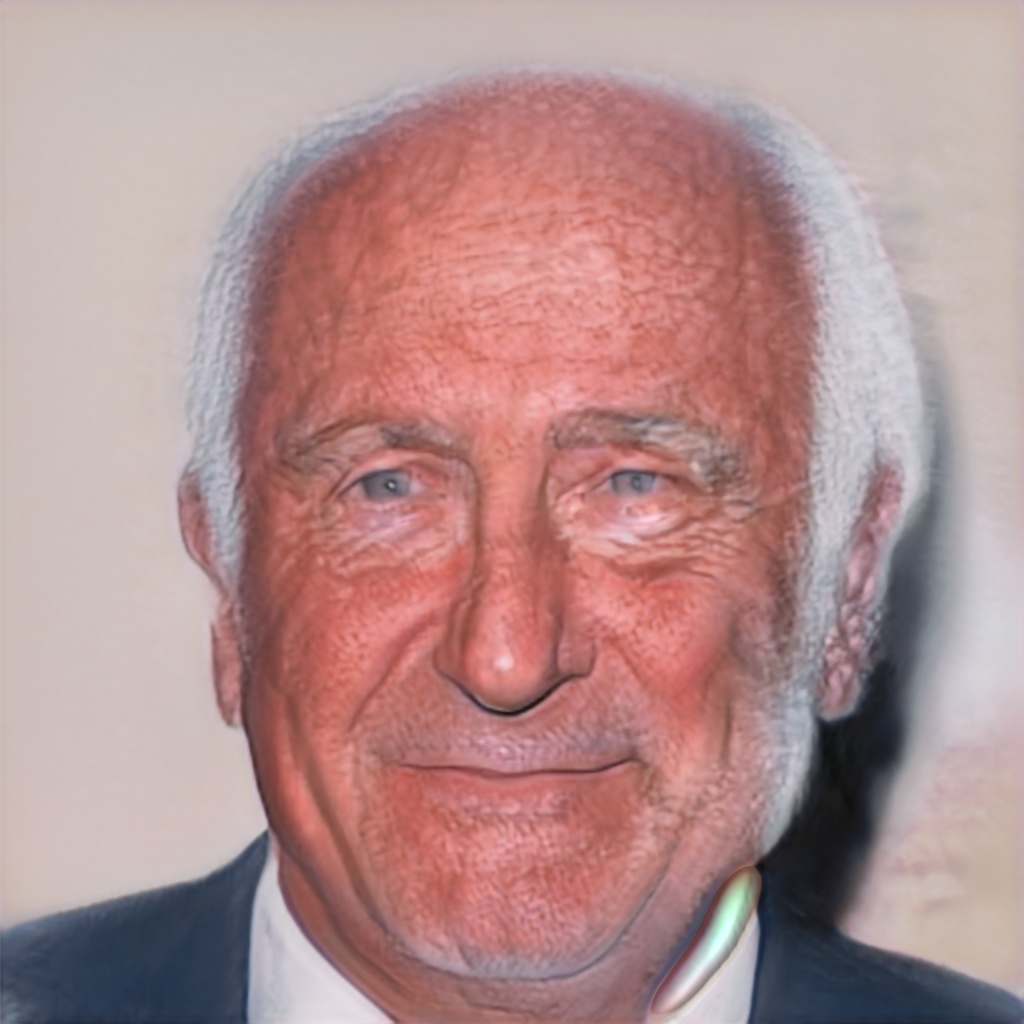} & \rot{Ours} \\
 & Eyeglasses  & Gender & Smile & Age &  Eyeglasses  & Gender & Smile & Age  & 
\end{tabular}

\caption{\textbf{Unconditional} attribute manipulation on PGGAN (left) and StyleGAN (right) with respect to Eyeglasses, Gender, Smile, and Age. We compare our method to \textit{Linear} and InterfaceGAN.}
\label{fig:uncond}
\end{figure*}

\vspace{0.1cm}\noindent\textbf{Unconditional Manipulation.} In the unconditional setting, we expect to see an effective transition as compared to the baselines since our method dynamically decides the moving direction in each step and follows the steepest direction. Note that although the main objective here is to change the target attribute, other attributes might change arbitrarily. Figure~\ref{fig:uncond} compares our method to two baselines, \textit{Linear} and InterfaceGAN, on 4 attributes. On PGGAN, all three methods successfully edit the attributes, while our method produces transition with much less distortion, e.g., smile and age. Also, we observe that our method and InterfaceGAN perform similarly on eyeglasses and gender. It may imply that for some attributes, the underlying manifold expands linearly as is derived by InterfaceGAN. 

On the other hand, our method behaves clearly differently from the two linear baselines on StyleGAN, showing that the manifold may be highly non-linear. We verify the benefit of the iterative scheme by comparing our method to the baselines. First, \textit{Linear} appears to fail to edit smiling even if it uses the same classifier as our framework. When comparing to InterfaceGAN, our method preserves more non-target contents than InterfaceGAN although the goal is not to preserve attributes. For example, age changes when editing eyeglasses, and hair color changes when editing smiles.

\begin{figure*}[htbp]
    \centering
    \includegraphics[width=\linewidth]{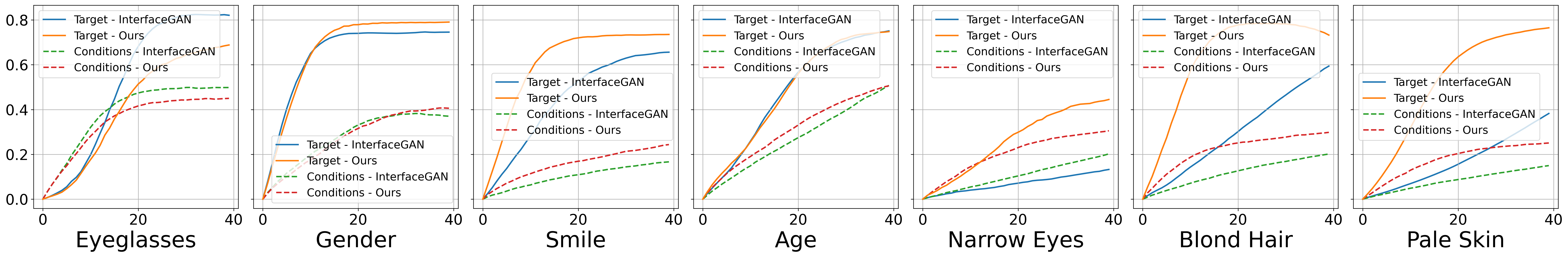}
    \caption{Logit changes over steps on StyleGAN. From left to right: Eyeglasses, Gender, Smile, Age, Narrow Eyes, Blond Hair, and Pale Skin. The solid lines represent the predictions of the target attributes while the dot lines represent the mean values over all the other non-target attribute predictions. Zoom in for better visualization.}
    \label{fig:logit_stylegan_uncond}
\end{figure*}

\begin{figure*}[htbp]
\setlength\tabcolsep{0.25em}
\centering
\begin{tabular}{ccccccc}
 \includegraphics[width=0.15\linewidth]{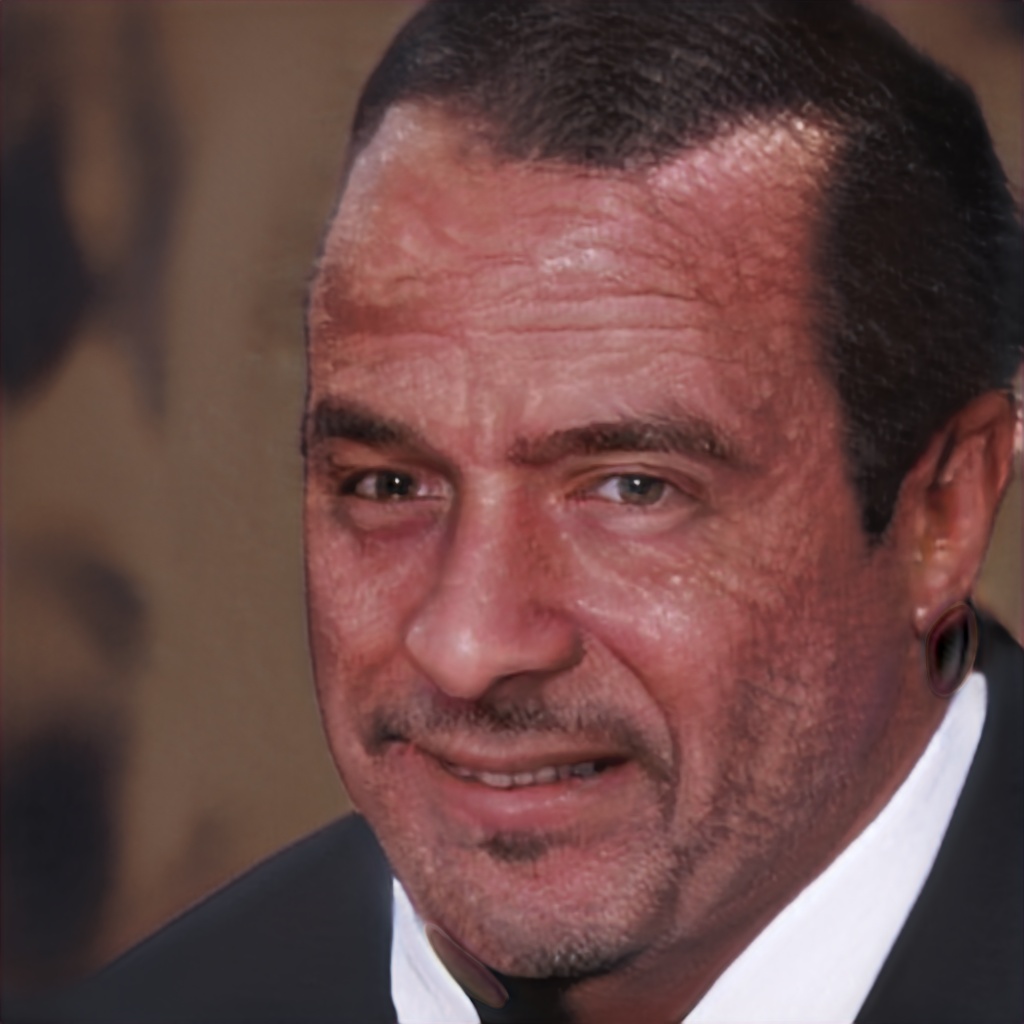} & \includegraphics[width=0.15\linewidth]{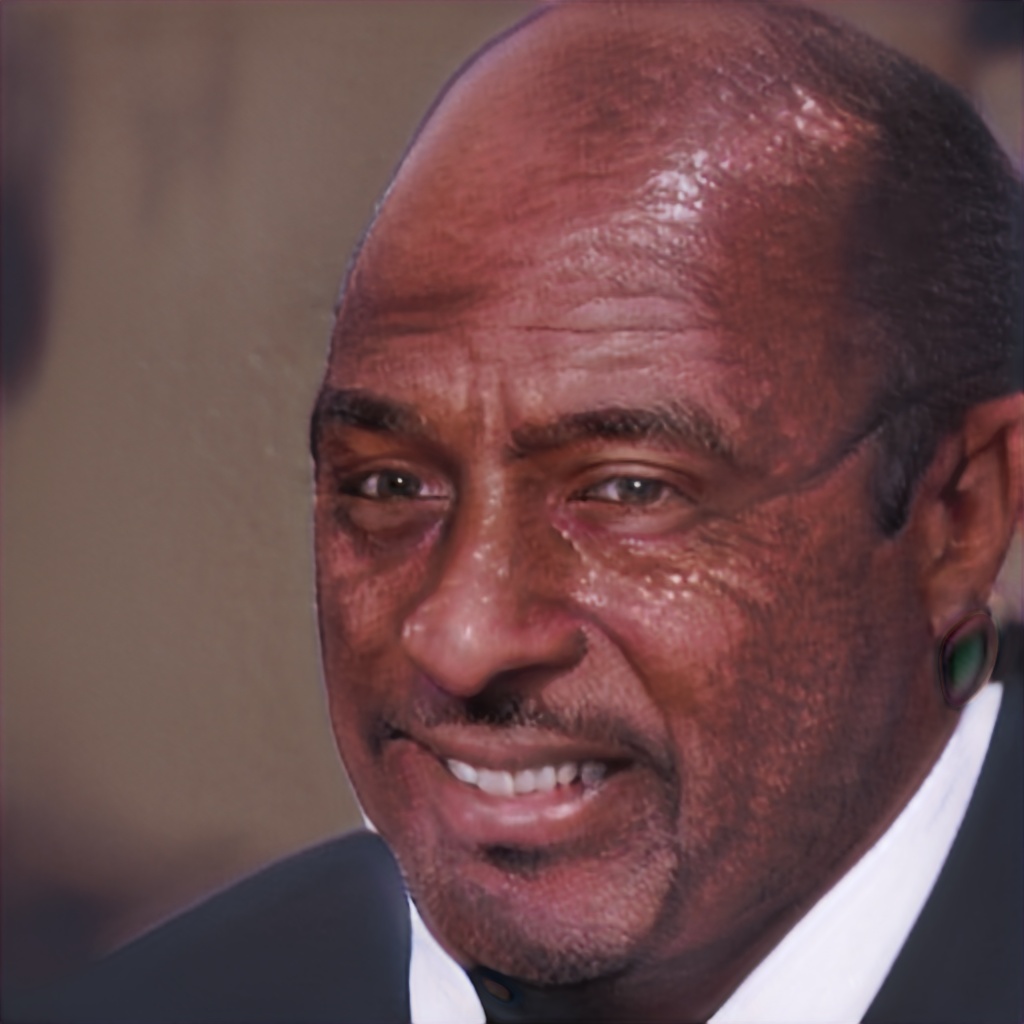} & \includegraphics[width=0.15\linewidth]{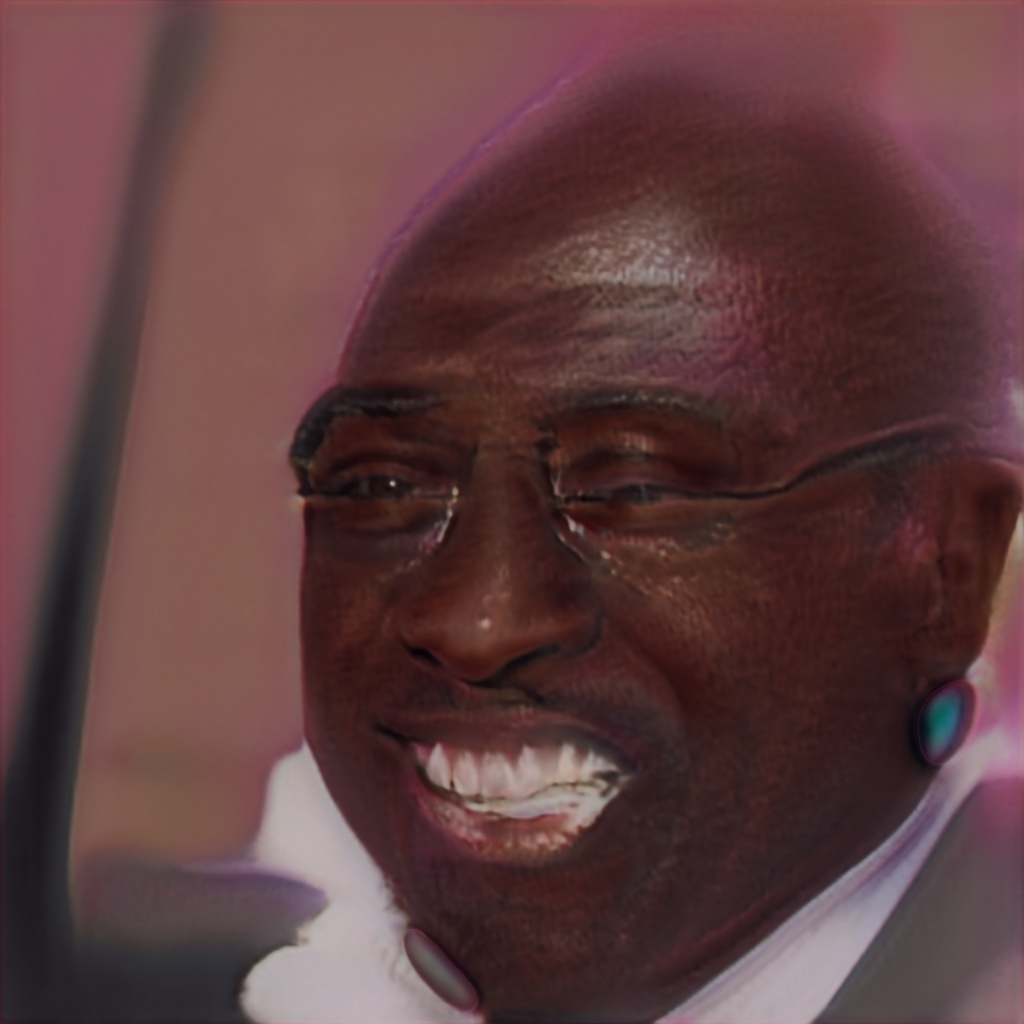} &  & \includegraphics[width=0.15\linewidth]{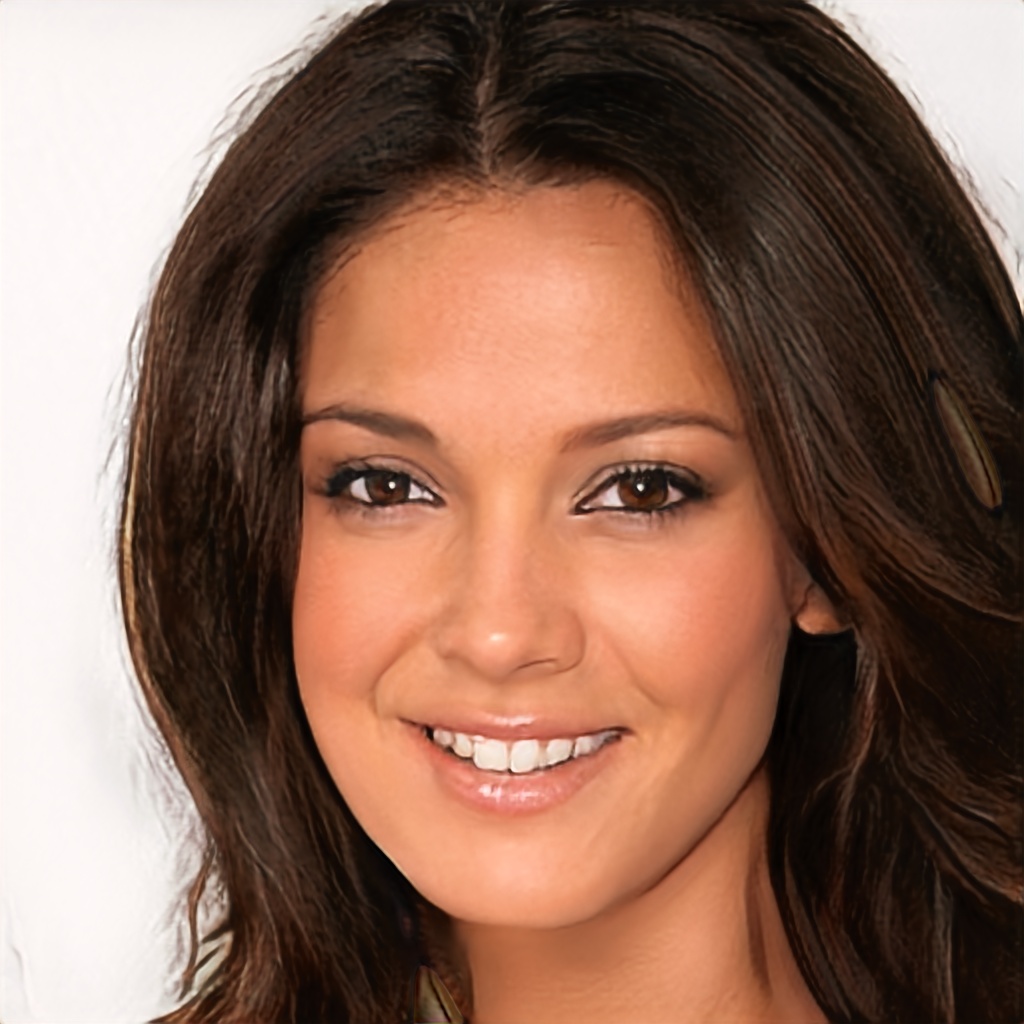} & \includegraphics[width=0.15\linewidth]{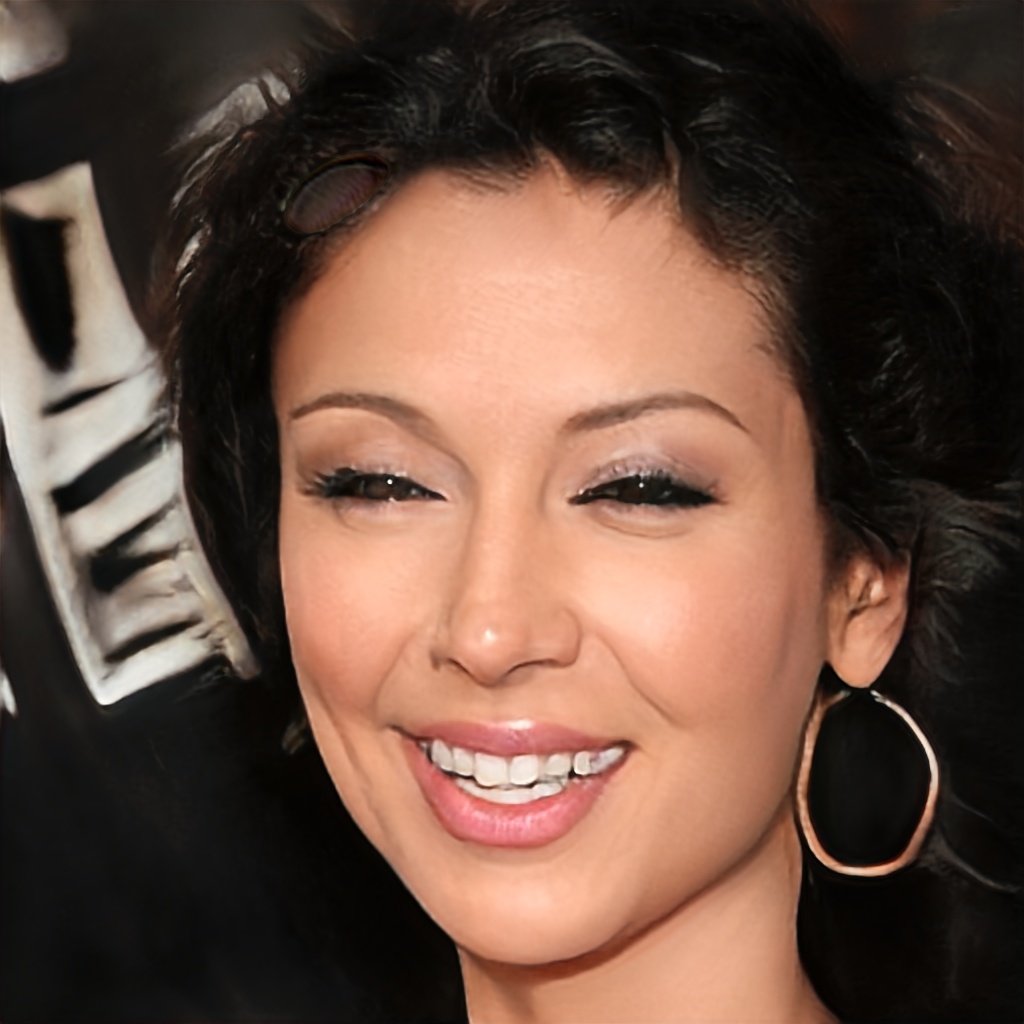} & \includegraphics[width=0.15\linewidth]{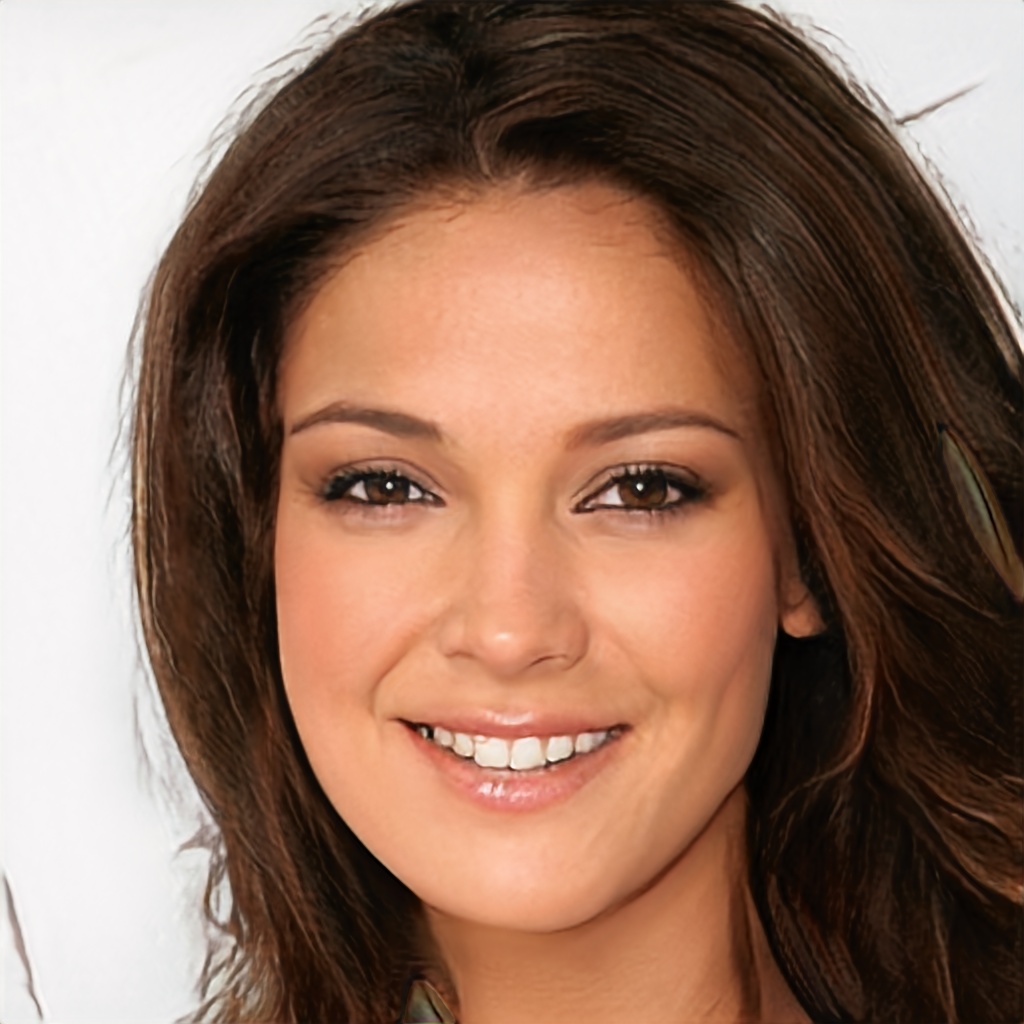} \\
 \multicolumn{3}{c}{Bald} & & \multicolumn{3}{c}{Narrow Eyes} \\
 \includegraphics[width=0.15\linewidth]{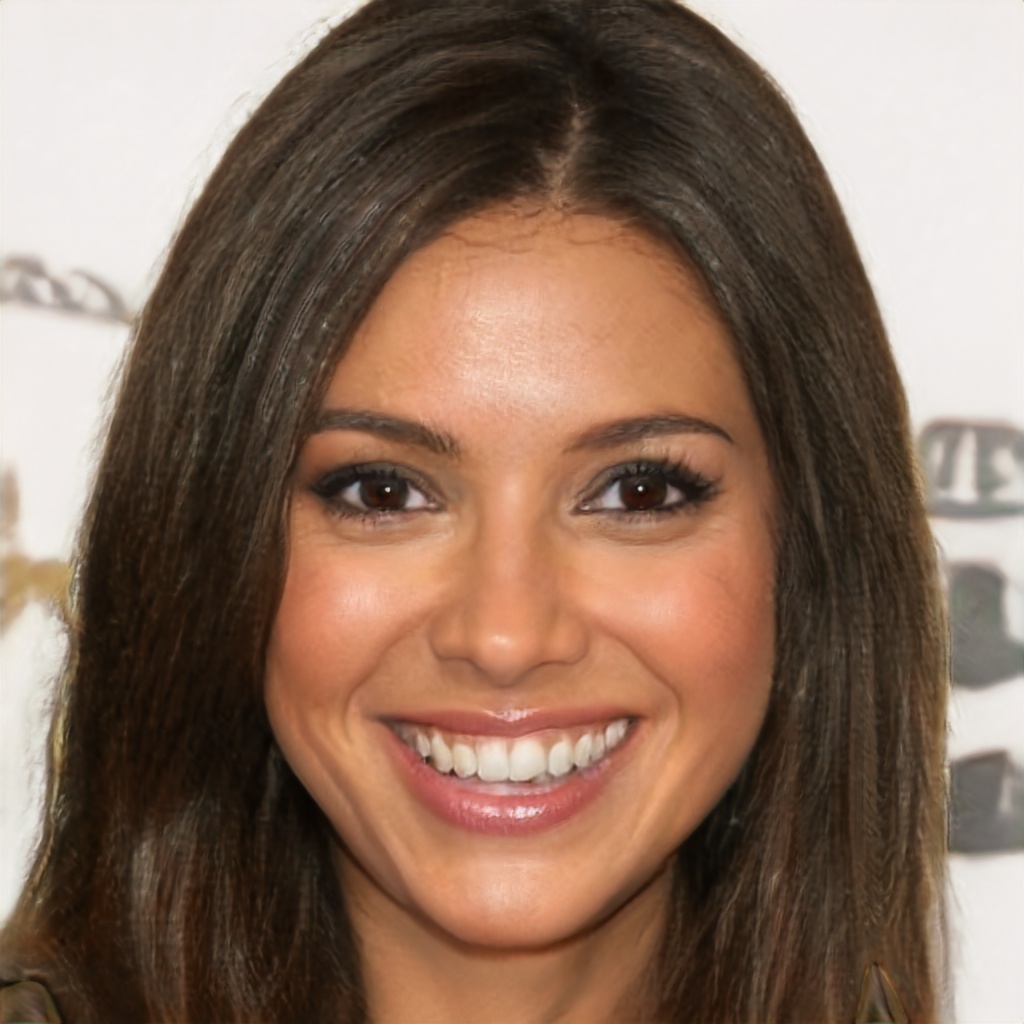} & \includegraphics[width=0.15\linewidth]{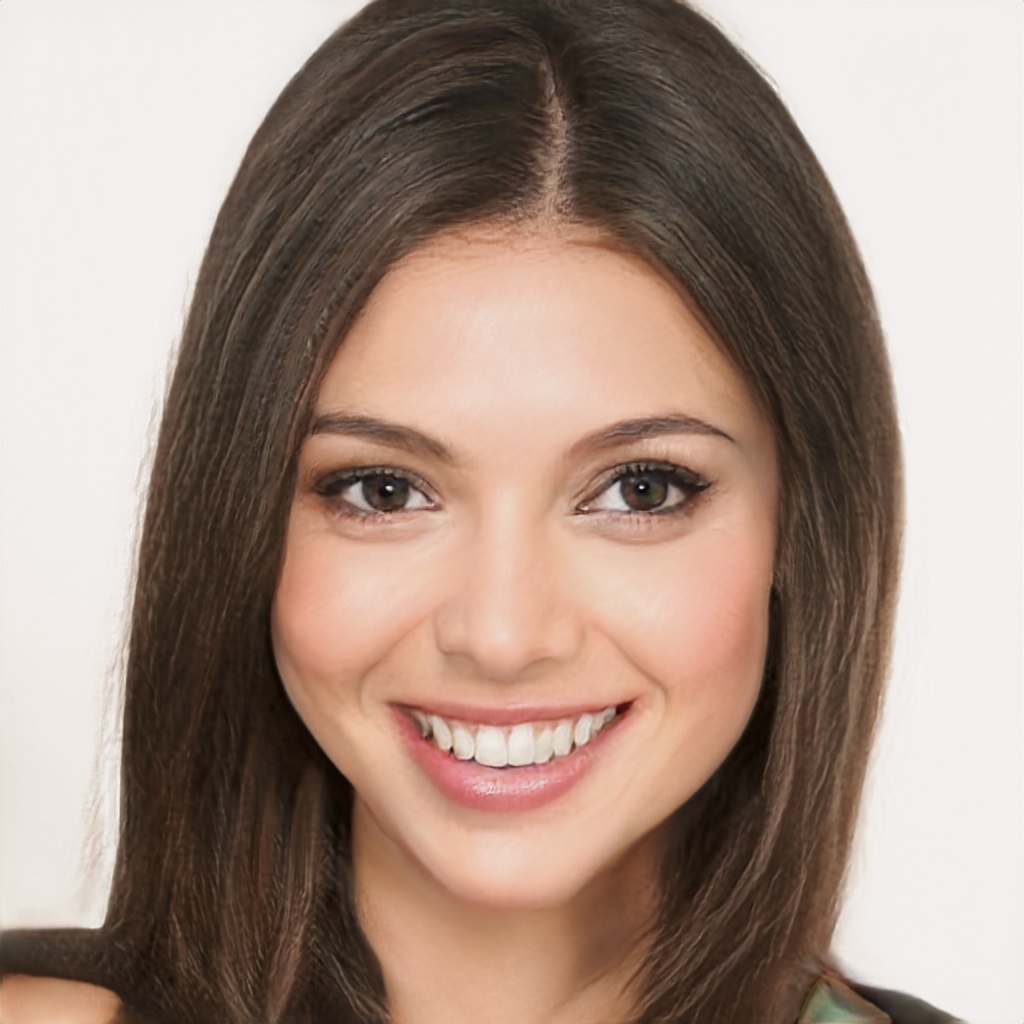} & \includegraphics[width=0.15\linewidth]{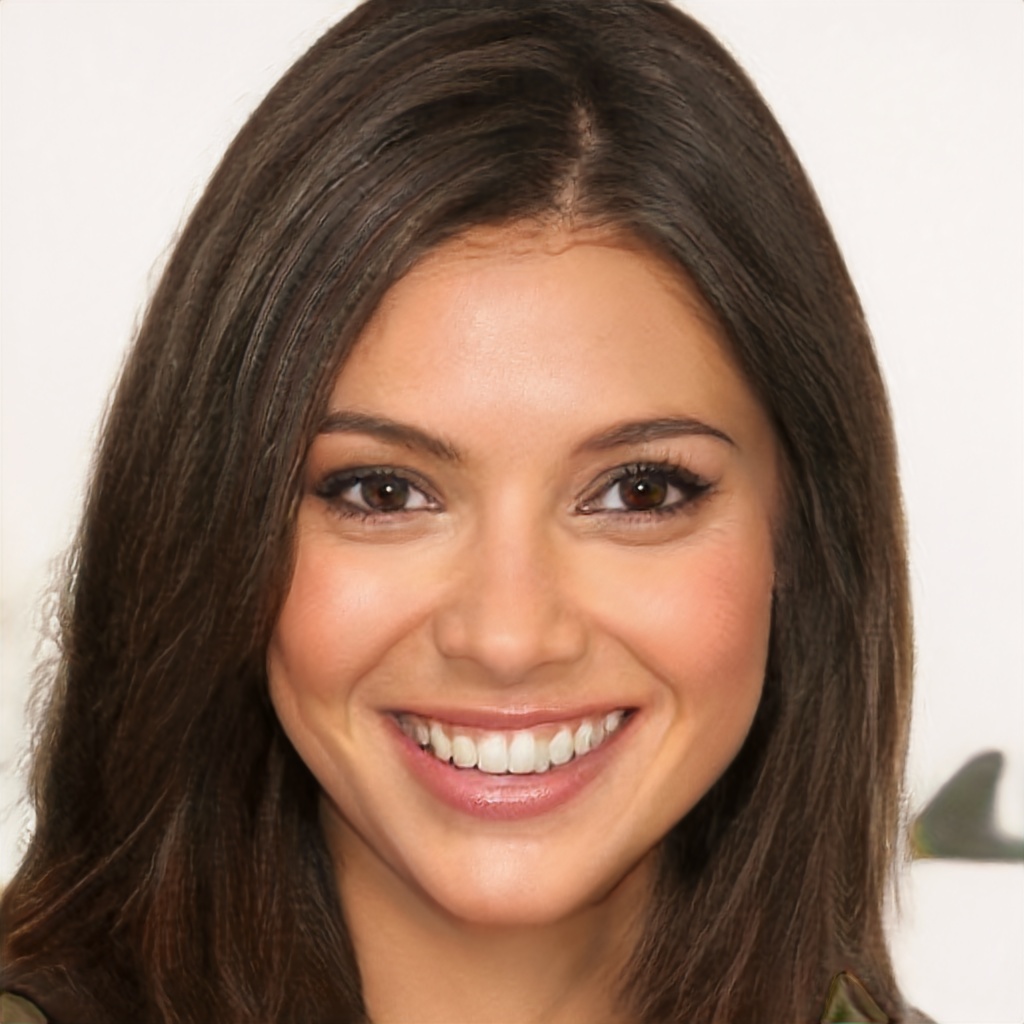} & &
 \includegraphics[width=0.15\linewidth]{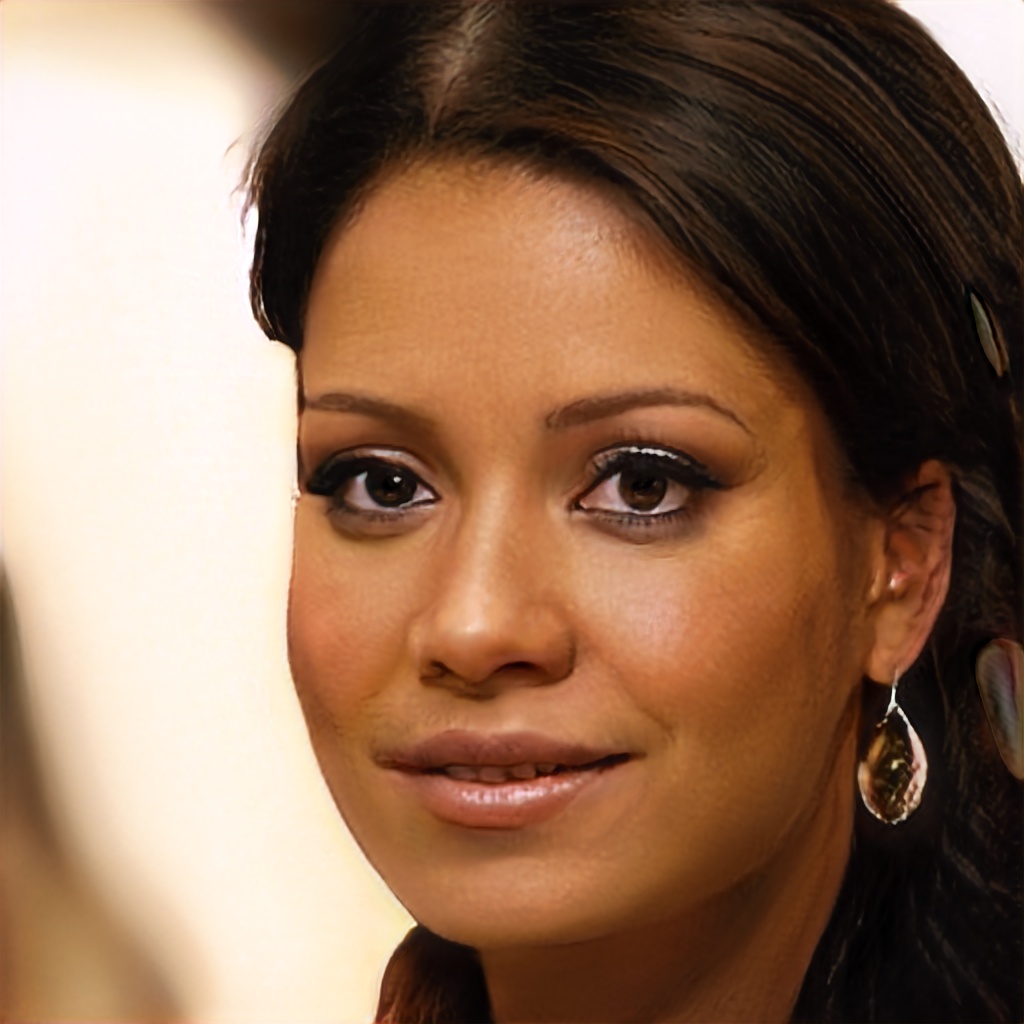} & \includegraphics[width=0.15\linewidth]{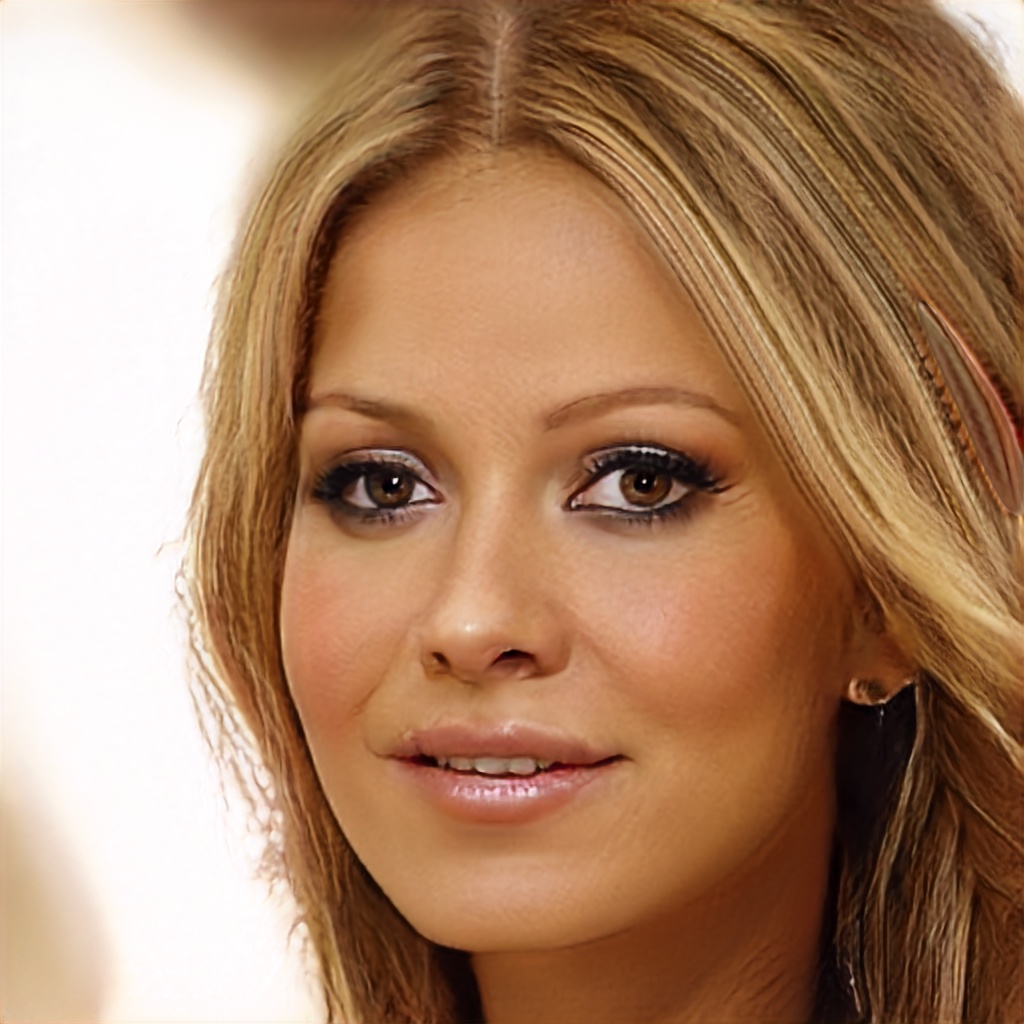} & \includegraphics[width=0.15\linewidth]{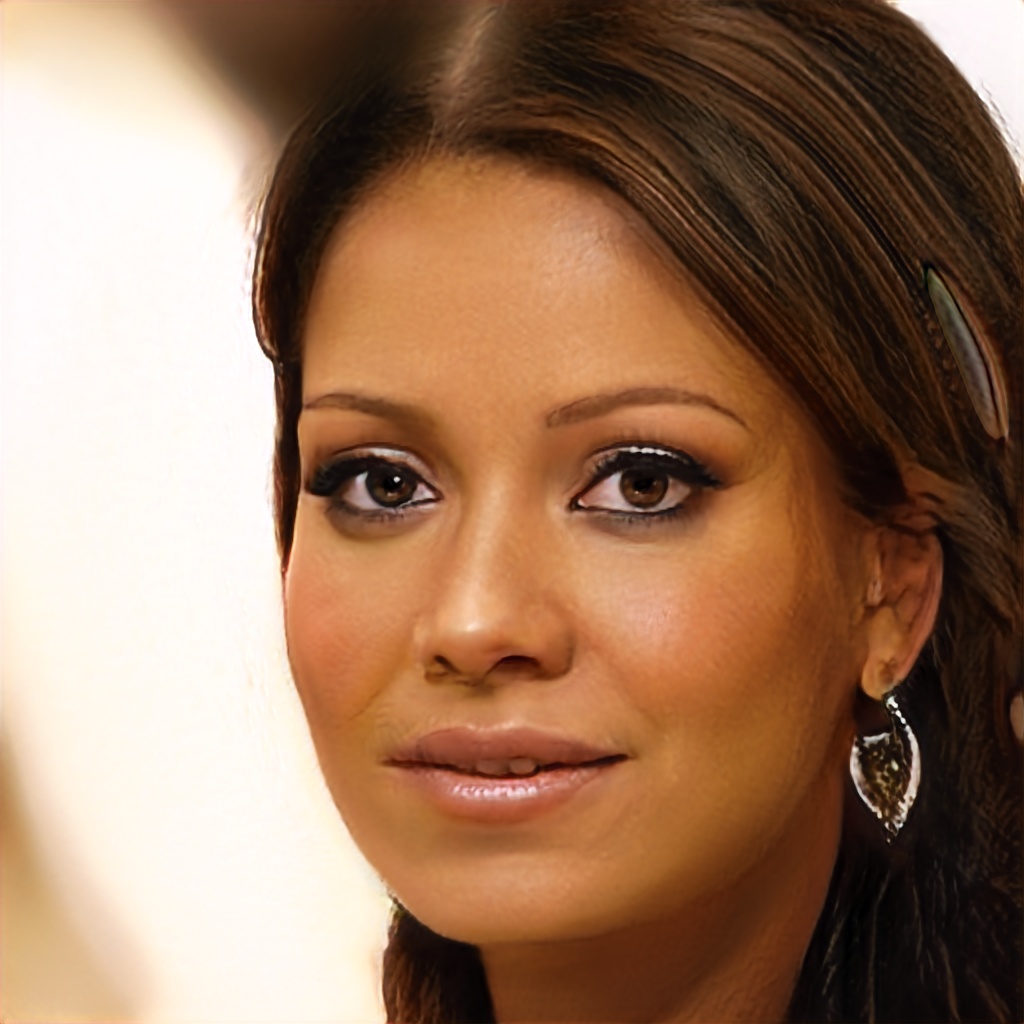} \\
 \multicolumn{3}{c}{Pale Skin} & & \multicolumn{3}{c}{Blond Hair} \\
\end{tabular}
\caption{\textbf{Unconditional} rare attribute manipulation on StyleGAN with respect to Bald, Narrow Eyes, Pale Skin, and Blond Hair. Each set consists of input (left), our method (middle), and InterfaceGAN (right).}
\label{fig:rare_attrs}
\end{figure*}

\vspace{0.1cm}\noindent\textbf{Effectiveness.} To further verify the effectiveness of the proposed method, we compare our method to InterfaceGAN on logit changes over steps. We experiment on StyleGAN since it tends to be more non-linear as observed in Figure~\ref{fig:uncond}. Specifically, we choose 7 attributes and sample 2500 trajectories for each attribute, each of which consists of 40 steps with the step size equal to 0.2. Next, we compute the mean values over trajectories and report the target and non-target logit changes, respectively. 

Figure~\ref{fig:logit_stylegan_uncond} shows that our method can rapidly change the target attribute on most attributes, especially on rare attributes such as blond hair, pale skin, and narrow eyes. The result aligns with our hypothesis that not all attributes in the latent space distribute linearly; thus, our framework can benefit from the non-linearity. We additionally visualize the manipulation over the rare attributes in Figure~\ref{fig:rare_attrs}, confirming the effectiveness of our method. In addition to effective manipulation, it is also seen that the non-target attributes of all methods are changed during the manipulation. It is thus crucial to limit such an undesired effect.

\begin{figure*}[htbp]
\setlength\tabcolsep{0.1em}
\centering
\begin{tabular}{ccccc:ccccc}
\multicolumn{5}{c:}{PGGAN} & \multicolumn{5}{c}{StyleGAN} \\
\rot{Input} & \includegraphics[width=0.115\linewidth]{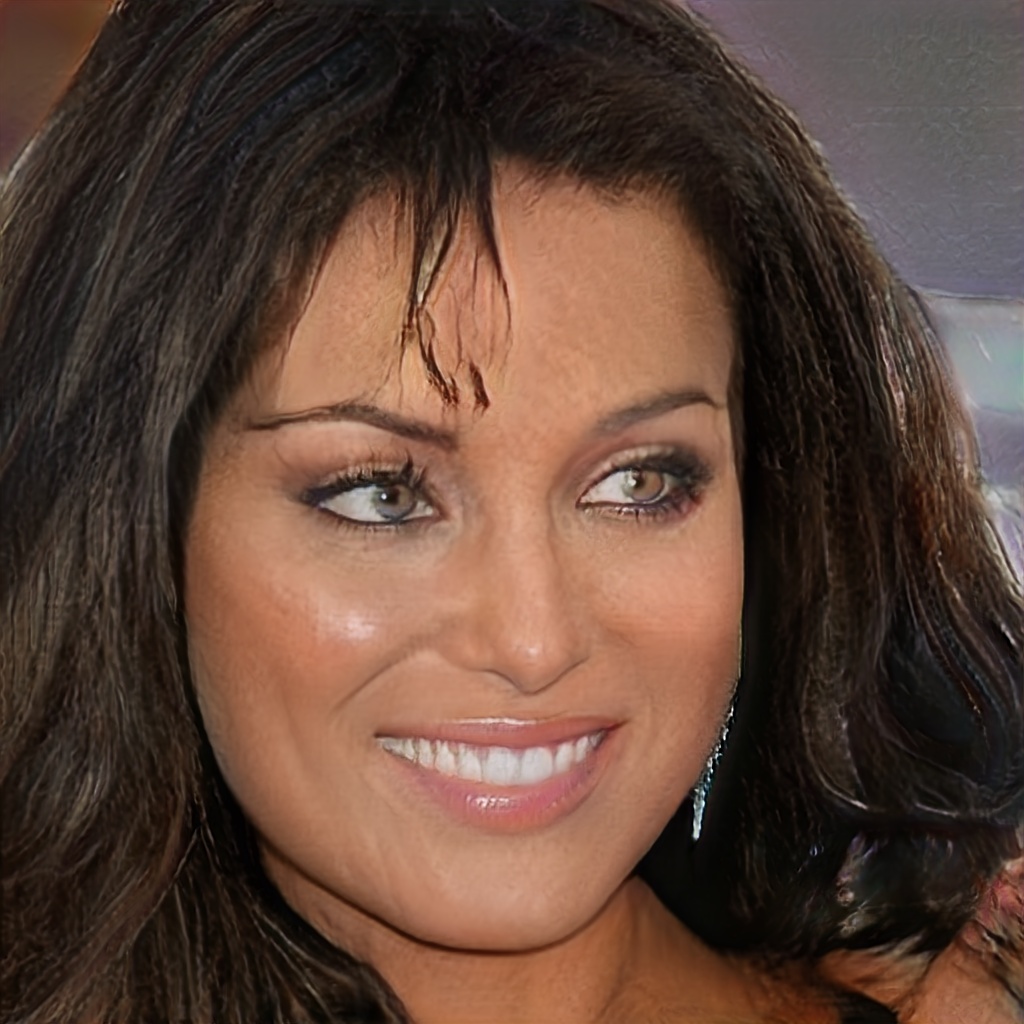} &  \includegraphics[width=0.115\linewidth]{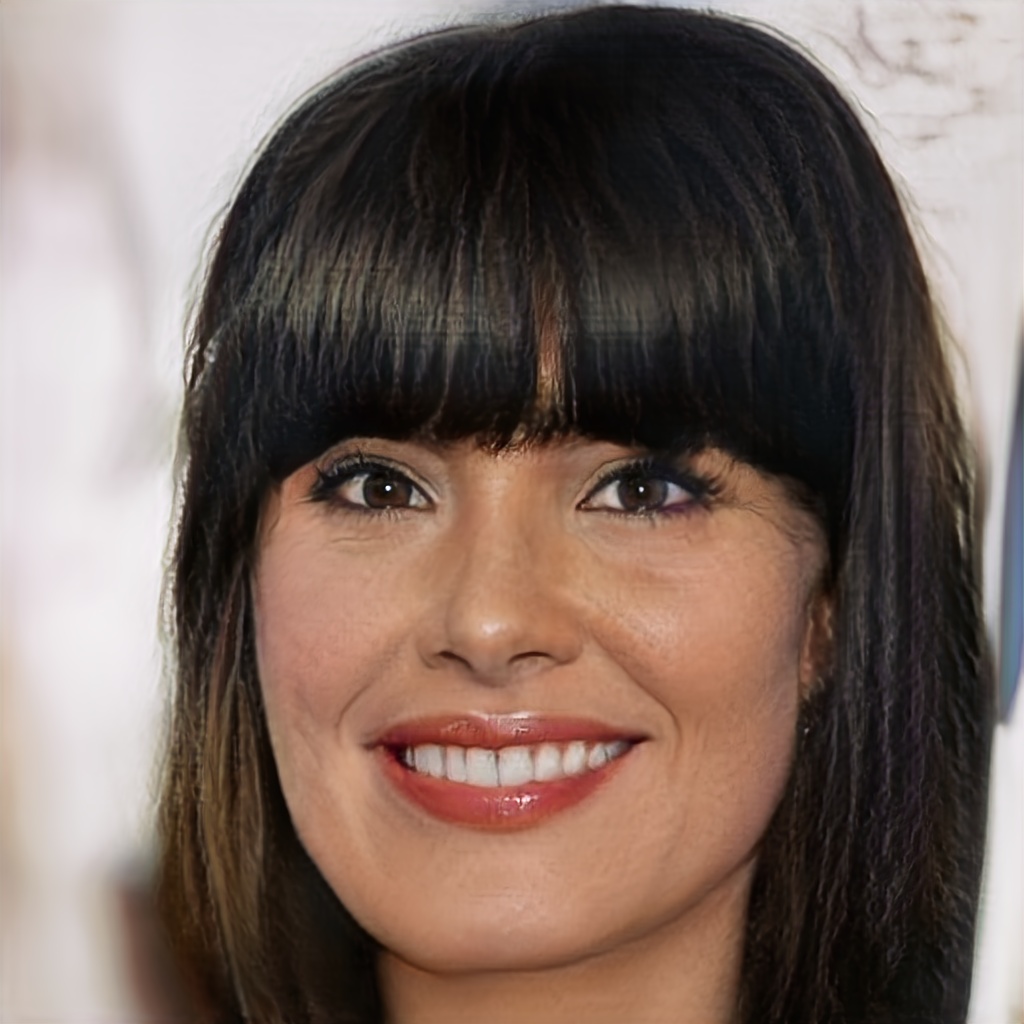} & \includegraphics[width=0.115\linewidth]{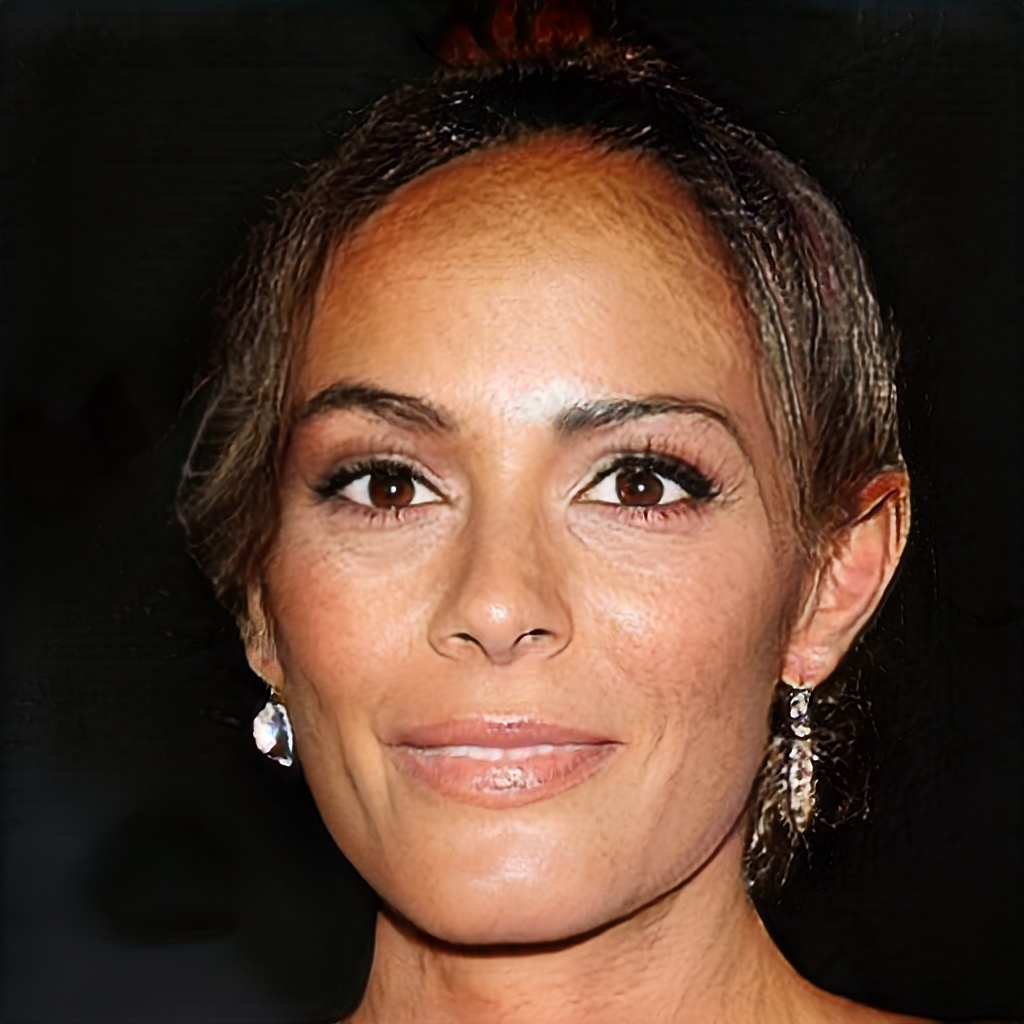} & \includegraphics[width=0.115\linewidth]{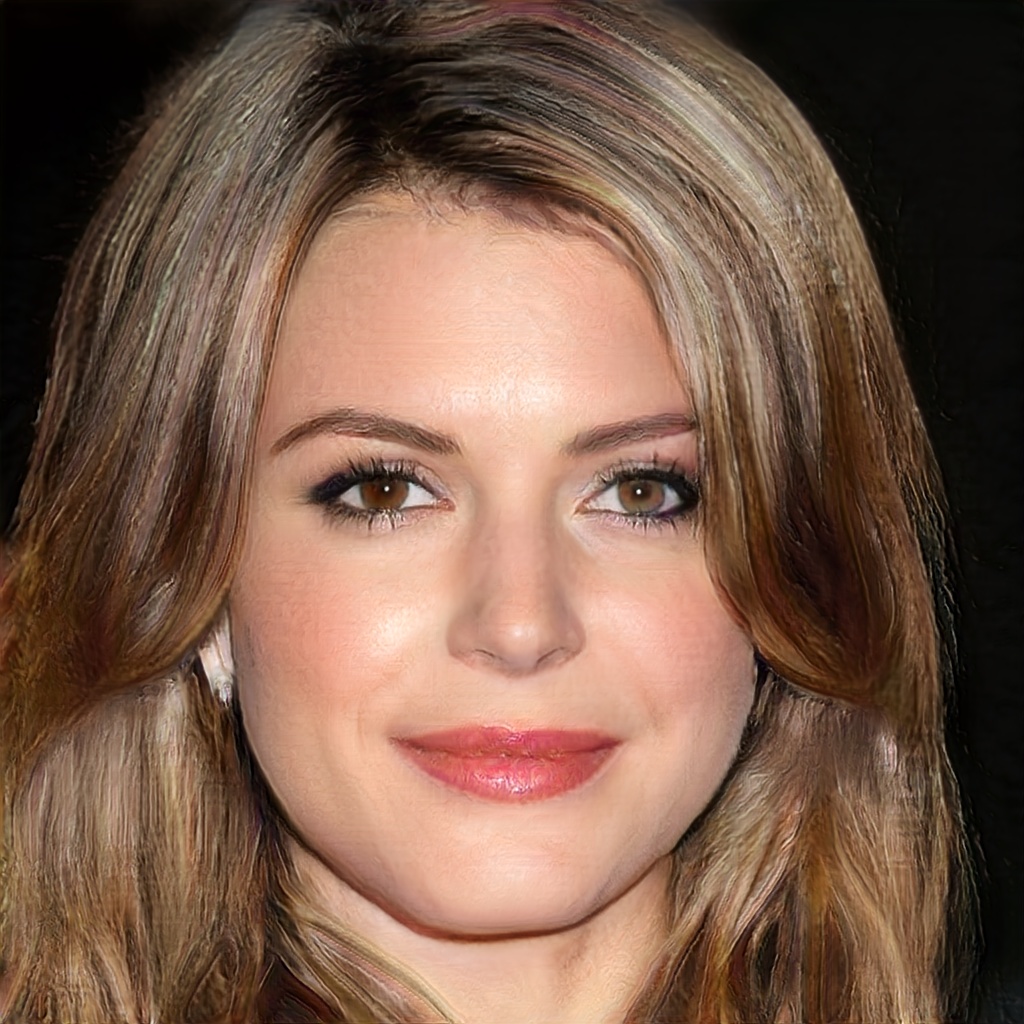} & \includegraphics[width=0.115\linewidth]{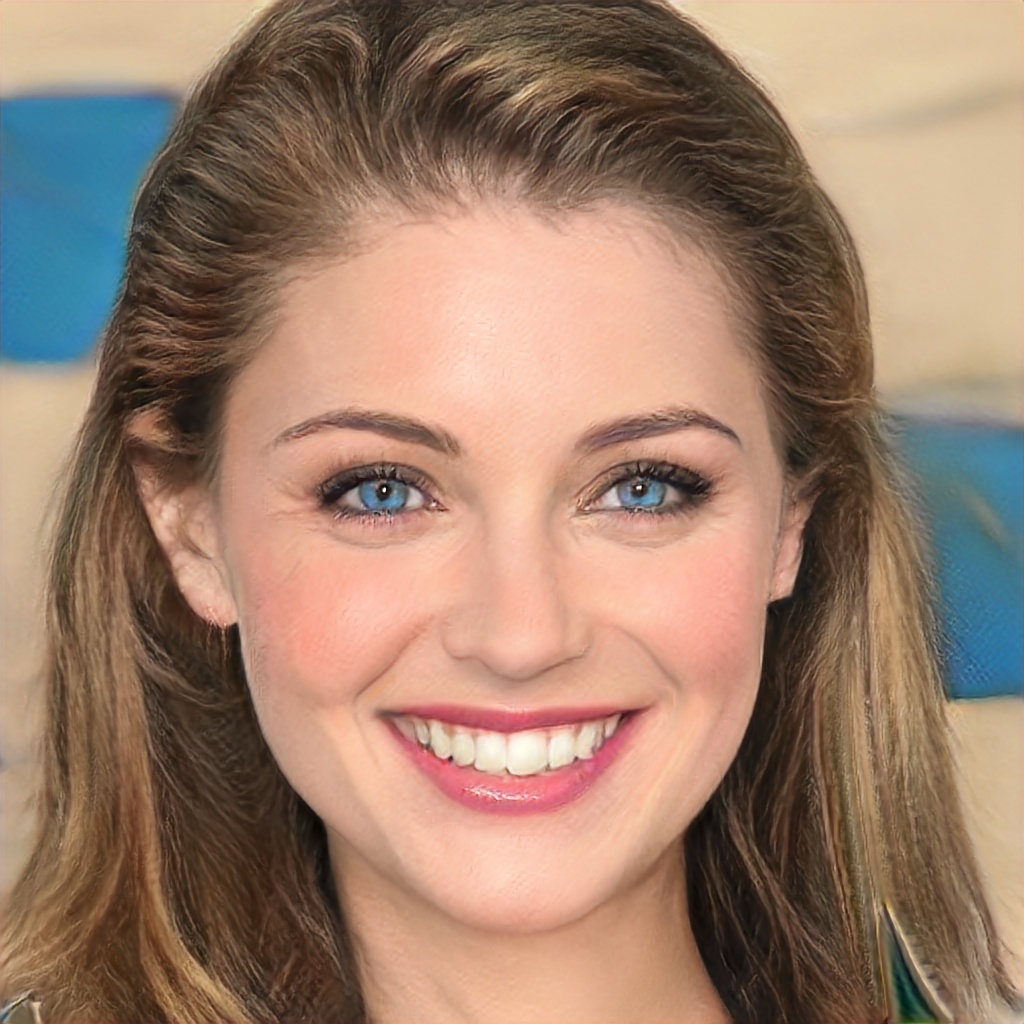} & \includegraphics[width=0.115\linewidth]{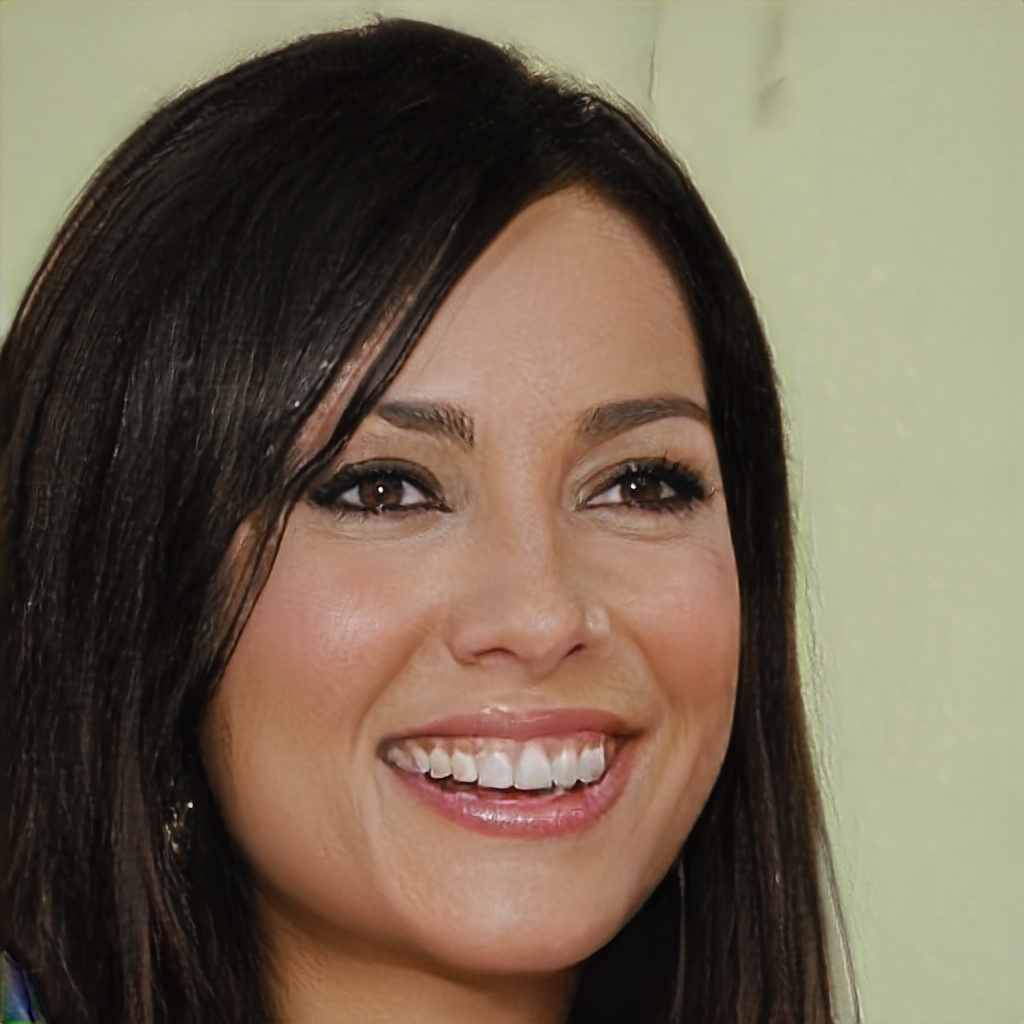} & \includegraphics[width=0.115\linewidth]{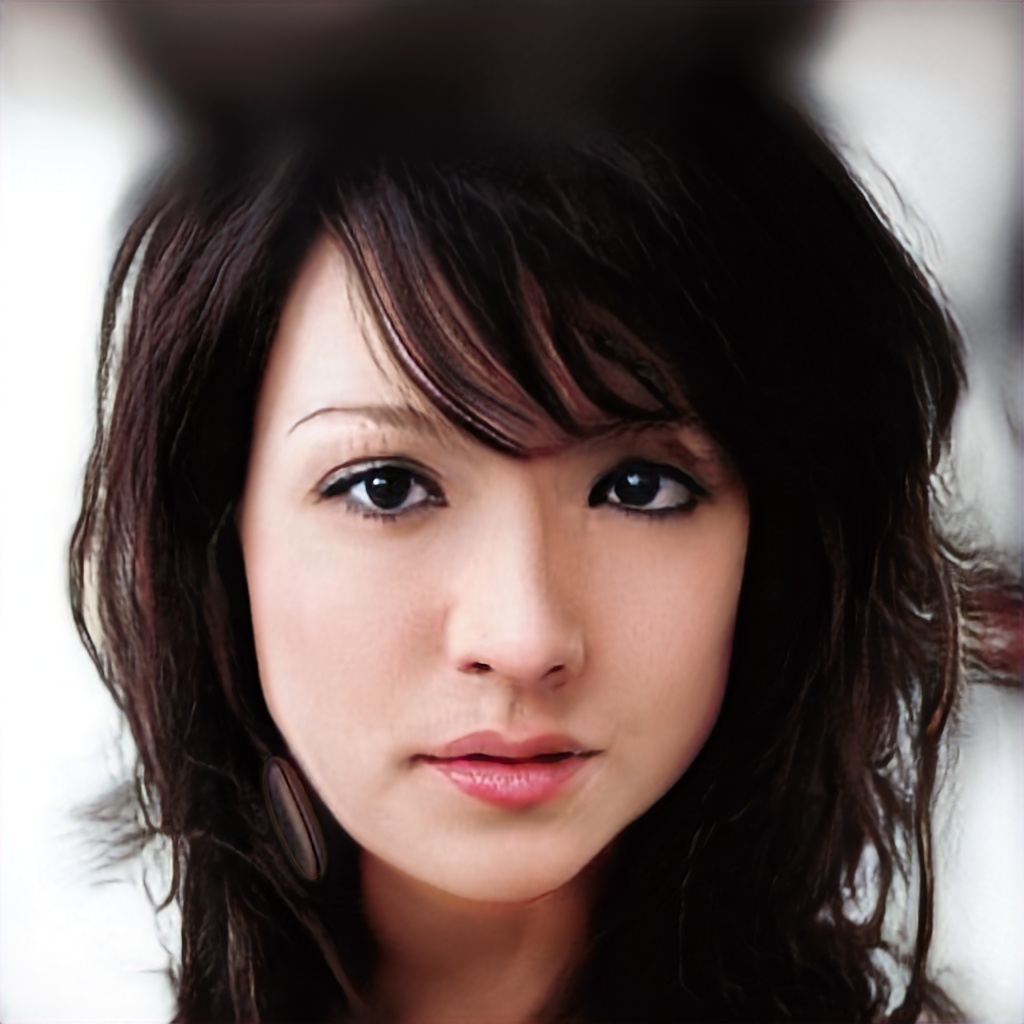} & \includegraphics[width=0.115\linewidth]{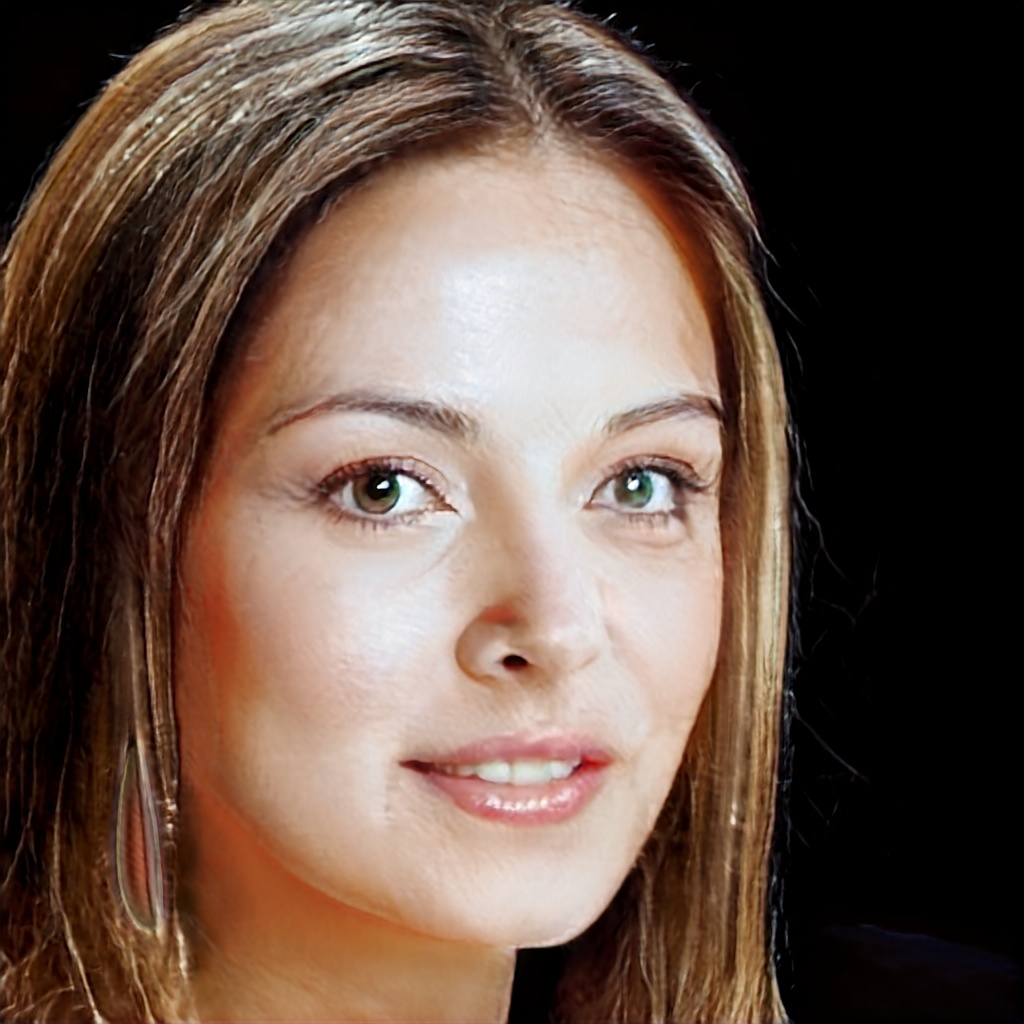} & \rot{Input} \\
\rot{InterfaceGAN} & \includegraphics[width=0.115\linewidth]{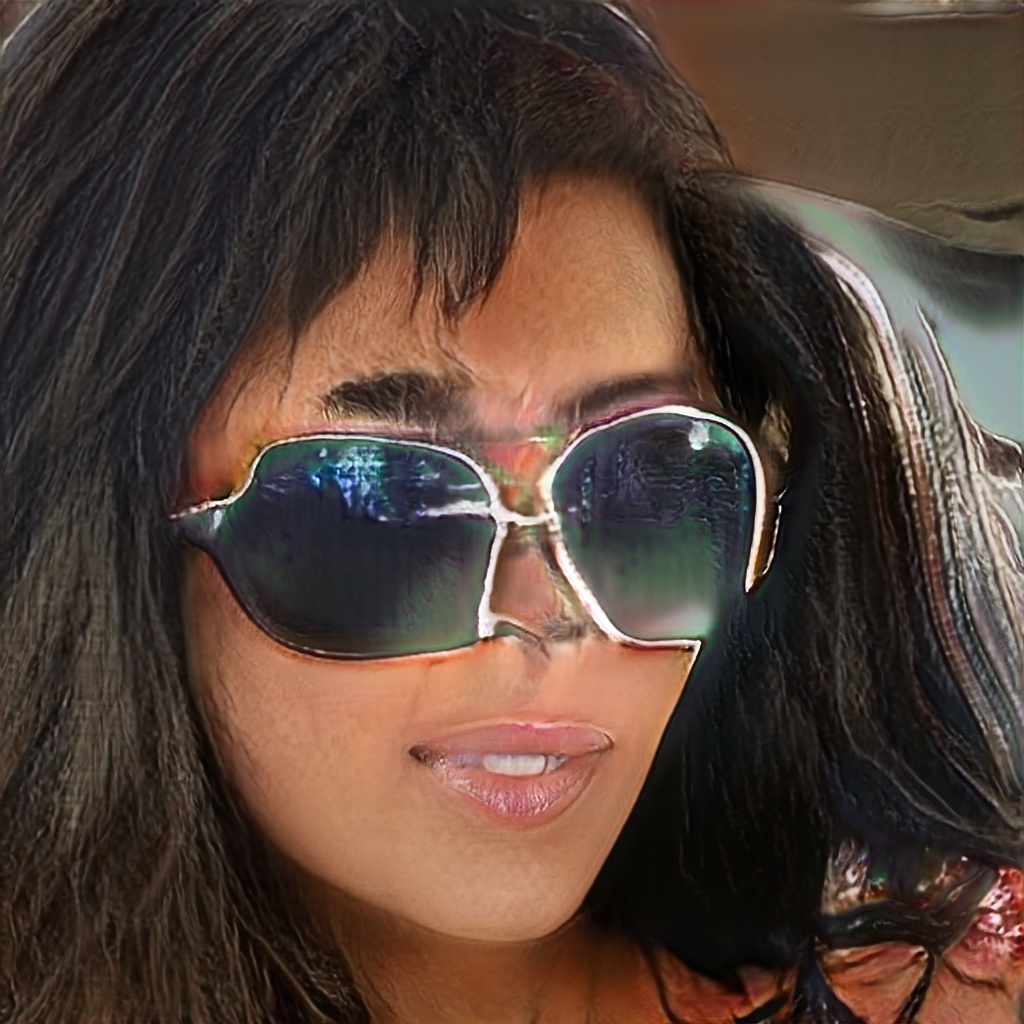} & \includegraphics[width=0.115\linewidth]{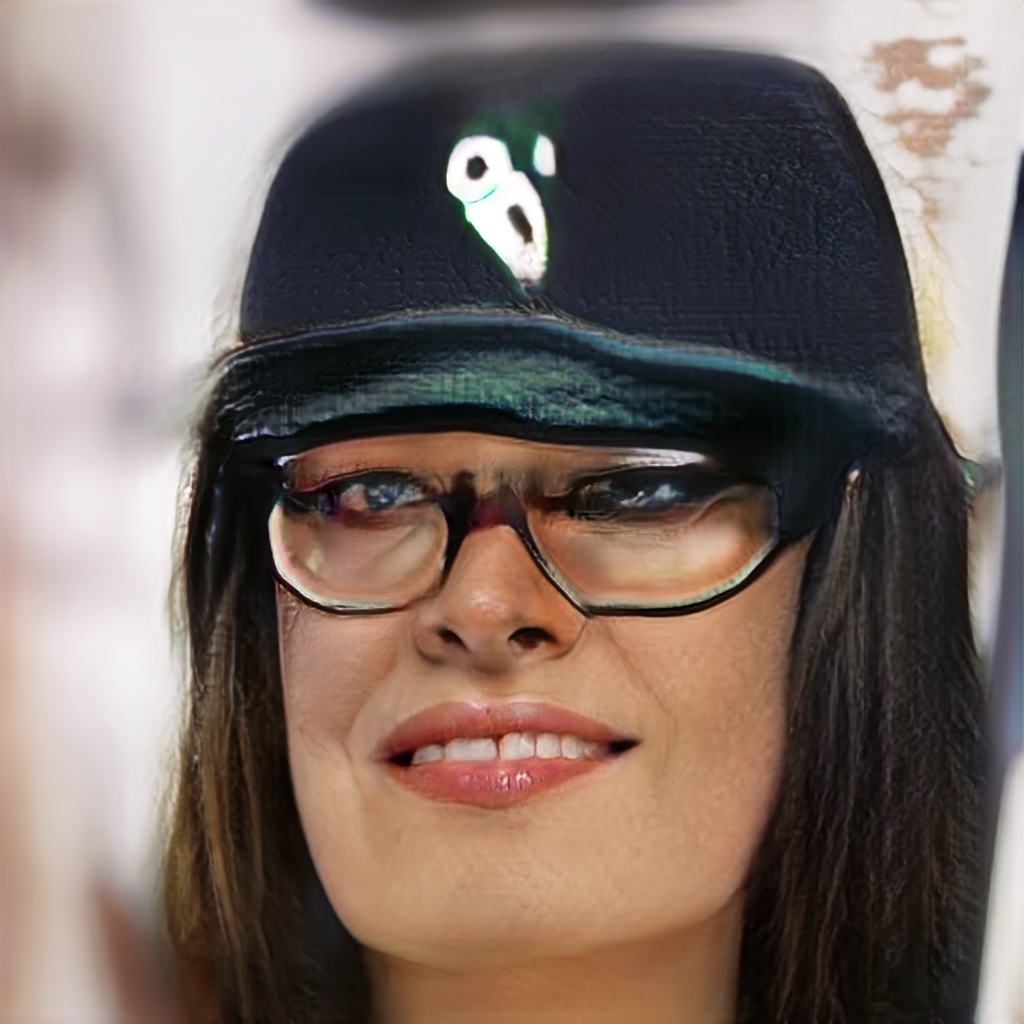} & \includegraphics[width=0.115\linewidth]{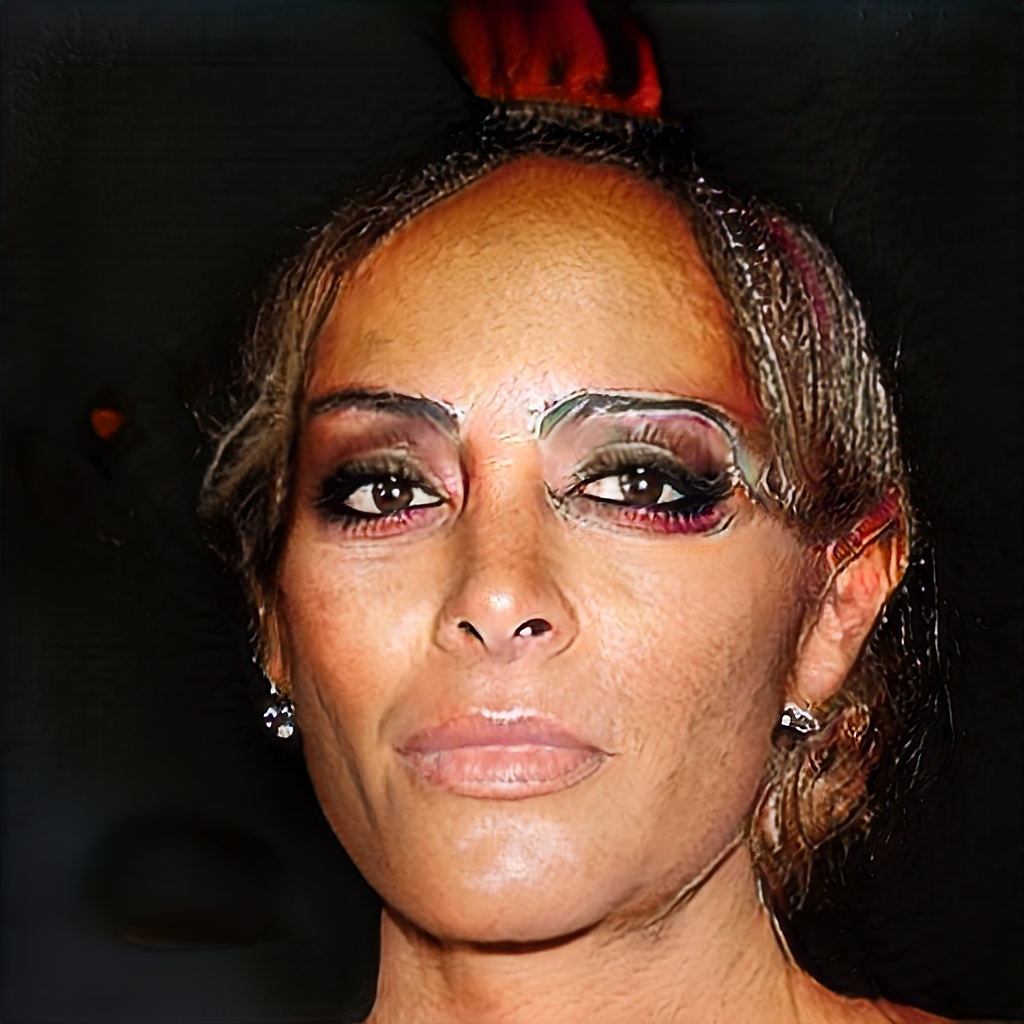} & \includegraphics[width=0.115\linewidth]{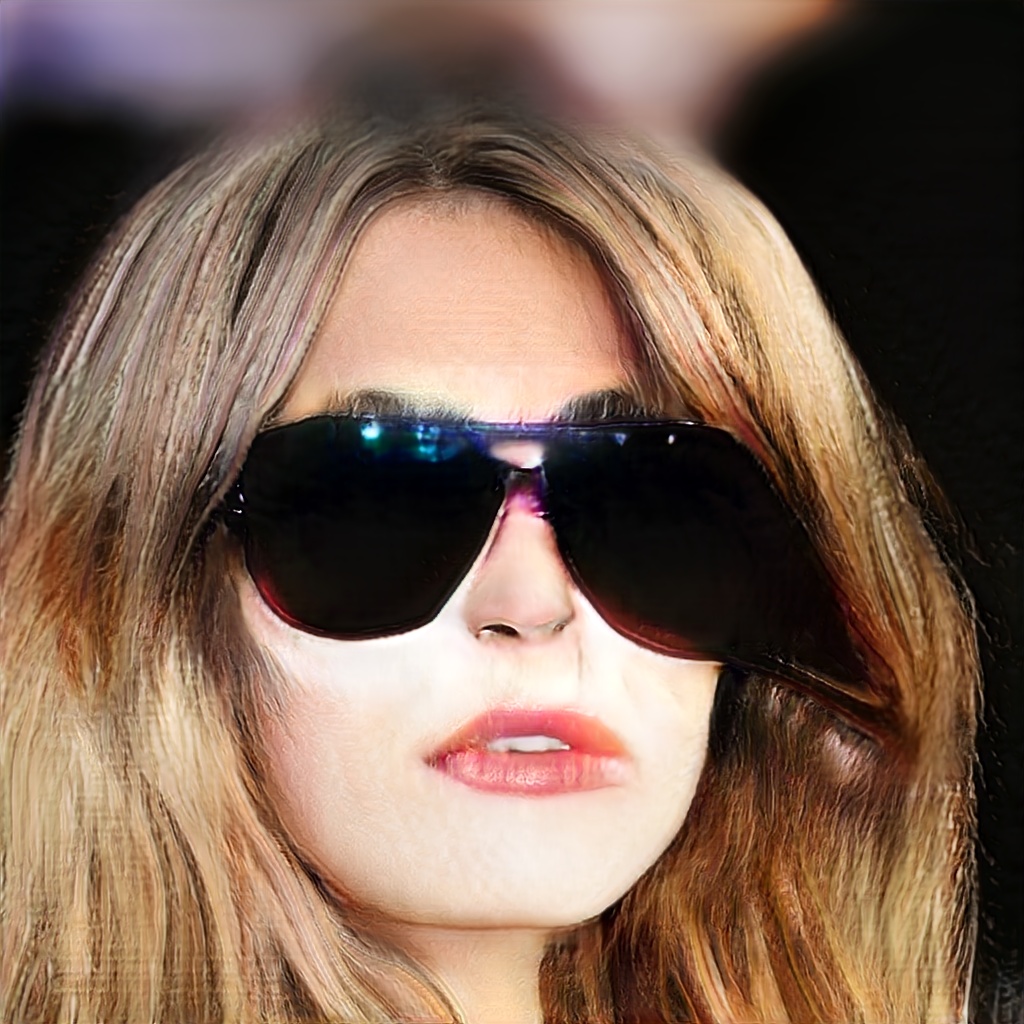} & \includegraphics[width=0.115\linewidth]{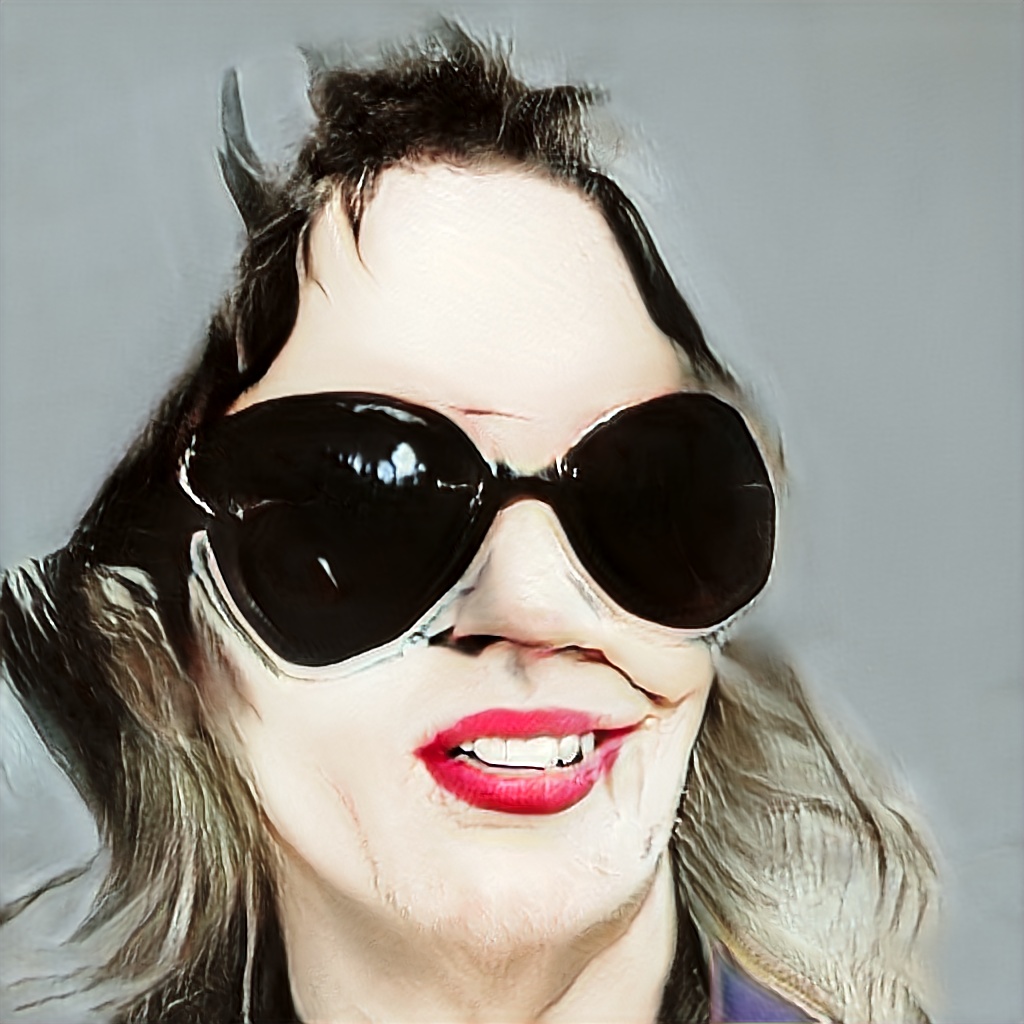} & \includegraphics[width=0.115\linewidth]{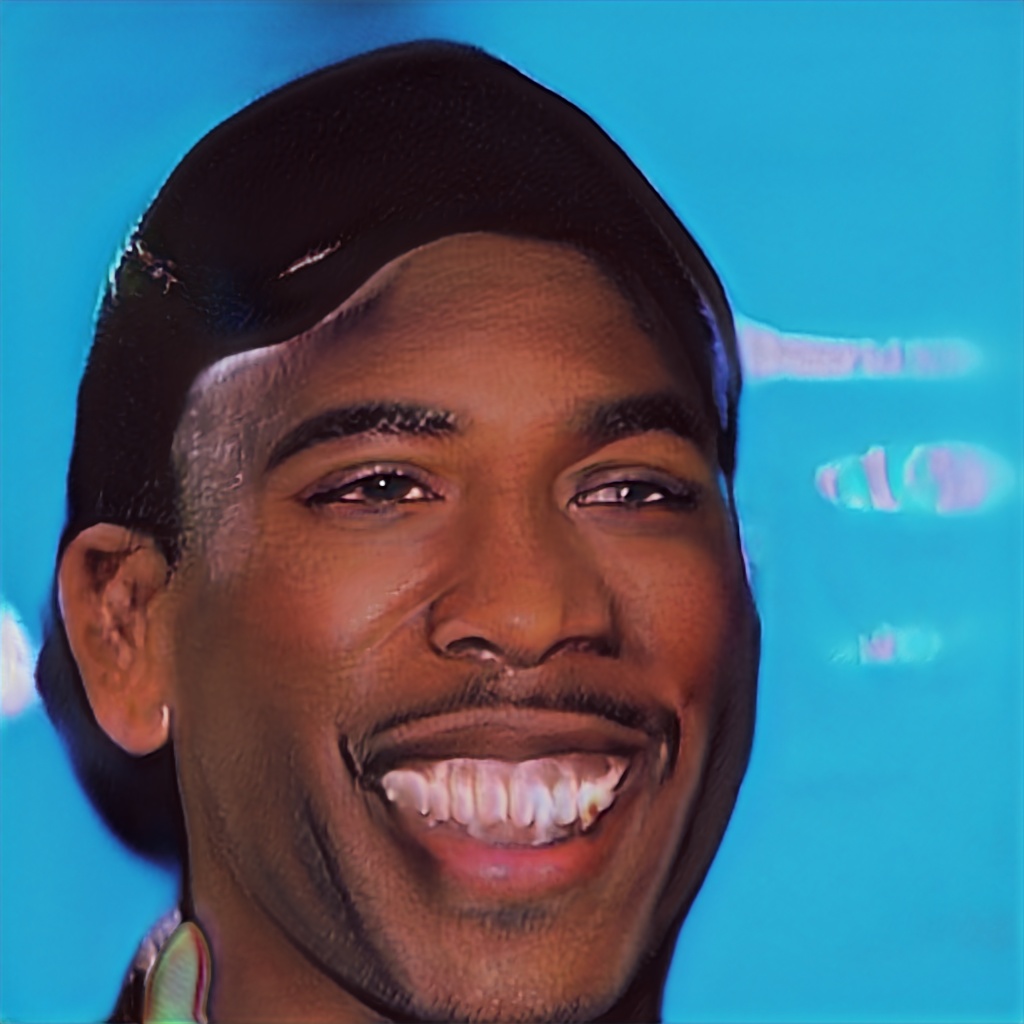} & \includegraphics[width=0.115\linewidth]{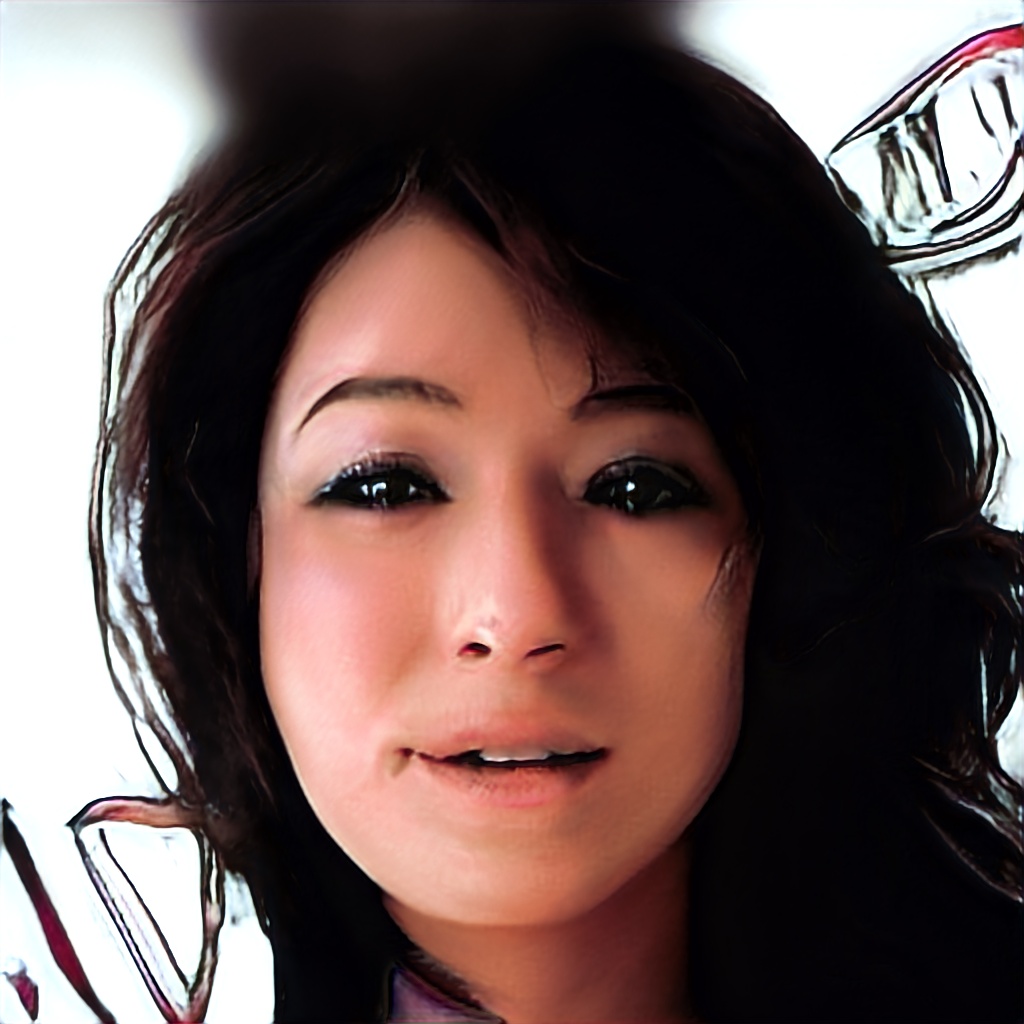} & \includegraphics[width=0.115\linewidth]{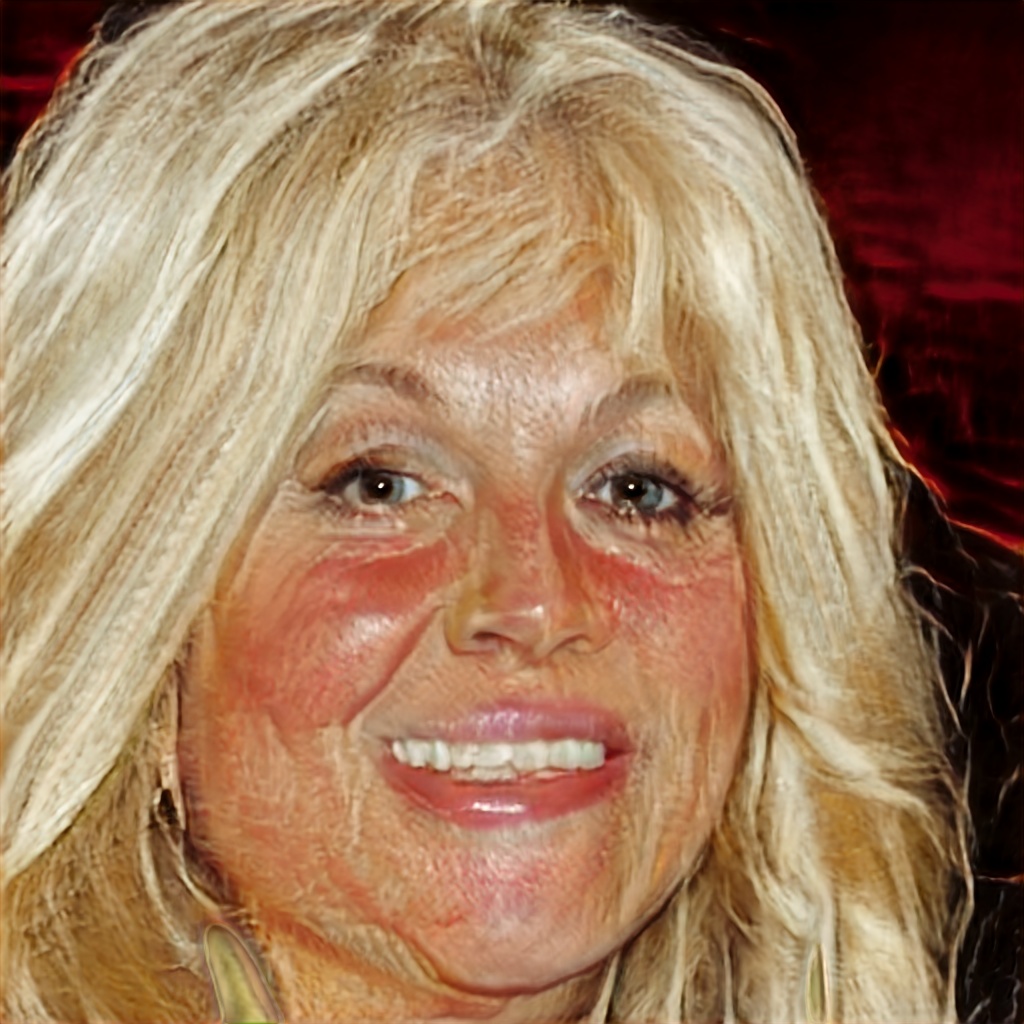}& \rot{InterfaceGAN} \\
\rot{Ours}& \includegraphics[width=0.115\linewidth]{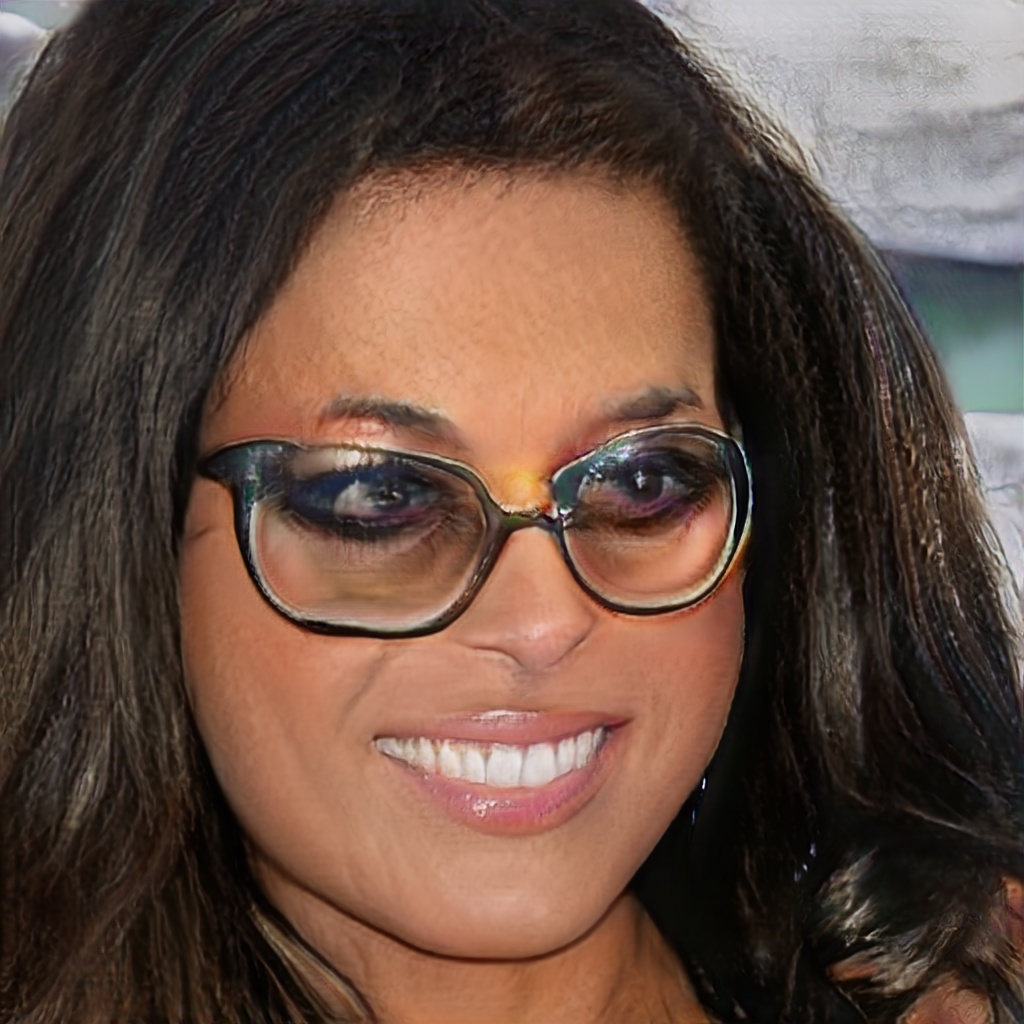} &  \includegraphics[width=0.115\linewidth]{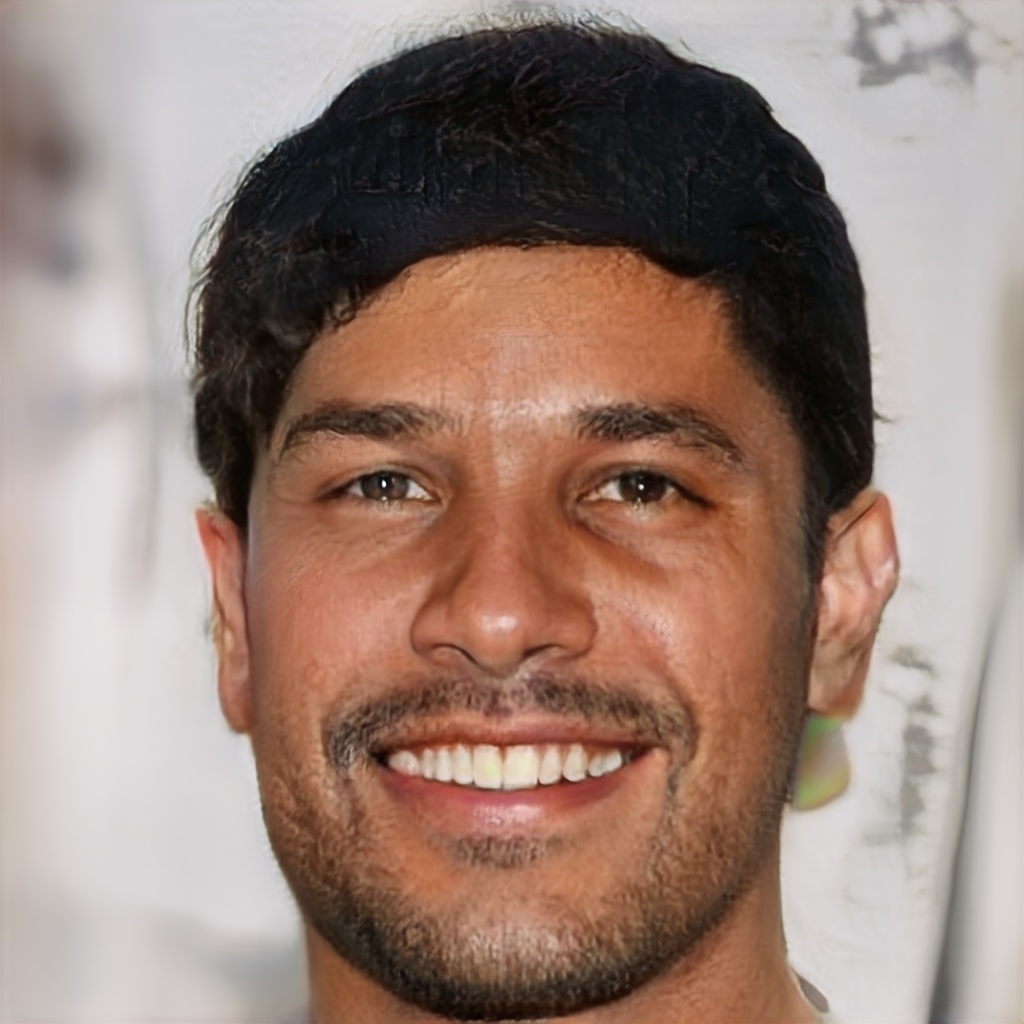} & \includegraphics[width=0.115\linewidth]{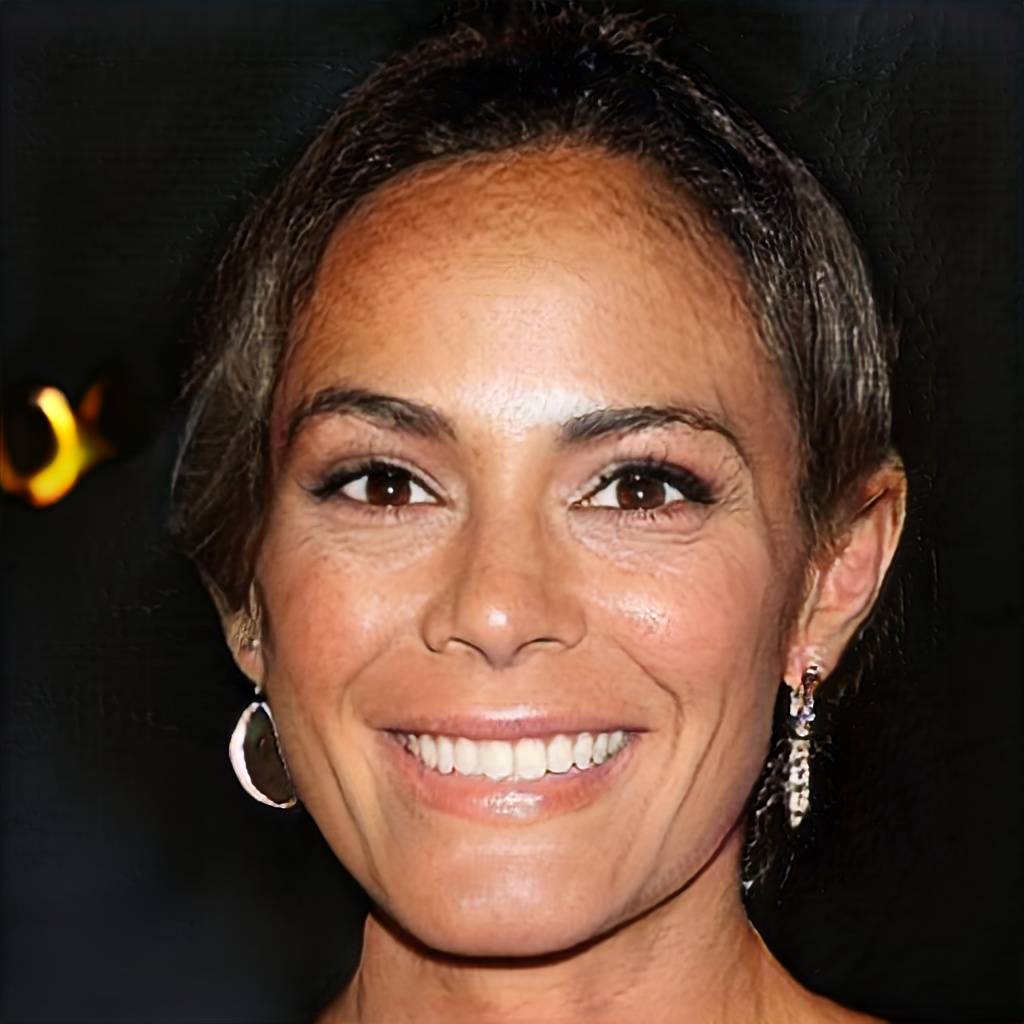} & \includegraphics[width=0.115\linewidth]{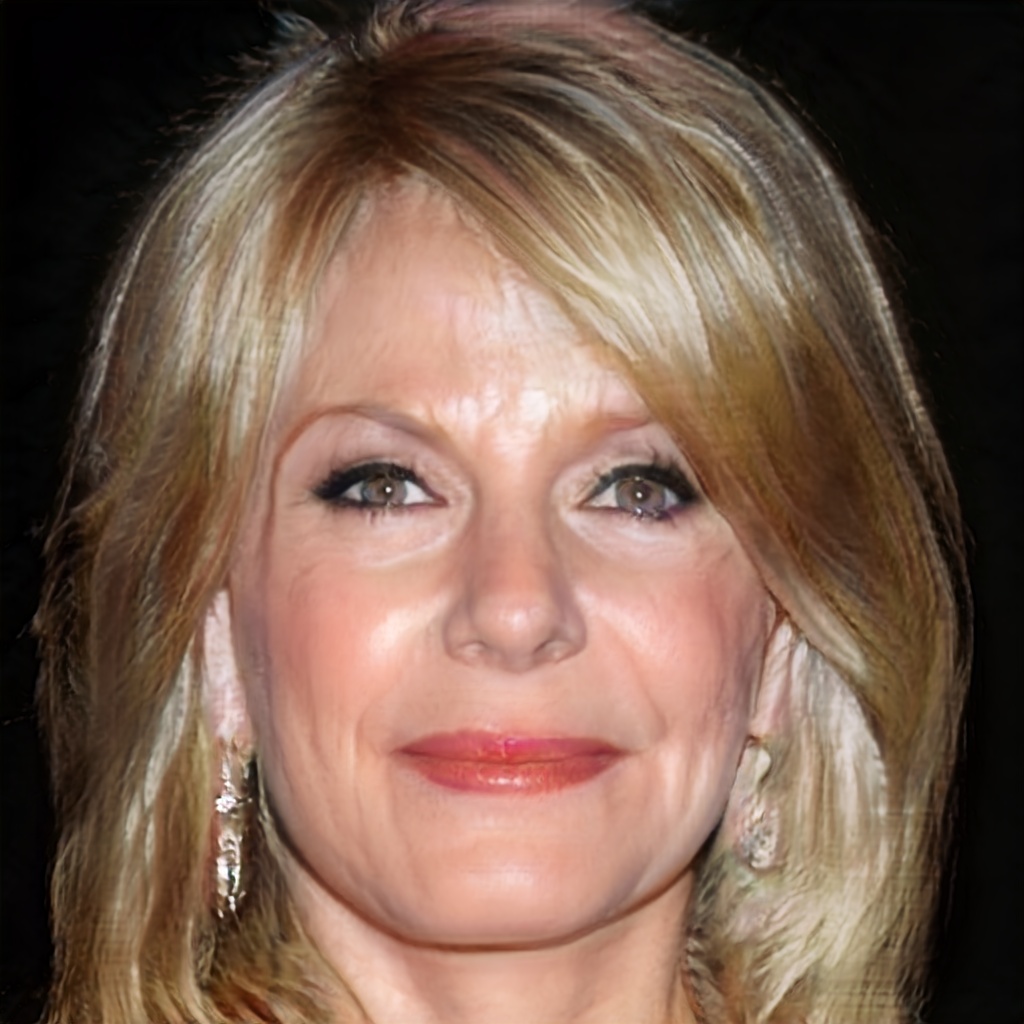} & \includegraphics[width=0.115\linewidth]{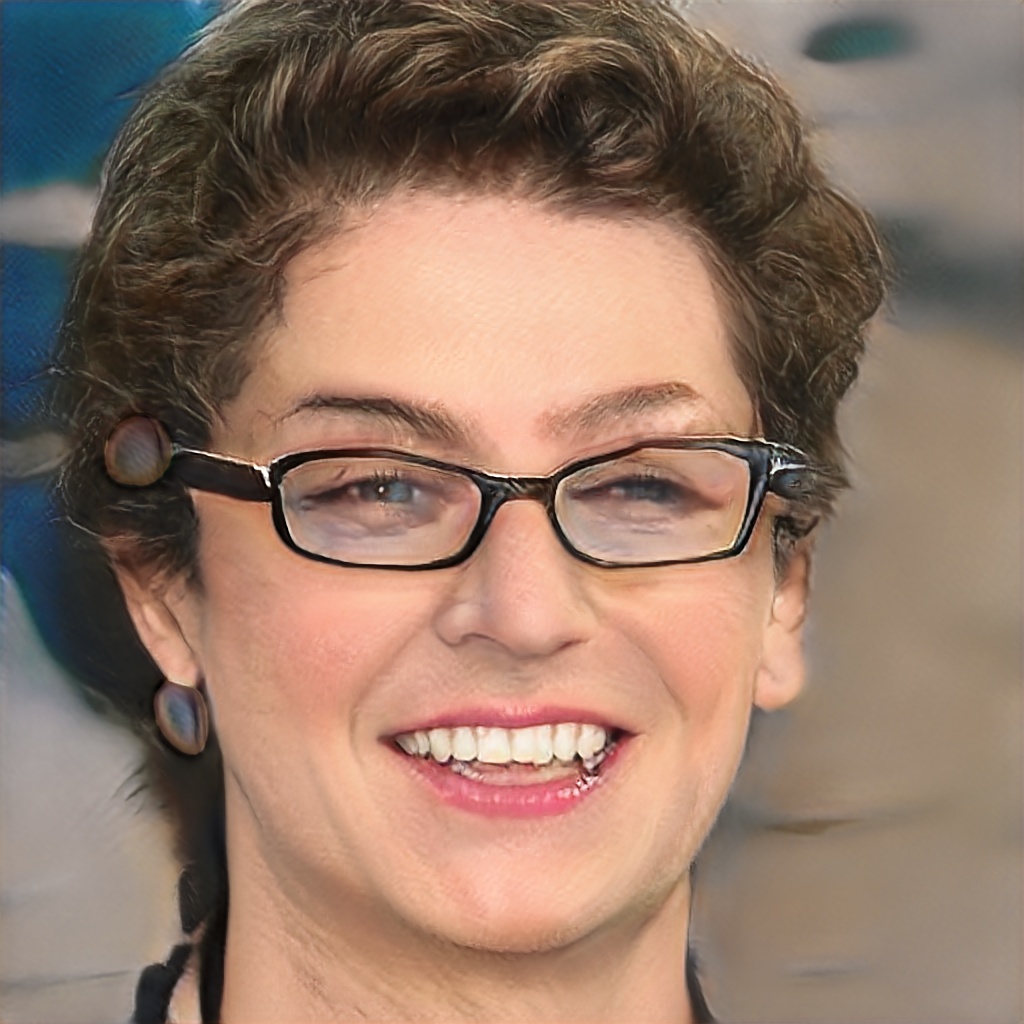} & \includegraphics[width=0.115\linewidth]{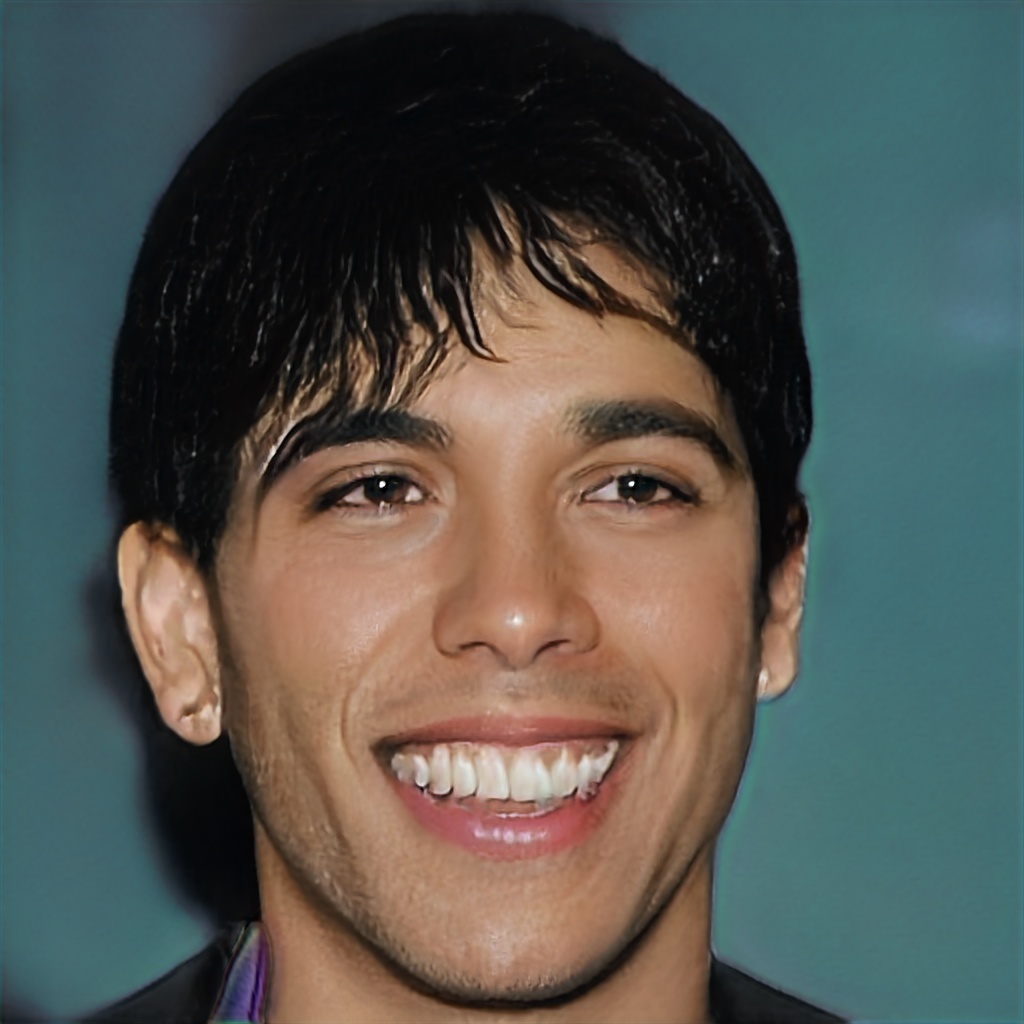} & \includegraphics[width=0.115\linewidth]{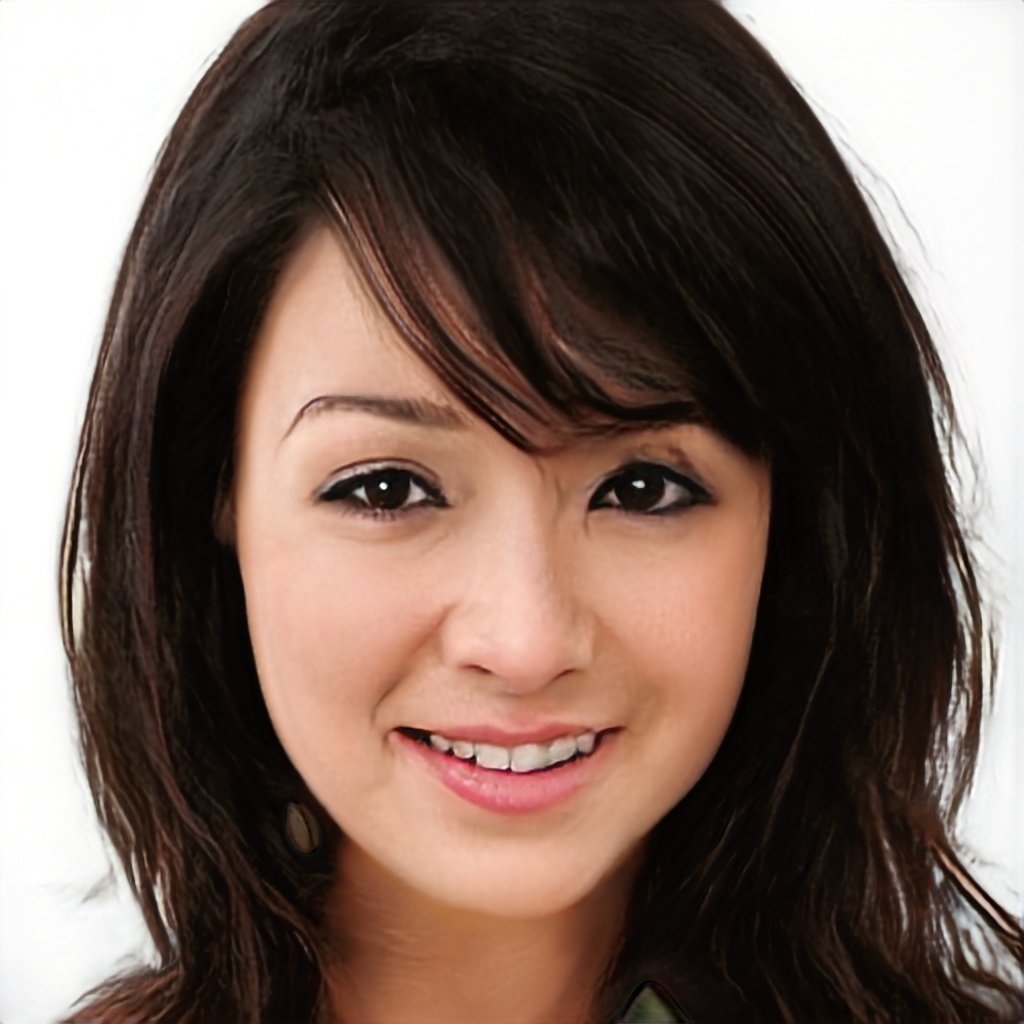} &\includegraphics[width=0.115\linewidth]{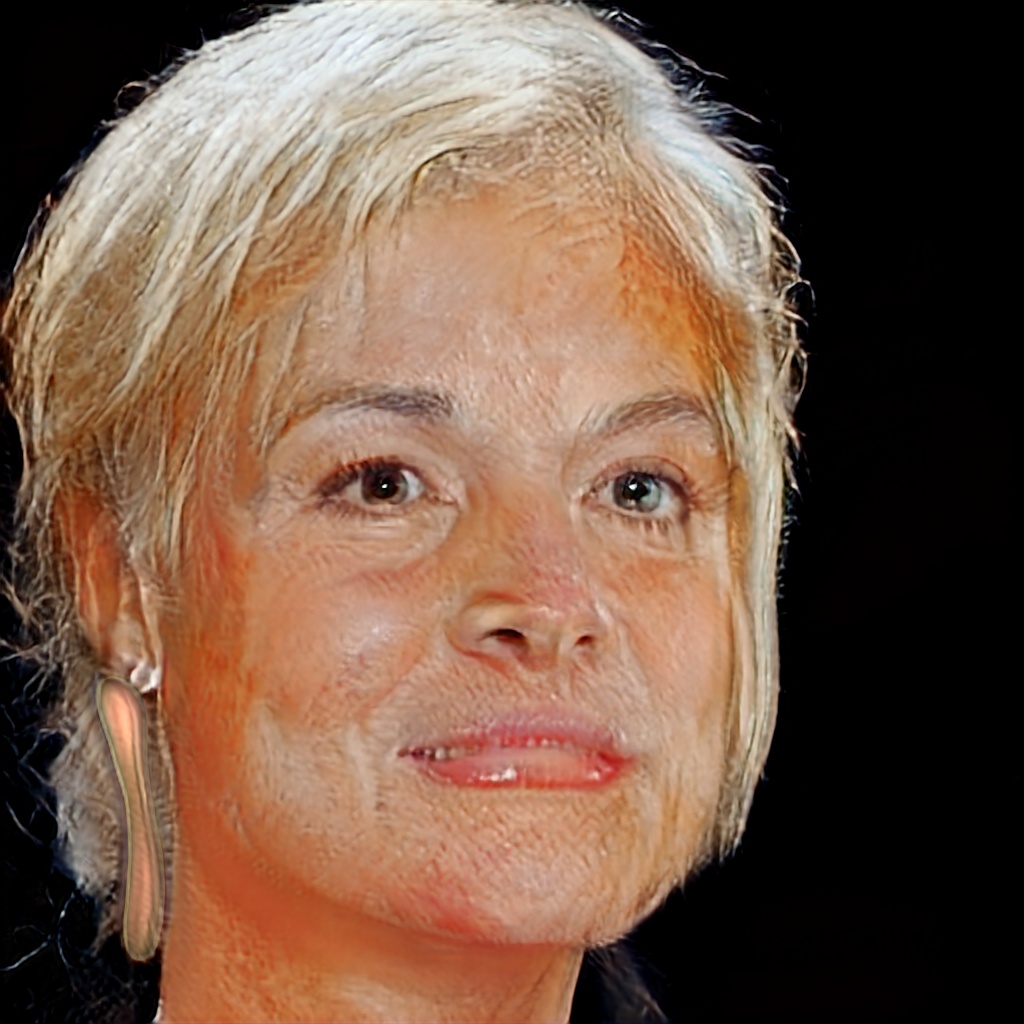}  & \rot{Ours} \\
 & Eyeglasses  & Gender & Smile & Age &  Eyeglasses  & Gender & Smile & Age  & 
\end{tabular}
\caption{\textbf{Conditional} attribute manipulation on PGGAN (left) and StyleGAN (right) with respect to Eyeglasses, Gender, Smile, and Age. We compare our method to InterfaceGAN.}
\label{fig:cond}
\end{figure*}

\vspace{0.1cm}\noindent\textbf{Conditional Manipulation.} We present that our framework along with the proposed constraint in Eq.~\ref{eq:constraint} can preserve the non-target attributes better. In this experiment, we consider 1 of 4 attributes to edit while conditioned on the other three. Figure~\ref{fig:cond} presents the results on PGGAN and StyleGAN, respectively. On PGGAN, our method strongly preserves the non-target attributes and effectively modify the target attribute. InterfaceGAN appears to be ineffective when it has to maintain non-target attributes. For example, the smiling is changed when editing eyeglasses. On StyleGAN, both methods succeed to edit the attributes while our method looks closer to the input. These results suggest that our method with the constraint can effectively edit the target attribute while retaining others.

\begin{figure}[htbp]
    \centering
    \includegraphics[width=\linewidth]{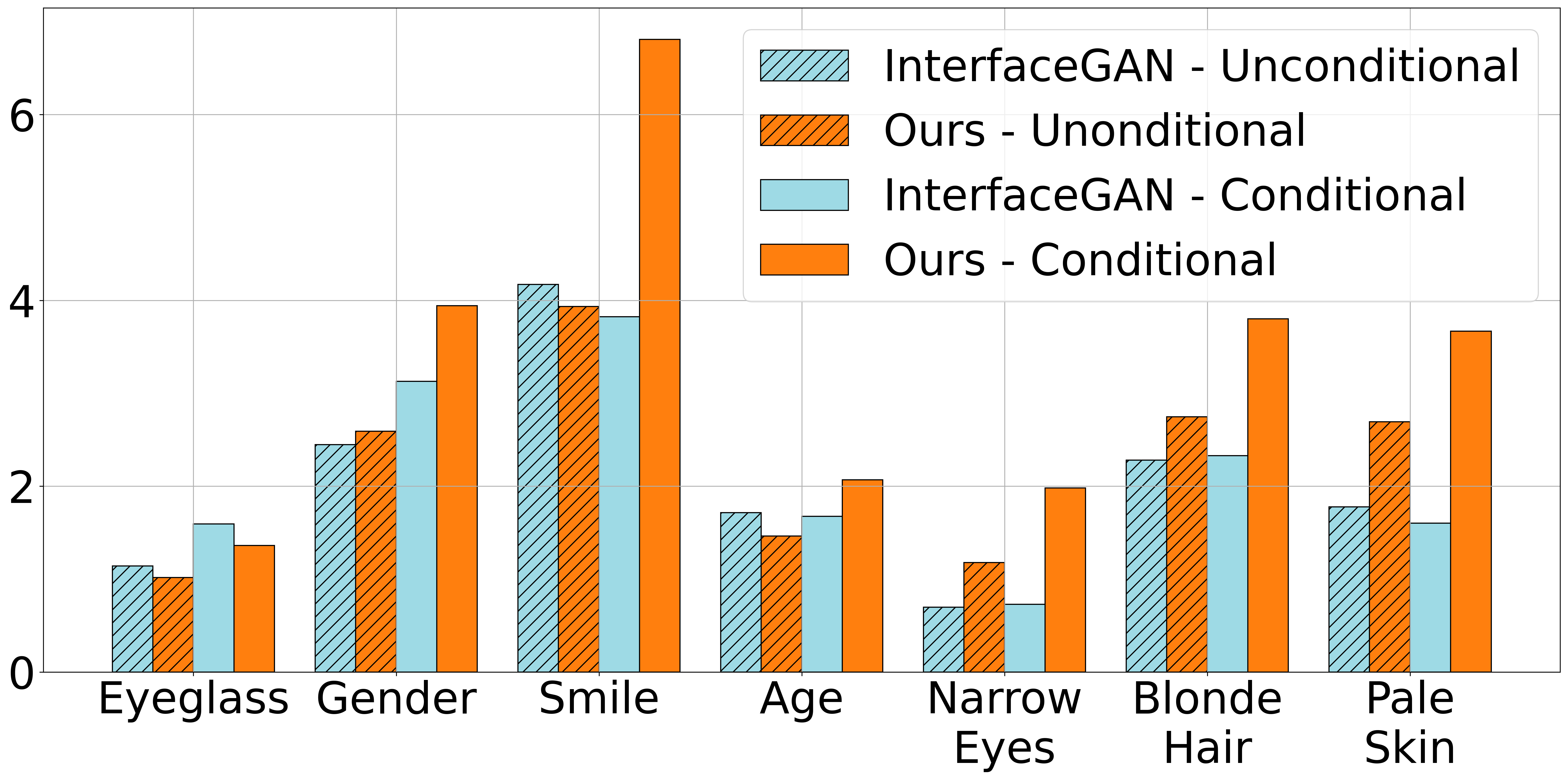}
    \caption{Attribute preservation ratio. We compare our method (orange) to InterfaceGAN (mist blue) in unconditional (non-stripe) and conditional (stripe) settings. The higher values, the better attributes are preserved during attribute manipulation.}
    \label{fig:preservation_ratio}
\end{figure}

\vspace{0.1cm}\noindent\textbf{Attribute Preservation.} To inspect how well our method can preserve the non-target attributes, we follow the same sampling strategy as in Figure~\ref{fig:logit_stylegan_uncond} and measure the ratio of target prediction changes to non-target prediction changes on StyleGAN. Higher values indicate that the subject can more effectively edit target attributes with lower perturbation to non-target attributes. We consider 7 attributes to edit. The former four attributes are conditioned on each other except themselves, and the latter three attributes are conditioned on all the former four. This design makes the latter three more challenging since they have to edit rare attributes while maintaining 4 attributes simultaneously. 

Figure~\ref{fig:preservation_ratio} compares our method to InterfaceGAN in both conditional (Eq.~\ref{eq:constraint}) and unconditional (Eq.~\ref{eq:iterative_traversal}) settings. It is immediately observed that our method with the proposed constraint achieves the highest ratio on almost all attributes, validating that the non-linear iterative scheme not only advances the attribute manipulation but also help the attribute preservation. Interestingly, the ratio of our unconditional method on the last three attributes are higher than InterfaceGAN in the conditional setting. It could be attributed to inefficiency of linear methods. As observed in Figure~\ref{fig:logit_stylegan_uncond}, linear methods are less activate on rare attributes. Moreover, it uses the same direction for every step, which may be sub-optimal for the points far away from the initial point. Both two reasons together lead to the phenomenon. 

\subsection{Analysis}
\label{ssec:analysis}
We conduct two analysis to provide more insights from the perspective of smoothness and function approximation.

\begin{table}[htbp]
\centering
\setlength\tabcolsep{0.25em}
\begin{tabular}{@{}ccx{1cm}x{1cm}x{1cm}x{1cm}@{}}
\toprule
 & \multicolumn{1}{c}{Cond.} & \multicolumn{1}{c}{Eyeglass} & \multicolumn{1}{c}{Gender} & \multicolumn{1}{c}{Smile} & \multicolumn{1}{c}{Age} \tabularnewline \midrule
\multicolumn{1}{l}{} & \multicolumn{5}{c}{PGGAN} \tabularnewline \cmidrule(l){2-6} 
InterfaceGAN & \multirow{2}{*}{N} &  \textbf{60.69} & 65.00 &  \textbf{54.49} & 61.16 \tabularnewline
Ours &  & 64.10 &  \textbf{62.50} & 56.55 &  \textbf{60.28} \tabularnewline \cmidrule(l){2-6} 
InterfacGAN & \multirow{2}{*}{Y} &  \textbf{54.51} &  \textbf{55.65} & 55.07 & 56.59 \tabularnewline
Ours &  & 57.40 & 59.79 &  \textbf{55.00} &  \textbf{55.30} \tabularnewline \midrule
\multicolumn{1}{l}{} & \multicolumn{5}{c}{StyleGAN} \\ \cmidrule(l){2-6} 
InterfaceGAN & \multirow{2}{*}{N} &  \textbf{99.15} & 101.65 &  \textbf{96.90} & 96.62 \tabularnewline
Ours &  & 103.08 &  \textbf{97.93} & 97.93 &  \textbf{91.86} \tabularnewline \cmidrule(l){2-6} 
InterfaceGAN & \multirow{2}{*}{Y} & 80.46 & 81.52 & 92.75 & 83.51 \tabularnewline
Ours &  &  \textbf{57.23} &  \textbf{71.67} &  \textbf{78.18} &  \textbf{55.41} \tabularnewline \bottomrule
\end{tabular}
\caption{Modified Perceptual Path Length measured on PGGAN and StyleGAN. Lower is better.}
\label{tab:ppl}
\end{table}

\vspace{0.1cm}\noindent\textbf{Smoothness.} We compare the trajectories generated by our method to those by linear methods such as InterfaceGAN in terms of smoothness. Previously, Karras~\etal~\cite{karras2019style} propose a metric, Perceptual Path Length (PPL), to measure smoothness over the whole latent space. However, we only focus on smoothness of trajectories; thus we propose to use modified PPL (mPPL) that only consider noise sampled from the generated trajectories. Formally, we have

\begin{equation}
\label{eq:modified_ppl}
\begin{aligned}
    l_z = \mathds{E}_{z^{(i)} \sim \mathcal{T}, \mathcal{T} \sim p(\mathcal{T})}[\frac{1}{\epsilon^2}d(& G(\text{lerp}(z^{(i)}, z^{(i+1)}, t)), \\ &G(\text{lerp}(z^{(i)}, z^{(i+1)}, t+\epsilon)))],
\end{aligned}
\end{equation}
where $\mathcal{T}$ is a generated trajectory, $z^{(i)}$ is the noise of i-th step in the trajectory, $G$ is the generator, $\text{lerp}$ is linear interpolation function with factor $t$ drawn from uniform distribution, $\epsilon$ is a small displacement, and $d(\cdot)$ is LPIPS metric~\cite{zhang2018unreasonable}. For each attribute, we sample 1000 trajectories and each consists of 600 steps with step size 0.01, leading to 600k images in total. Following Karras~\etal~\cite{karras2019style}, we set $\epsilon$ to 1e-4 and report the mean value over all samples. Lower values mean smoother trajectories.

Table~\ref{tab:ppl} presents the evaluation results on both PGGAN and StyleGAN. In the unconditional setting, both methods produce comparably smooth trajectories. It also happens in the condition setting on PGGAN. It could be attributed to the relatively linear manifolds. Note that smoothness does not mean the effectiveness. One could generate images with no change to gain smoother transition. Lastly, our method surpasses the baseline on StyleGAN in the conditional setting by a large margin since it can strongly retain the non-target attributes, leading to smoother transition.

\begin{table}[htbp]
\centering
\begin{tabular}{@{}lrrrrr@{}}
\toprule
 & \multicolumn{1}{c}{$<1$} & \multicolumn{1}{c}{$<2$} & \multicolumn{1}{c}{$<3$} & \multicolumn{1}{c}{$>=3$} & \multicolumn{1}{c}{Avg.} \\ \midrule
 & \multicolumn{5}{c}{Eyeglasses} \\ \cmidrule(l){2-6} 
InterfaceGAN & 1.760 & 2.779 & 3.644 & \textbf{1.481} & 2.416 \\
Ours & \textbf{1.675} & \textbf{2.401} & \textbf{2.469} & 1.557 & \textbf{2.026} \\ \midrule
 & \multicolumn{5}{c}{Gender} \\ \cmidrule(l){2-6} 
InterfaceGAN & 5.702 & 4.045 & 1.798 & 0.891 & 3.109 \\
Ours & \textbf{4.469} & \textbf{3.694} & \textbf{1.790} & \textbf{0.812} & \textbf{2.692} \\ \midrule
 & \multicolumn{5}{c}{Smile} \\ \cmidrule(l){2-6} 
InterfaceGAN & \textbf{1.764} & \textbf{1.783} & 1.693 & 0.961 & \textbf{1.550} \\
Ours & 3.191 & 2.391 & \textbf{1.611} & \textbf{0.921} & 2.028 \\ \midrule
 & \multicolumn{5}{c}{Age} \\ \cmidrule(l){2-6} 
InterfaceGAN & 2.350 & 2.434 & 2.312 & 1.354 & 2.113 \\
Ours & \textbf{0.969} & \textbf{1.109} & \textbf{1.893} & \textbf{1.285} & \textbf{1.314} \\ \bottomrule
\end{tabular}
\caption{Errors of first order Taylor approximation. Errors are reported according to the L2-norm between the initial point and the estimated points. $f$ consists of the generator and the task model.}
\label{tab:taylor}
\end{table}
\begin{figure*}[tb]
    \centering
    \setlength\tabcolsep{0.25em}
    \begin{tabular}{cccccc}
    \includegraphics[width=0.15\linewidth]{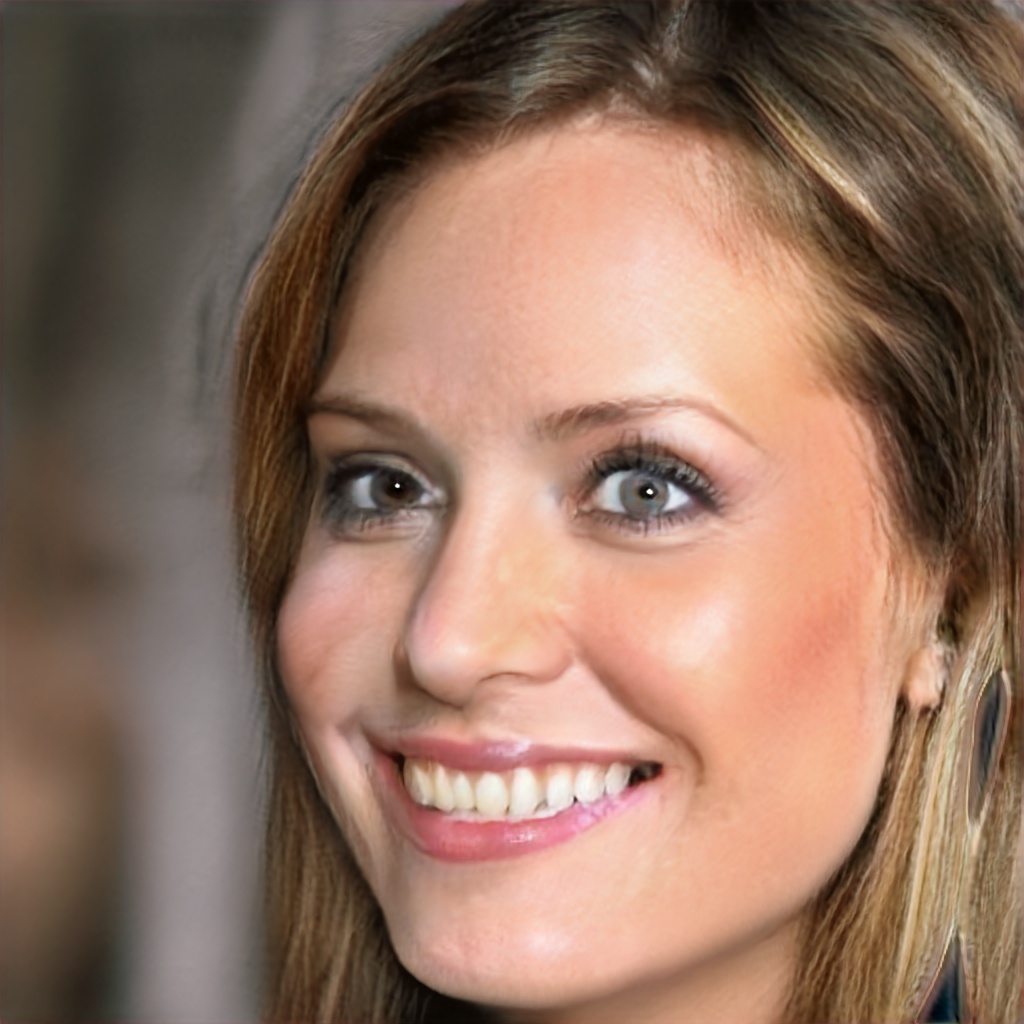} &
     \includegraphics[width=0.15\linewidth]{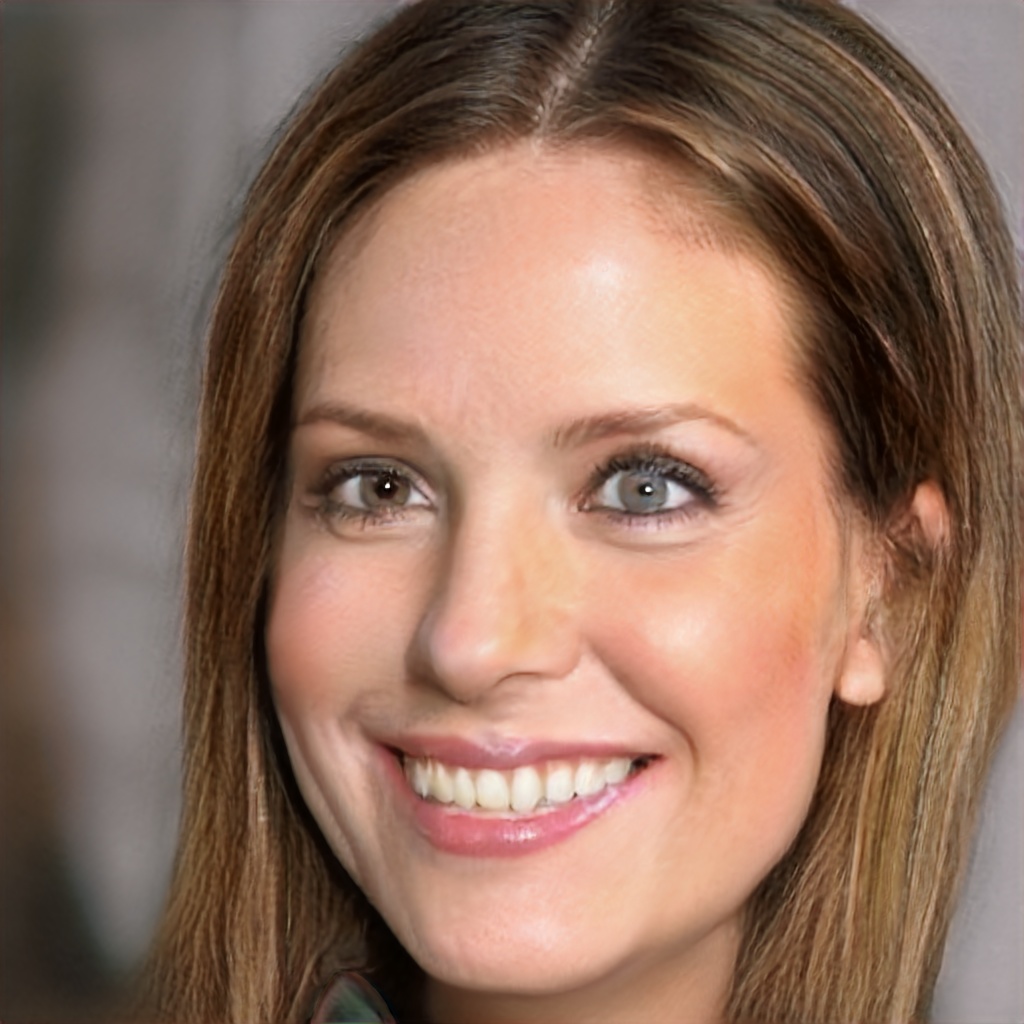}&  \includegraphics[width=0.15\linewidth]{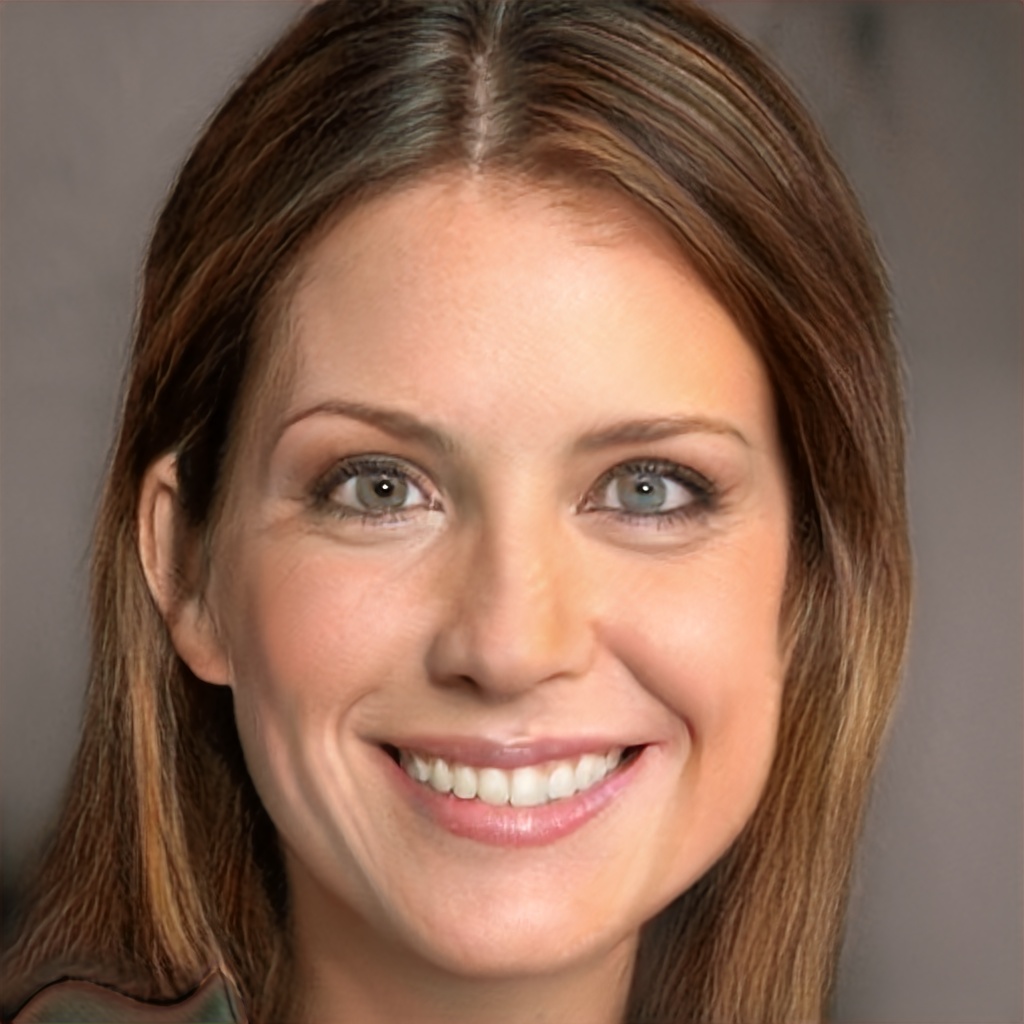} & \includegraphics[width=0.15\linewidth]{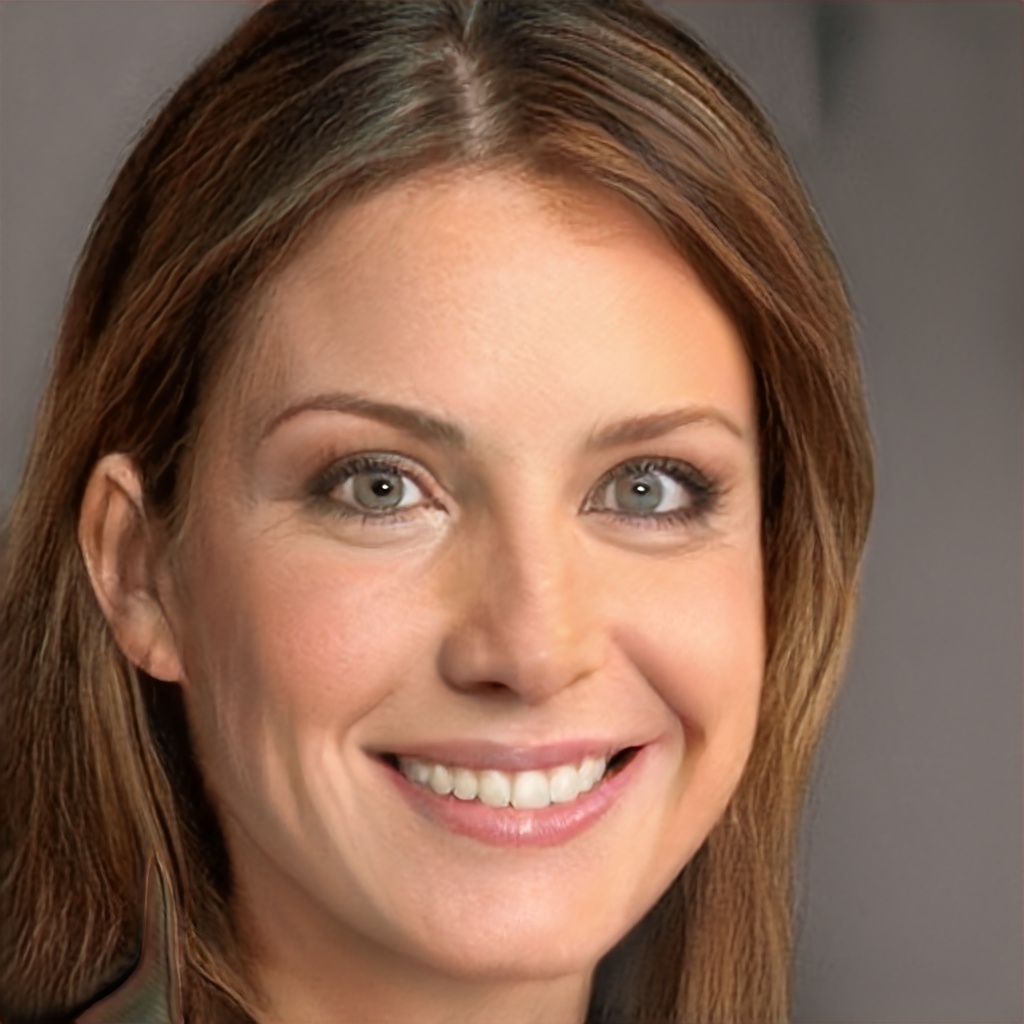} & \includegraphics[width=0.15\linewidth]{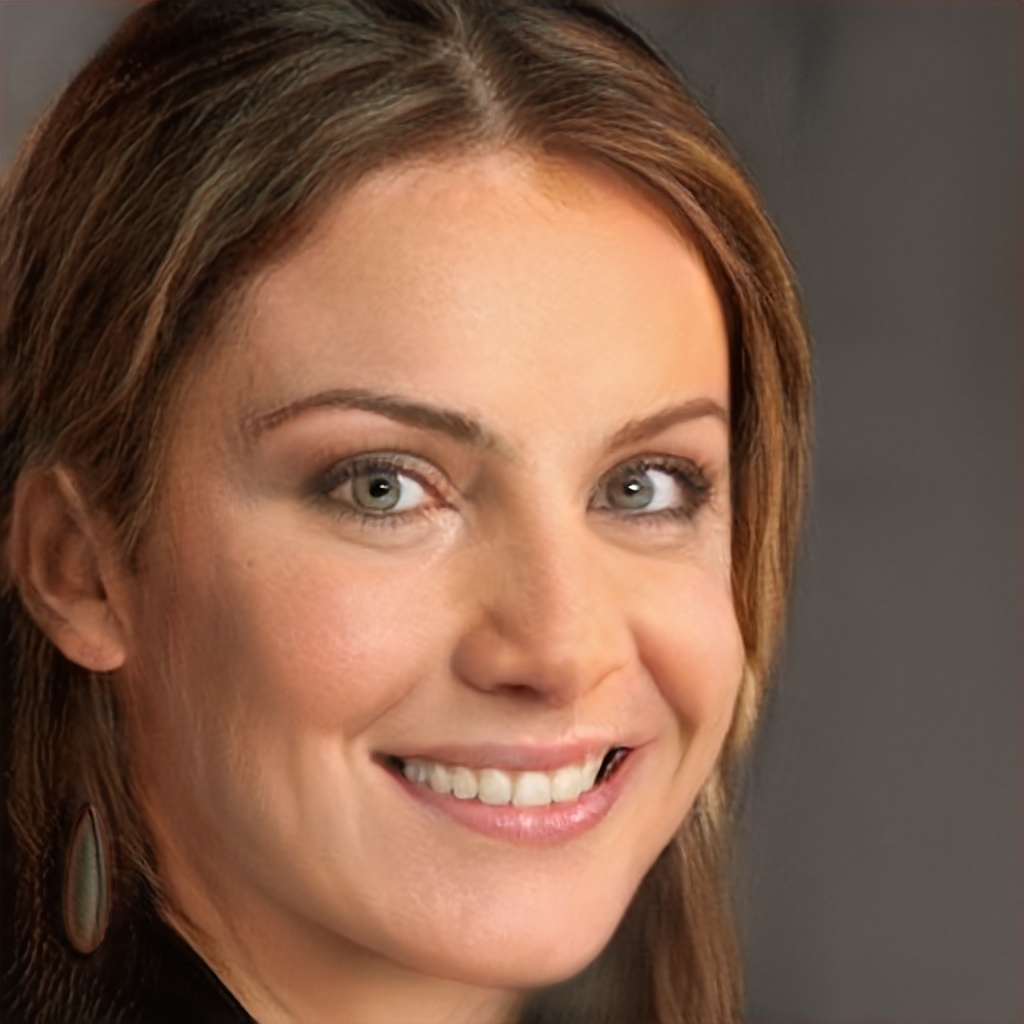} & \includegraphics[width=0.15\linewidth]{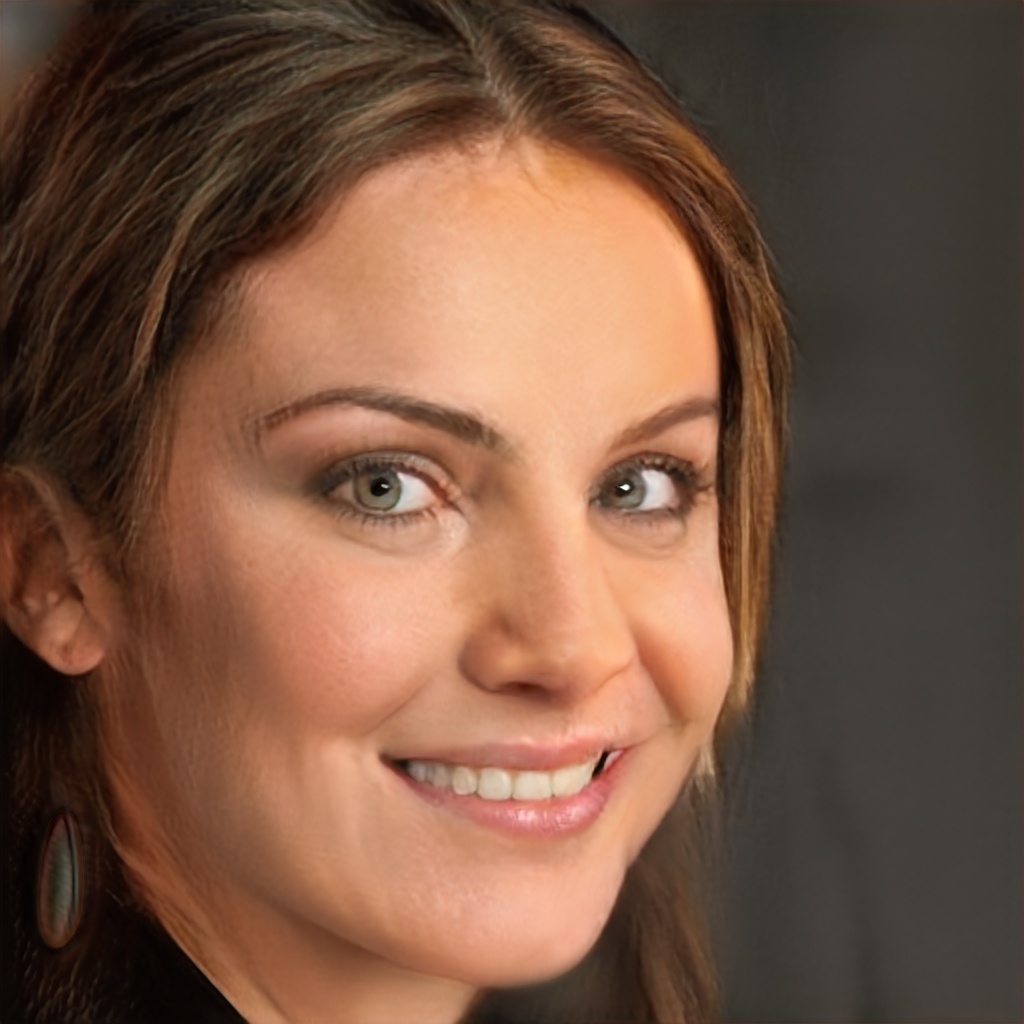} \\
     \includegraphics[width=0.15\linewidth]{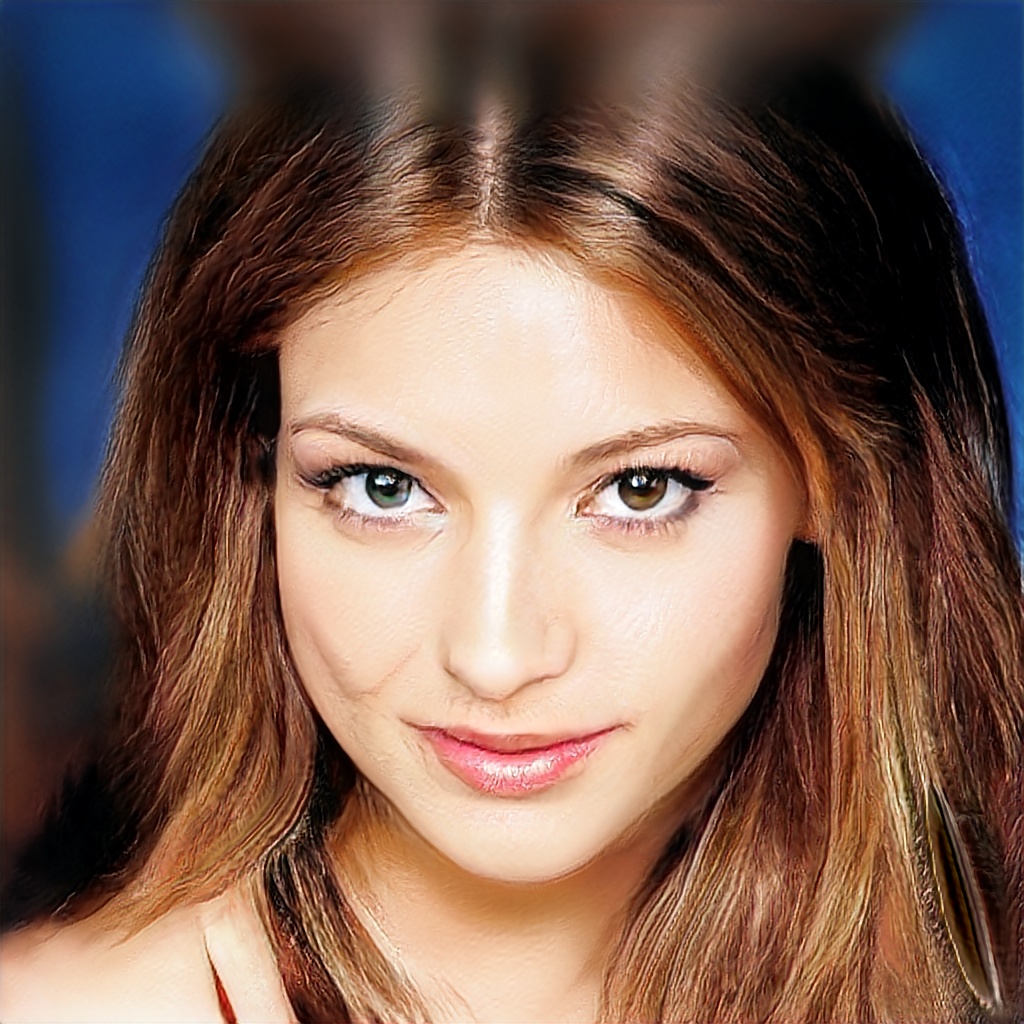} &
     \includegraphics[width=0.15\linewidth]{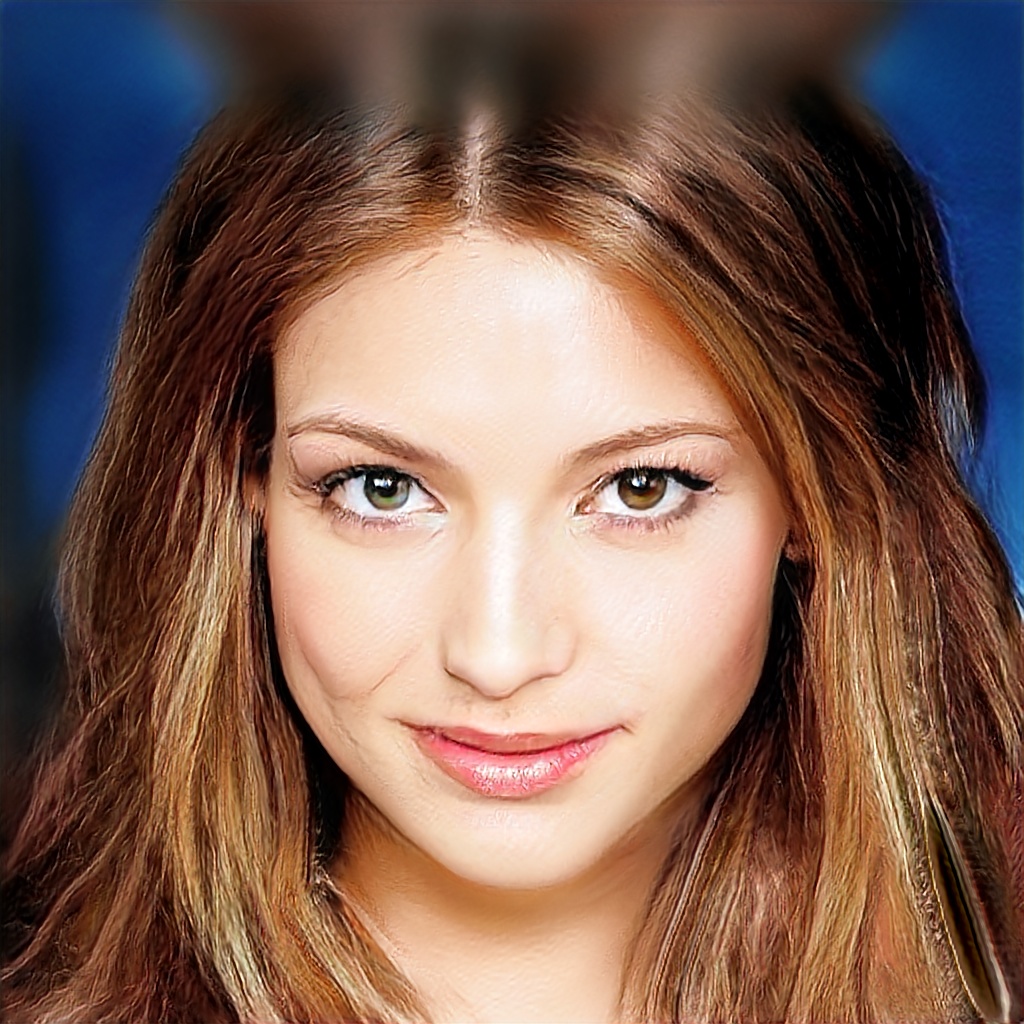}&  \includegraphics[width=0.15\linewidth]{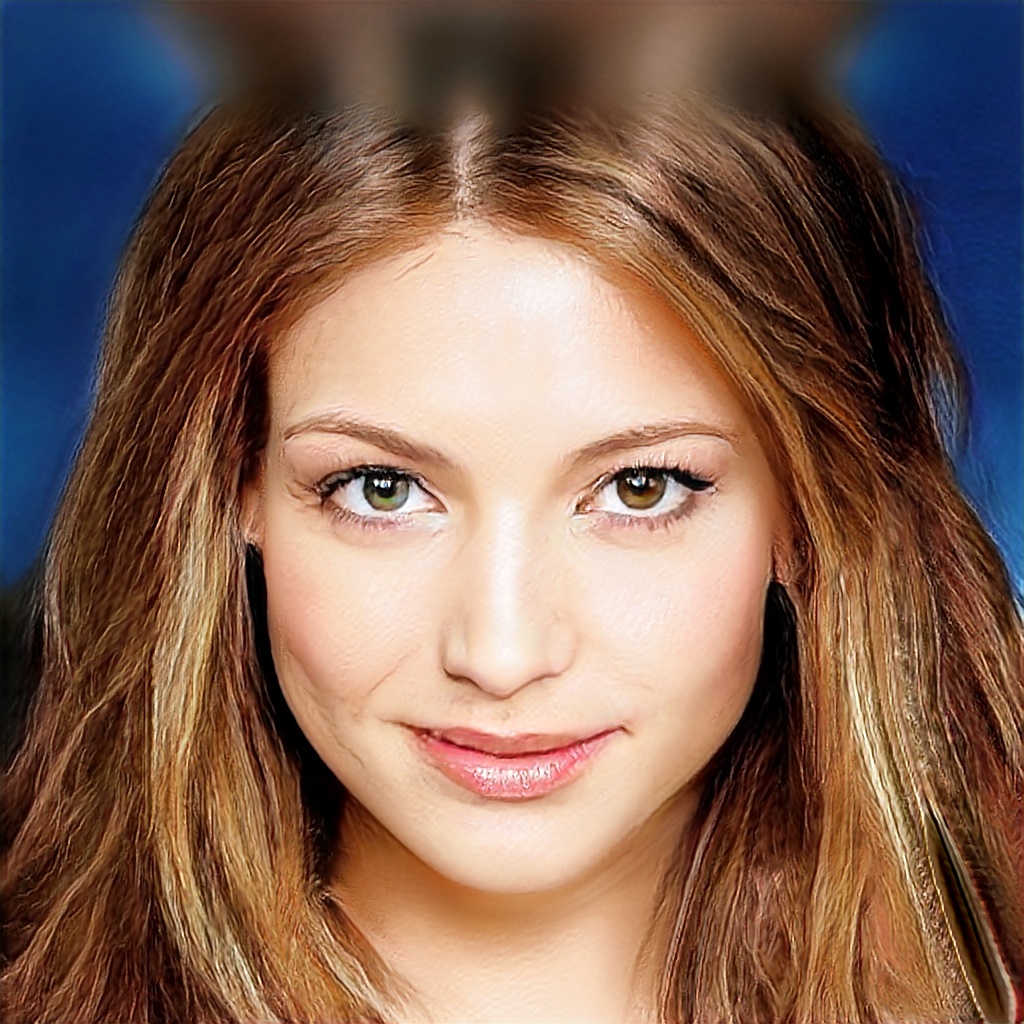} & \includegraphics[width=0.15\linewidth]{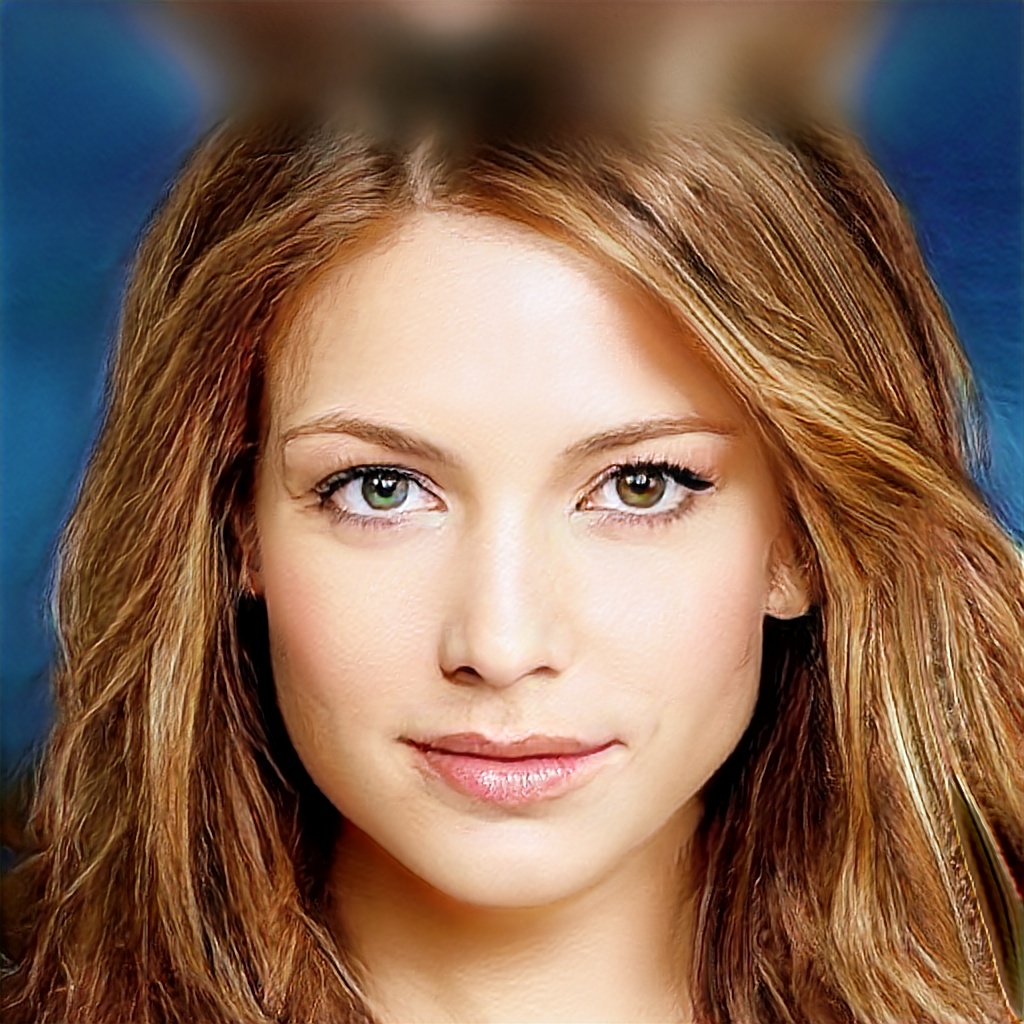} & \includegraphics[width=0.15\linewidth]{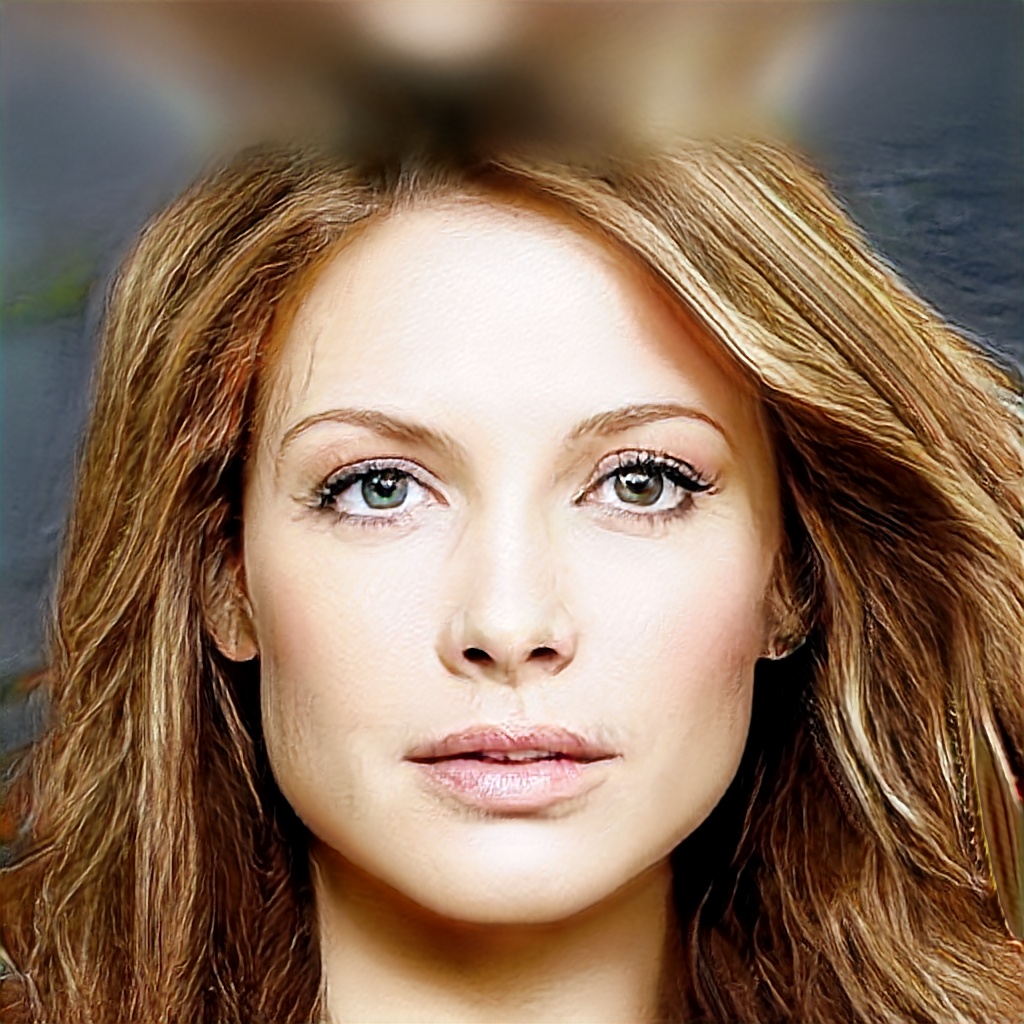} & \includegraphics[width=0.15\linewidth]{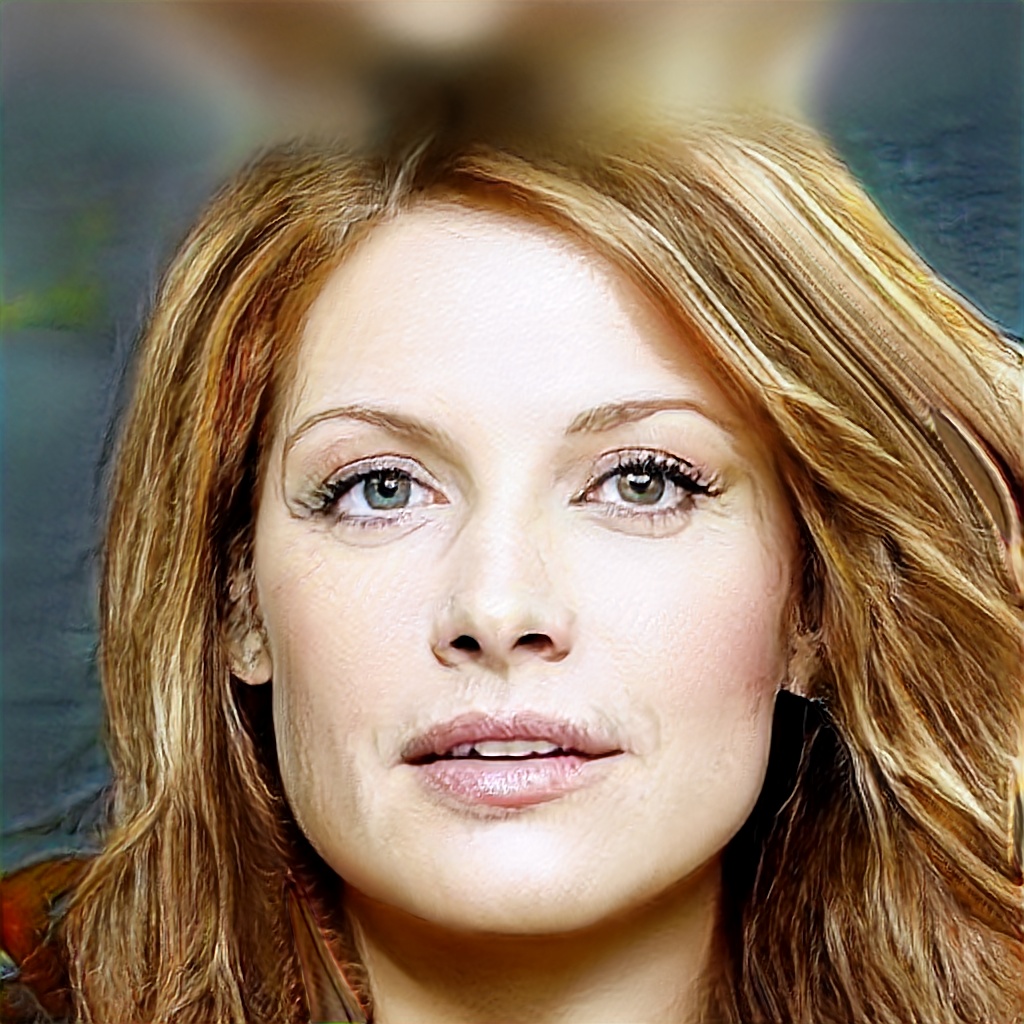}
    \end{tabular}
    \caption{Pose manipulation on StyleGAN with respect to yaw (top) and pitch (bottom).}
    \label{fig:pose}
\end{figure*}
\begin{figure*}[tb]
\setlength\tabcolsep{0.25em}
\centering
\begin{tabular}{ccccccc}
 \includegraphics[width=0.15\linewidth]{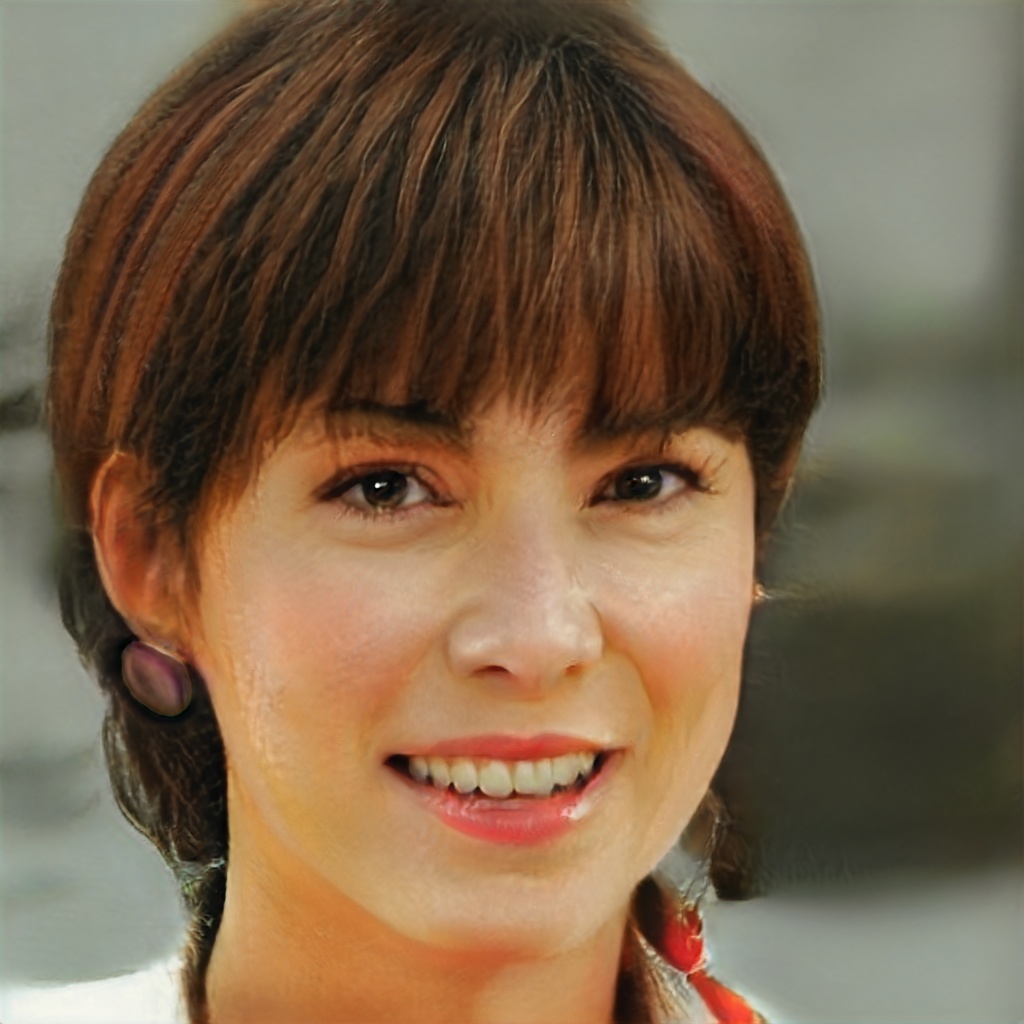}& \includegraphics[width=0.15\linewidth]{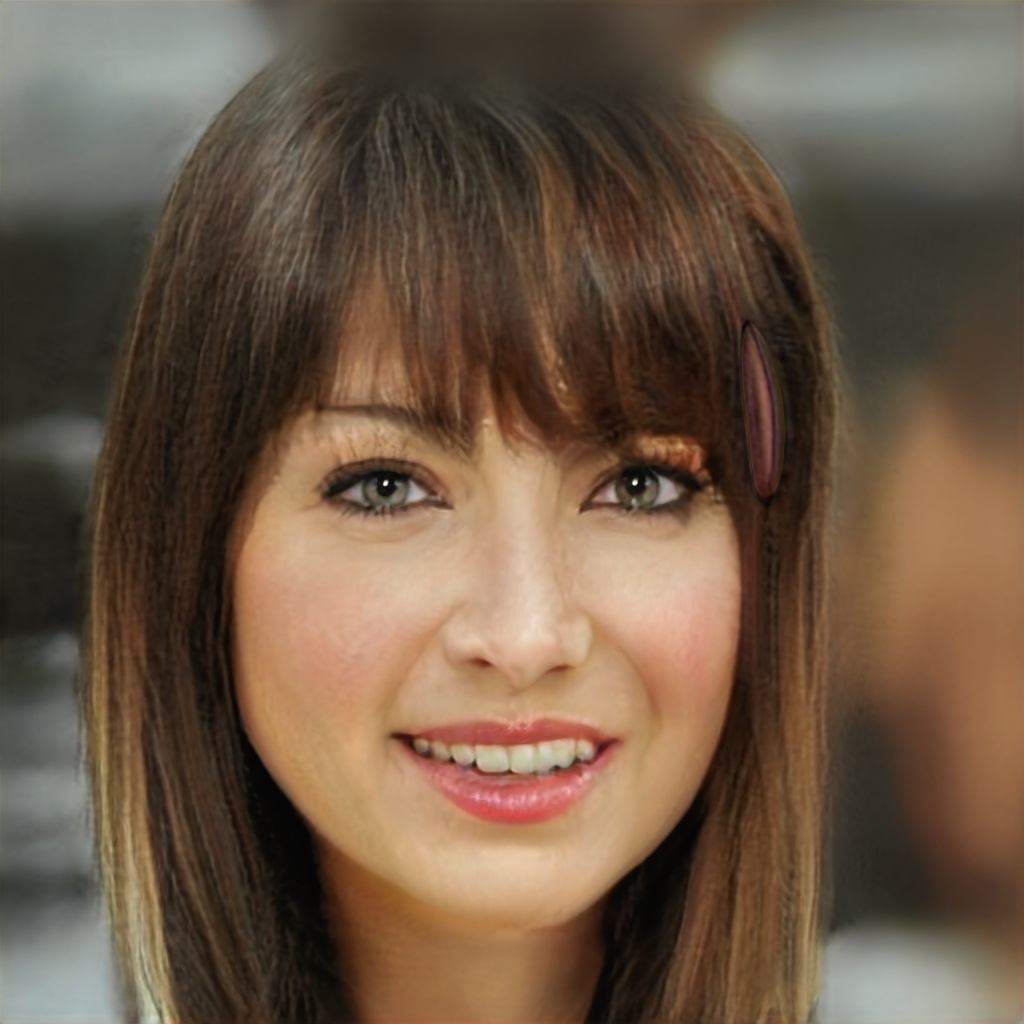} & \includegraphics[width=0.15\linewidth]{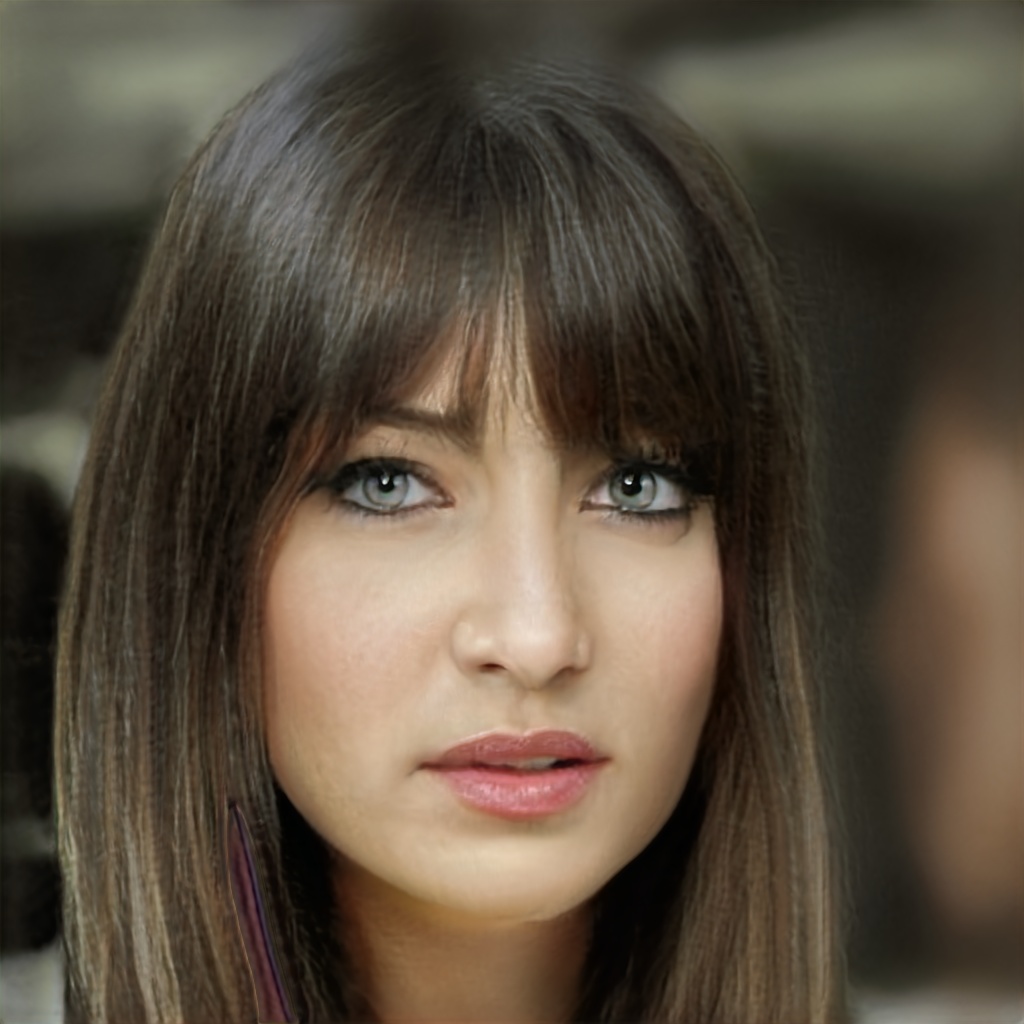} &  & \includegraphics[width=0.15\linewidth]{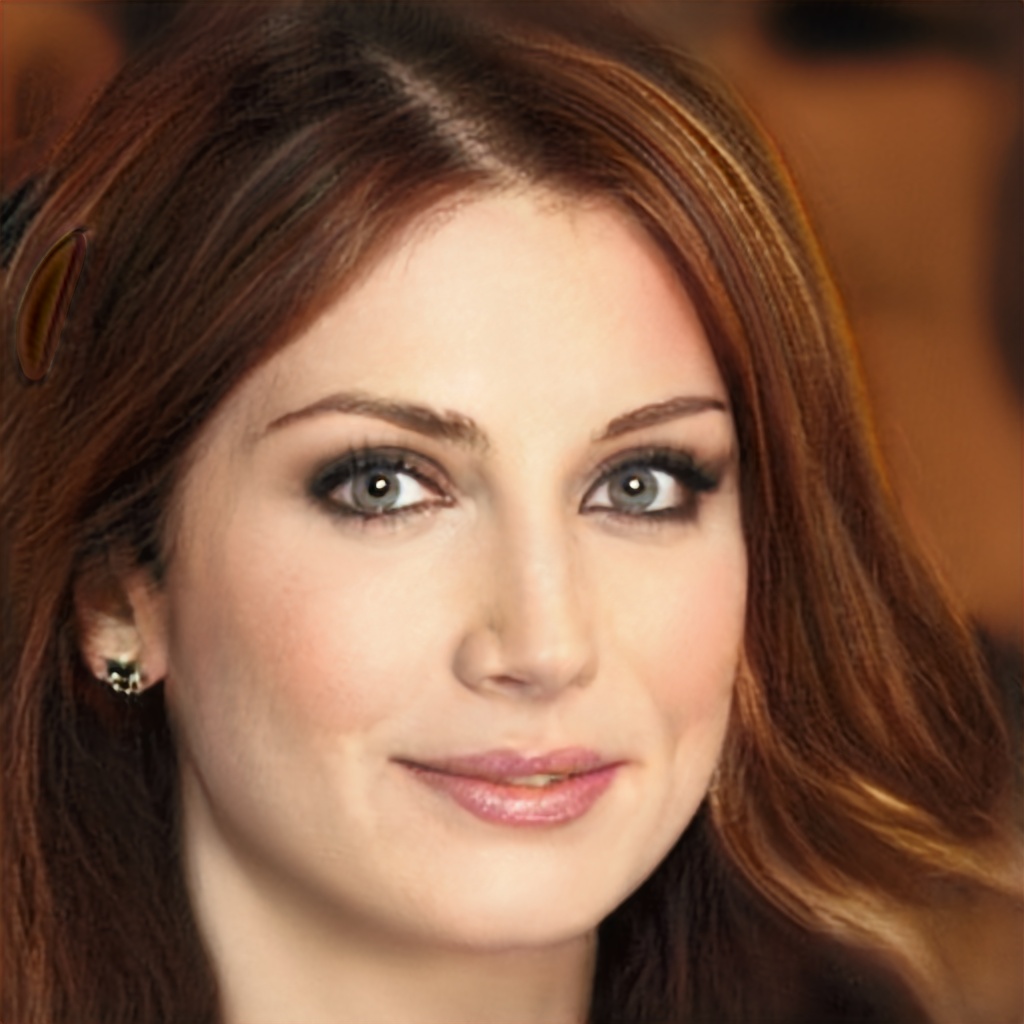} & \includegraphics[width=0.15\linewidth]{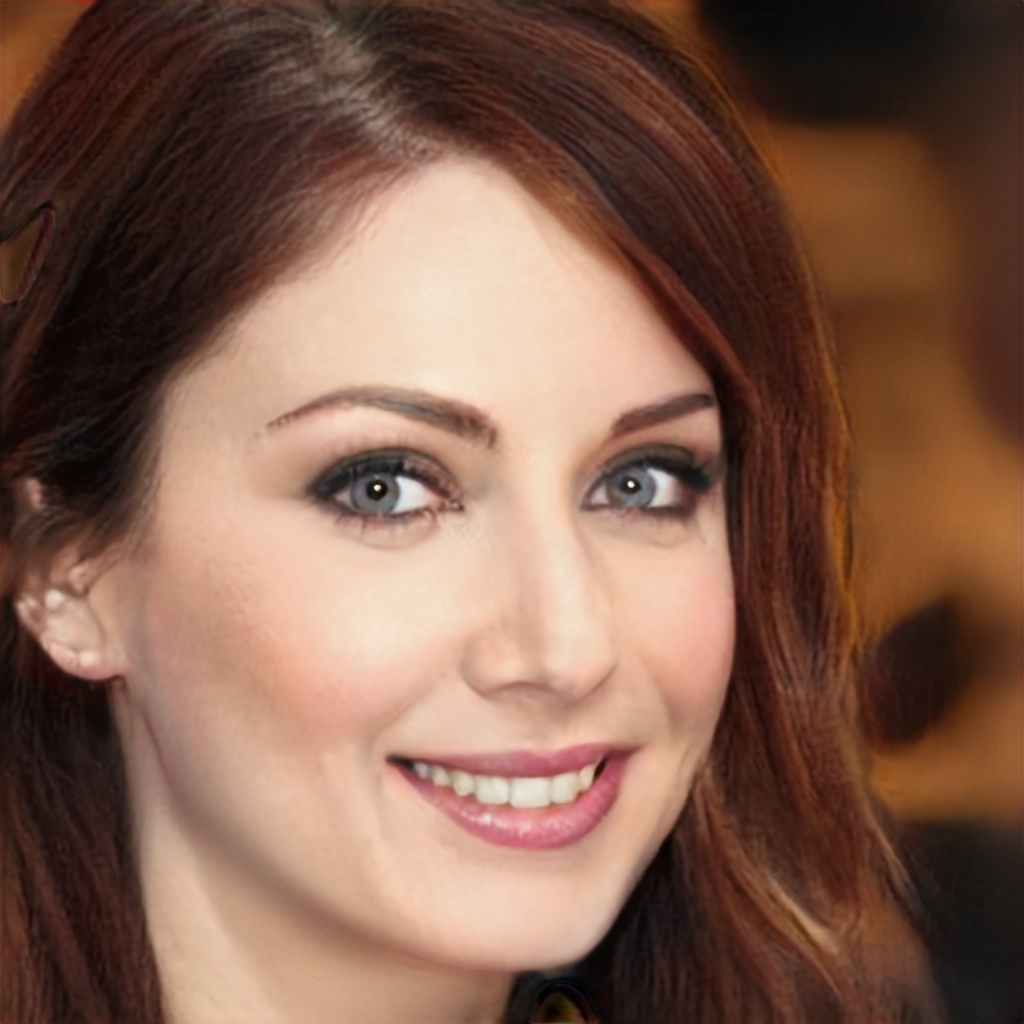} & \includegraphics[width=0.15\linewidth]{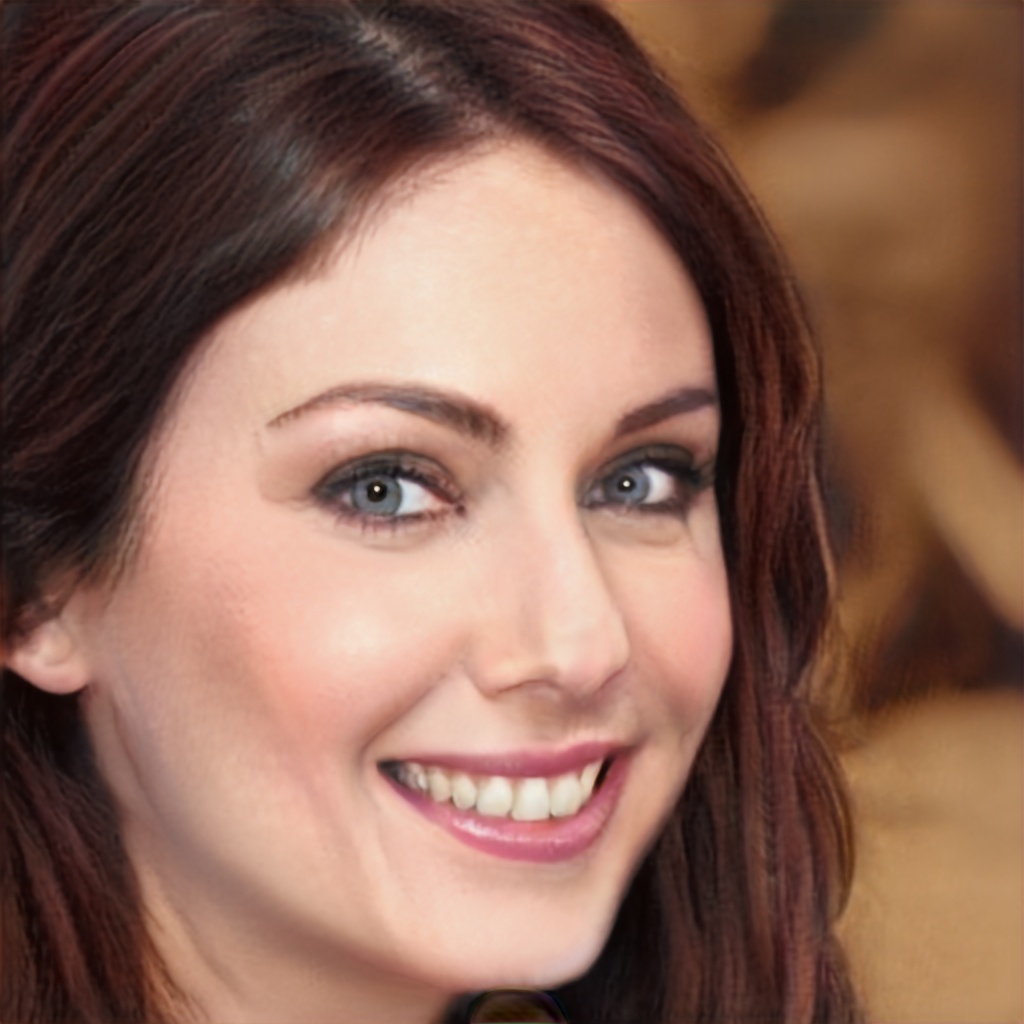} \\
 \multicolumn{3}{c}{Nose up/down} & & \multicolumn{3}{c}{Nose left/right}\\
 \includegraphics[width=0.15\linewidth]{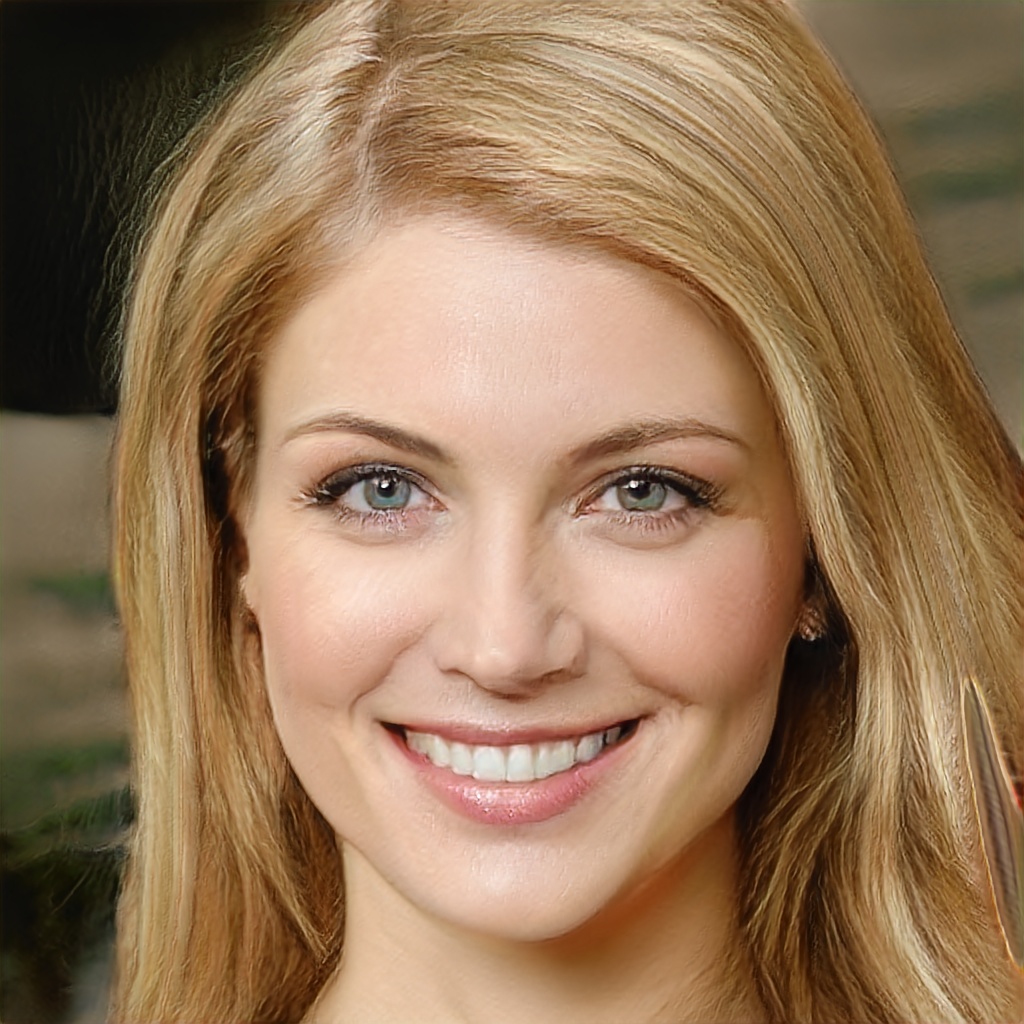} & \includegraphics[width=0.15\linewidth]{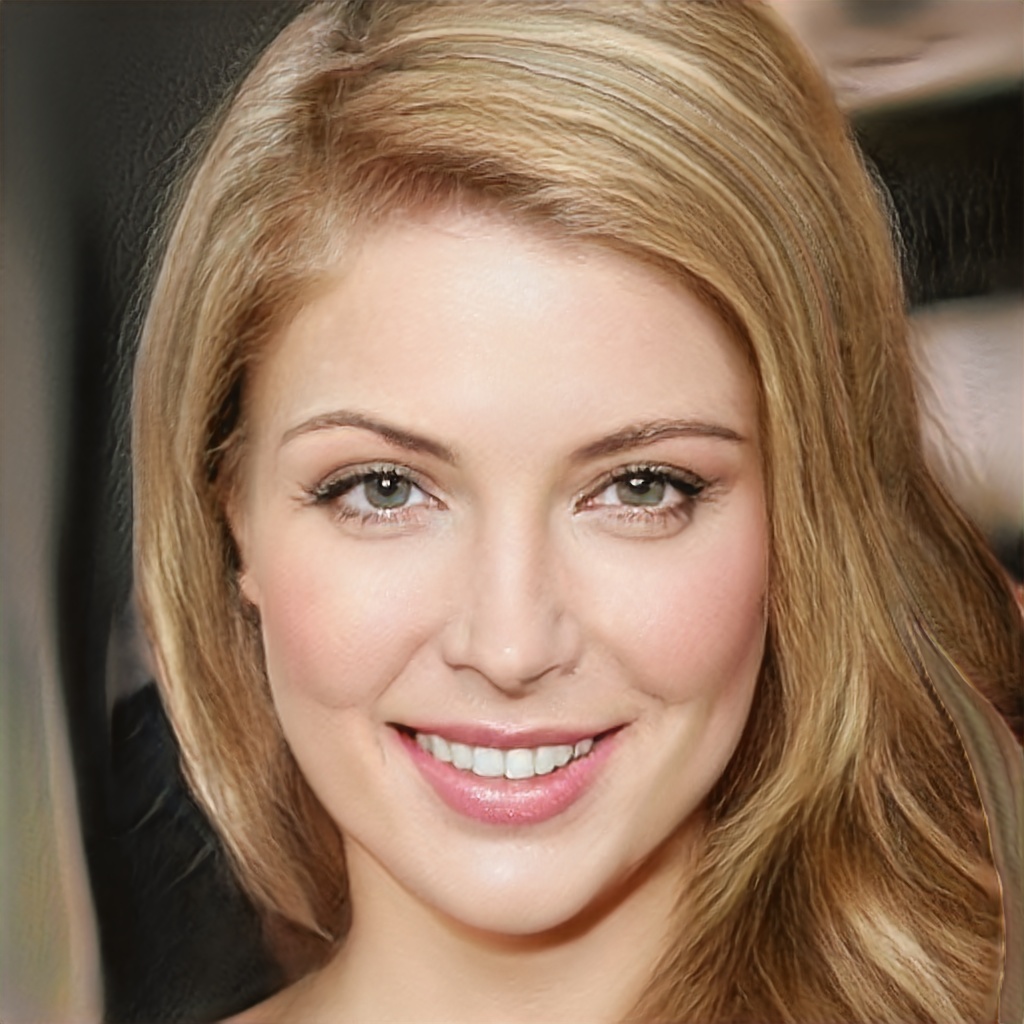} & \includegraphics[width=0.15\linewidth]{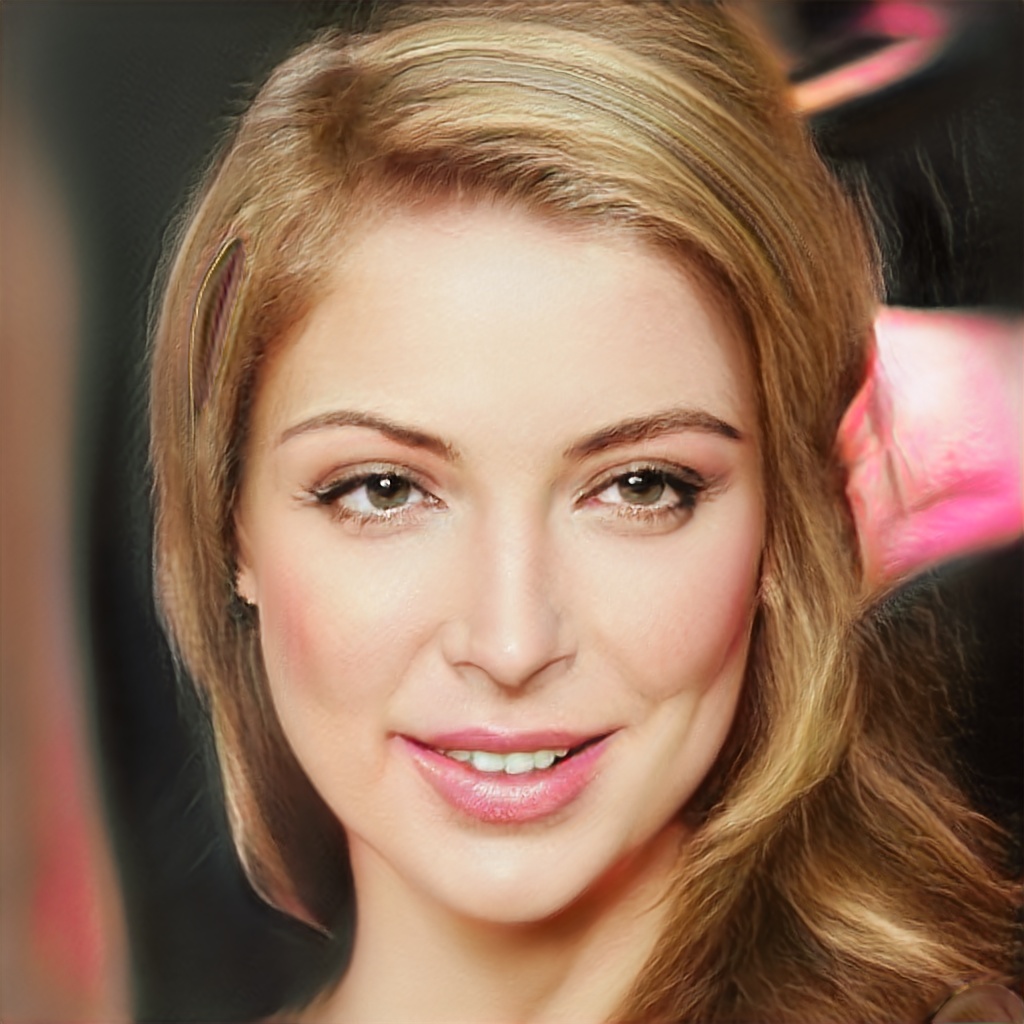} &  & \includegraphics[width=0.15\linewidth]{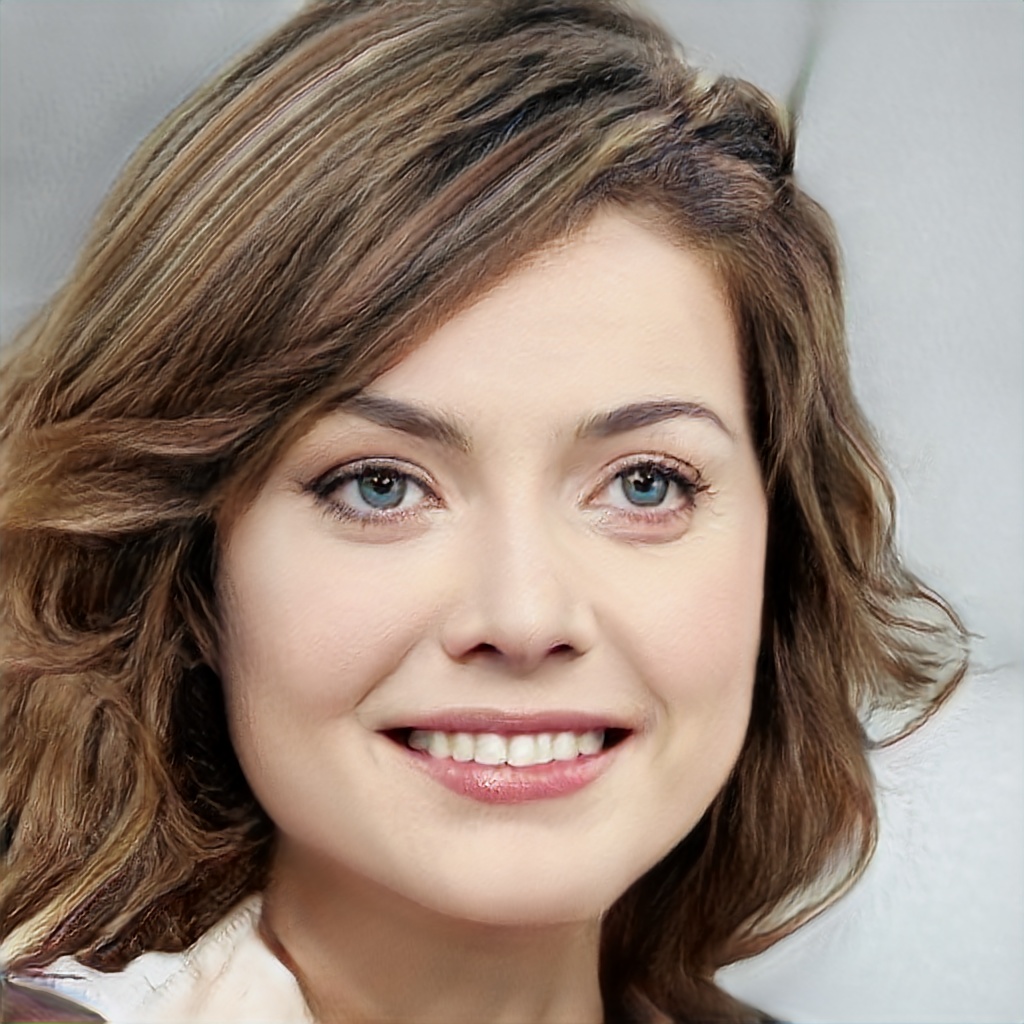} & \includegraphics[width=0.15\linewidth]{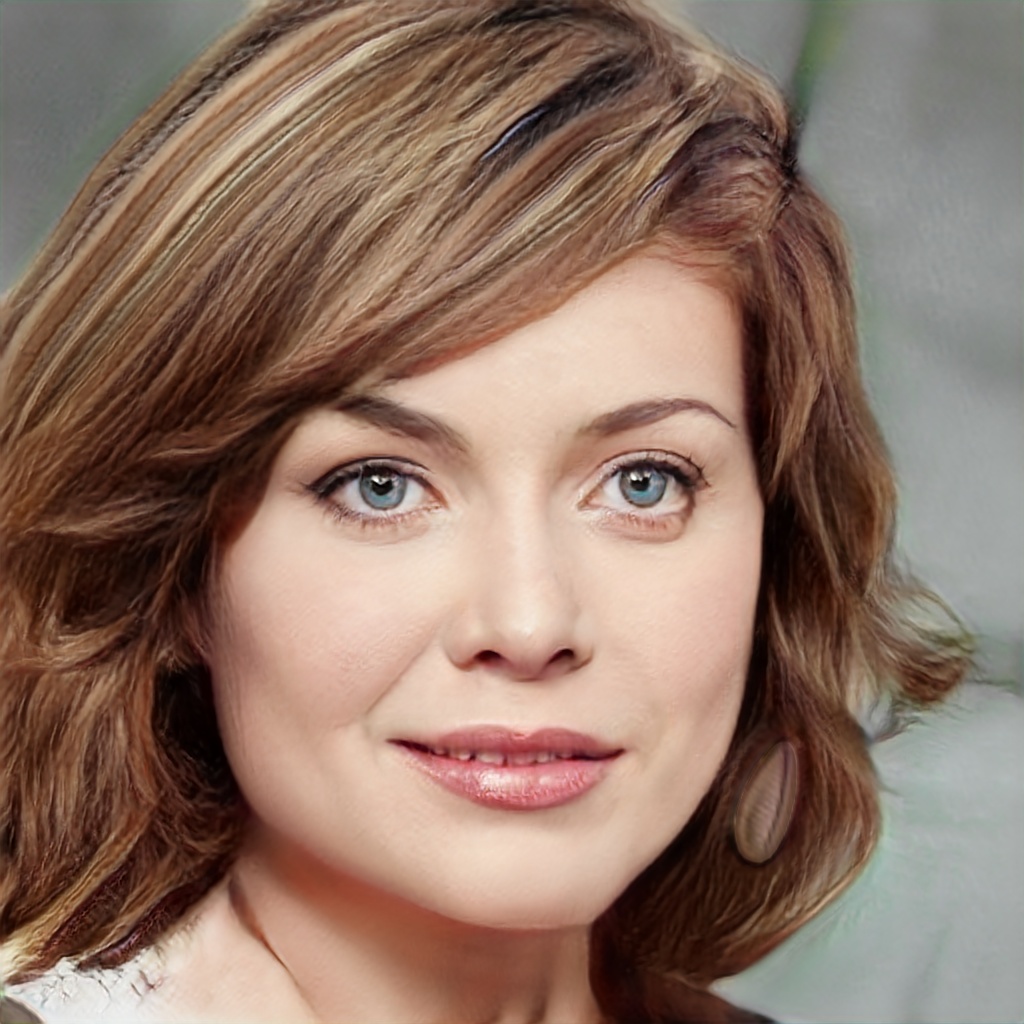} & \includegraphics[width=0.15\linewidth]{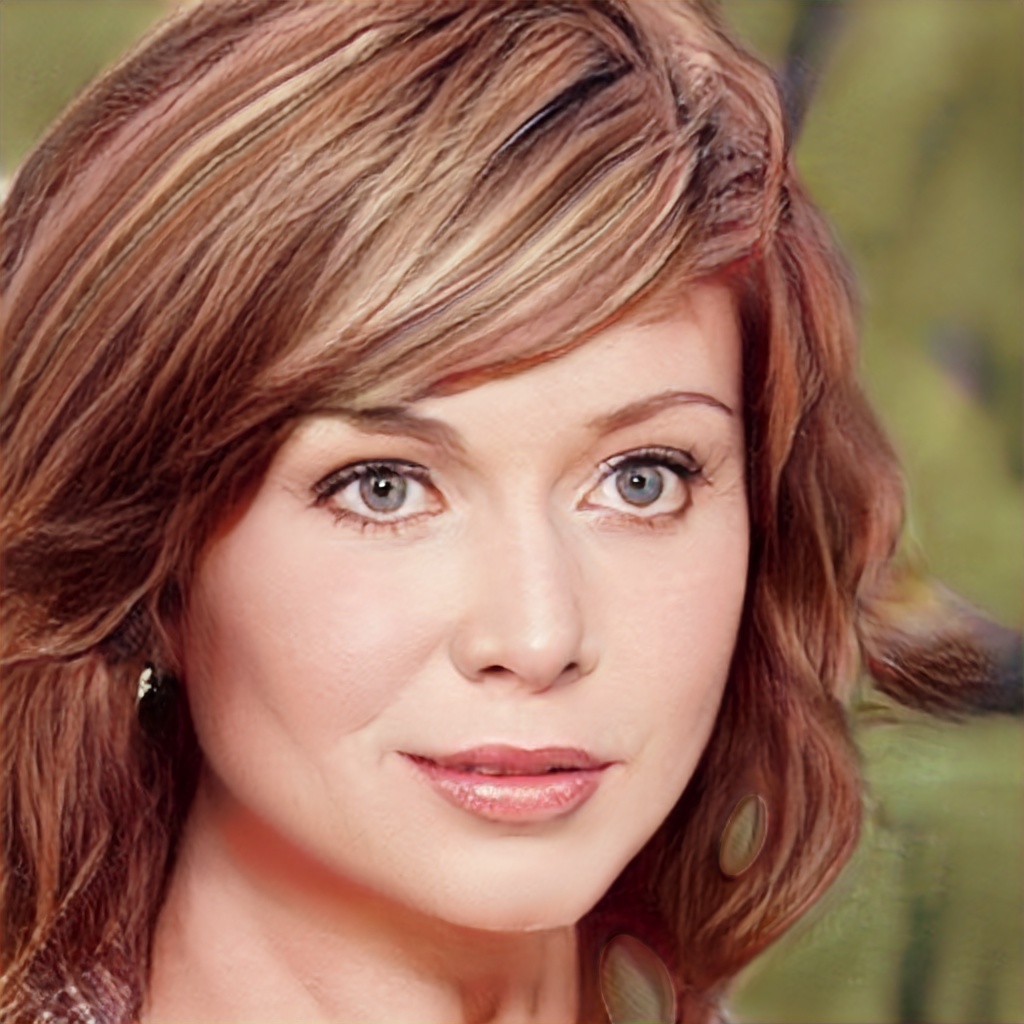}\\
 \multicolumn{3}{c}{Left corner of mouth} & & \multicolumn{3}{c}{Right conrner of mouth}
\end{tabular}
\caption{Landmark editting on StyleGAN with respect to Nose up/down, Nose left/right, Left corner of mouth, Right corner of mouth.}
\label{fig:landmark}
\end{figure*}
\vspace{0.1cm}\noindent\textbf{Function Approximation.} We show that our iterative scheme can better traverse the highly non-linear latent space from the perspective of function approximation. We first recall the first order Taylor approximation,
\begin{equation}
    \label{eq:taylor}
    f(x) = f(a) + f^\prime(a)(x-a).
\end{equation}
If a method models the underlying manifold better, it is expected to have lower error between the real logit $f(z^{(i+1)})$ and the estimated logit $f(z^{(i)})+f^\prime(z^{(i)})(z^{(i+1)} - z^{(i)})$. Note that the real gradient $f^\prime$ in Eq.~\ref{eq:taylor} is unavailable for all baselines and our model. Instead, all methods use estimated gradients, namely $\mathbf{J}_j$ (Eq.~\ref{eq:iterative_traversal}) for our method and $v$ (see Section~\ref{ssec:exp_setup}) for InterfaceGAN. 

Since InterfaceGAN uses normal vectors of SVM decision boundaries as guiding signals, we expect it to work well in a local region while becoming less accurate in the farther distance. To verify this assumption, we measure the errors and categorize them according to the L2 distance between the estimated and initial points. Table~\ref{tab:taylor} compares our method to InterfaceGAN. It shows that our method approximates the underlying function more accurately on average. As for the smiling, since we discard the data with lower confidence (see Section~\ref{ssec:implementation}), it may make the proxy model less robust at the region where many ambiguous samples happen. Lastly, due to the iterative scheme, our method achieves lower errors even when the estimated points moves far away from the initial points. 

\subsection{Head Pose Manipulation}
\label{ssec:head_pose}
To further study the possibility of unintended usage of pretrained GANs, we employ our framework to achieve head pose manipulation by leveraging pre-trained HopeNet~\cite{ruiz2018fine} as the task model. In particular, we attempt to gain control over the yaw and pitch axes since the training dataset of the pre-trained GANs, CelebAHQ, is well-aligned by eyes; thus it is almost impossible to control the roll axis. It also suggests that the framework could be still limited by the training dataset of pretrained GANs even though we are capable of gaining control over the generation process. To generate fine-grained transition, we choose small step size 0.01 and produces 1000 steps in total.

Due to the proposed constraint in Eq.~\ref{eq:constraint}, our framework can solely edit one axis by conditioning on four attributes and the other axis. As shown in Figure~\ref{fig:pose}, we smoothly interpolate the images for yaw and pitch axes. We also note that some artifacts or undesired changes happen when forcing the pose to go beyond certain degrees. With our framework, we can generate faces with arbitrary poses if they remain feasible in the training data of pre-trained GANs.

\subsection{Landmark Editing}
By utilizing pre-trained MTCNN~\cite{zhang2016joint}, we take one step further to consider landmark editing, which is a relatively challenging application. The reasons are three-fold. First, a landmark is two coordinates, which are extremely local attributes as compared to, e.g., age or gender. Second, since landmarks are coordinates, it becomes very sensitive for annotation. The performance may be degraded if the accuracy of task models are not satisfactory. Lastly, landmarks are often correlated with other features. Based on the results of previous experiments, we edit landmarks while conditioning on 4 attributes, poses, and a subset of landmarks. Note that since the training dataset is well aligned, the landmark of eyes are not editable.  

Figure~\ref{fig:landmark} demonstrates the results when editing the coordinates of noses and mouths. Surprisingly, we find some attributes are highly entangled. For instance, \textit{mouth landmarks} is highly correlated with \textit{smile}. To measure the correlation, we additionally compute the cosine similarity between direction vectors (Eq.~\ref{eq:iterative_traversal}). Higher similarity indicates higher entanglement. We analyze the correlation of \textit{mouth landmarks vs. smile} (0.296) and \textit{mouth landmarks vs. gender} (0.035). The former one is around 8.5 times larger than the latter. Additionally, \textit{nose landmarks vs. yaw} are highly correlated by 0.5014. In this analysis, we do find measurable entanglement.
 
\raggedbottom
\section{Conclusion}
\label{sec:conclusion}
In this paper, we demonstrate that publicly released state-of-the-art GANs can be applied for a wide range of applications beyond the creators' intention. To achieve this, we propose a framework to gain high-level control over unconditional image generation. Extensive results show that the framework is advantageous over non-iterative linear baselines in many aspects and can be readily applied to various unintended tasks. While this allows to re-use high quality GANs that are becoming increasingly costly to train for new applications, we hope that this work also raises awareness regarding potential unintended usage, urging the creators and owners to be cautious about the possible risks of their models before a full release.
\section*{Acknowledgment}
This work is partially funded by the Helmholtz Association within the project "Protecting Genetic Data with Synthetic Cohorts from Deep Generative Models (PRO-GENE-GEN)" (ZT-I-PF-5-23).

\FloatBarrier

{\small
\bibliographystyle{ieee_fullname}
\bibliography{egbib}
}

\clearpage
\newpage
\appendix
\setcounter{table}{0}
\renewcommand{\thetable}{A\arabic{table}}
\setcounter{figure}{0}
\renewcommand{\thefigure}{A\arabic{figure}}
\setcounter{equation}{0}
\renewcommand{\theequation}{A\arabic{equation}}

\SetKw{KwTo}{in}\SetKwFor{For}{for}{\string do:}{end}%

\noindent \textbf{\LARGE Appendix} \\

\section{Quality of Generated Images}
We follow the evaluation protocol of StyleGAN and report average FID scores. As shown in Table~\ref{tab:appendix_fid}, we observe that both methods degrade the image quality of StyleGAN. As compared to InterfaceGAN, our method performs slightly better in the unconditional setting and significantly better in the conditional setting. The experiment confirms the effectiveness of our non-linear method on the quality of generated images. Note that following InterfaceGAN, we do not normalize latent codes and find the vanilla setting slightly worse than the one reported by StyleGAN (5.04).

\begin{table}[h]
\centering
\begin{tabular}{@{}lcc@{}}
\toprule
 & Unconditional & Conditional \\ \cmidrule(l){2-3} 
Interfacegan & 26.70 &  21.46\\
Ours & \textbf{23.79} &  \textbf{14.90} \\ \midrule
Vanilla GAN & \multicolumn{2}{c}{11.14 (5.04)} \\ \bottomrule
\end{tabular}
\caption{FID evaluation on StyleGAN. Lower is better.}
\label{tab:appendix_fid}
\end{table}

\section{Detailed Iterative Algorithm}
We detail the proposed iterative framework in Algorithm~\ref{alg:overall}. We first recall our proposed framework. The Jacobian matrix of the proxy model is computed as follows,
\begin{equation}
    \label{eq:appendix_jacobian}
    \renewcommand\arraystretch{2}
    \mathbb{J} = \begin{bmatrix}
        \frac{\partial \mathcal{P}_1}{\partial z_1} & \cdots & \frac{\partial \mathcal{P}_1}{\partial z_n} \\
        \vdots & \ddots & \vdots \\
        \frac{\partial \mathcal{P}_m}{\partial z_1} & \cdots & \frac{\partial \mathcal{P}_m}{\partial z_n} \\
    \end{bmatrix},
\end{equation}
where $\mathcal{P}_j$ denotes j-th attributes predicted by the proxy. We iteratively explore the latent space by dynamically discovering the most representative direction, which can be shown as below.

\begin{equation}
    \label{eq:appendix_iterative_traversal}
    z^{(i+1)} = z^{(i)} - \lambda \mathbb{J}^{(i)}_j,
\end{equation}
where $\lambda$ is a hyper-parameter deciding moving speed and $\mathbb{J}^{(i)}_j$ is associated with the attribute of interest at step i. To encourage disentanglement, we further propose an orthogonal constraint.

\begin{equation}
\label{eq:appendix_constraint}
\begin{aligned}
 & \underset{n}{\text{maximize}} && \mathbb{J}_{j}^Tn \\
& \text{subject to} && An = 0,
\end{aligned}
\end{equation}
where $n$ is the direction vector of interest, and each row of $A$ consists of the attribute vector $\mathbb{J}_{k \neq j}$ on which we want to condition. Note that we take decreasing logit values as an example for simplicity. One could achieve the opposite direction by gradient ascent.

\begin{algorithm}
\SetAlgoLined
\KwIn{Trajectory length $L$, step size $\lambda$, initial point $z^{(0)}$, target attribute index $j$, proxy model $\mathcal{P}$, condition index set $K$} 
\KwOut{Trajectory $\mathcal{T}=\{z^{(1)}, \ldots, z^{(N)}\}$}
Initialize $\mathcal{T} \leftarrow \emptyset$ \;
 \For{i \KwTo $\{0, \ldots, L-1\}$}{
  Compute the Jacobian matrix $\mathbb{J}^{(i)}$ by Eg.~\ref{eq:appendix_jacobian} \;
  \eIf{$K \neq \emptyset$}{
   Construct matrix $A$ by vectors $\mathbb{J}_{k \in K}^{(i)}$ \;
   Solve orthogonal vector $n$ by Eq.~\ref{eq:appendix_constraint} \;
   }{
   $n \leftarrow \mathbb{J}_{j}^{(i)}$}
  $z^{(i+1)} \leftarrow z^{(i)} - \lambda n$ (Eq.~\ref{eq:appendix_iterative_traversal})\;
  Add $z^{(i+1)}$ to $\mathcal{T}$ \;
 }
 \KwRet{$\mathcal{T}$} \
 \caption{HijackGAN}
 \label{alg:overall}
\end{algorithm}

\section{Visualization of Smoothness}
\begin{figure*}
    \centering
    \includegraphics[width=\linewidth]{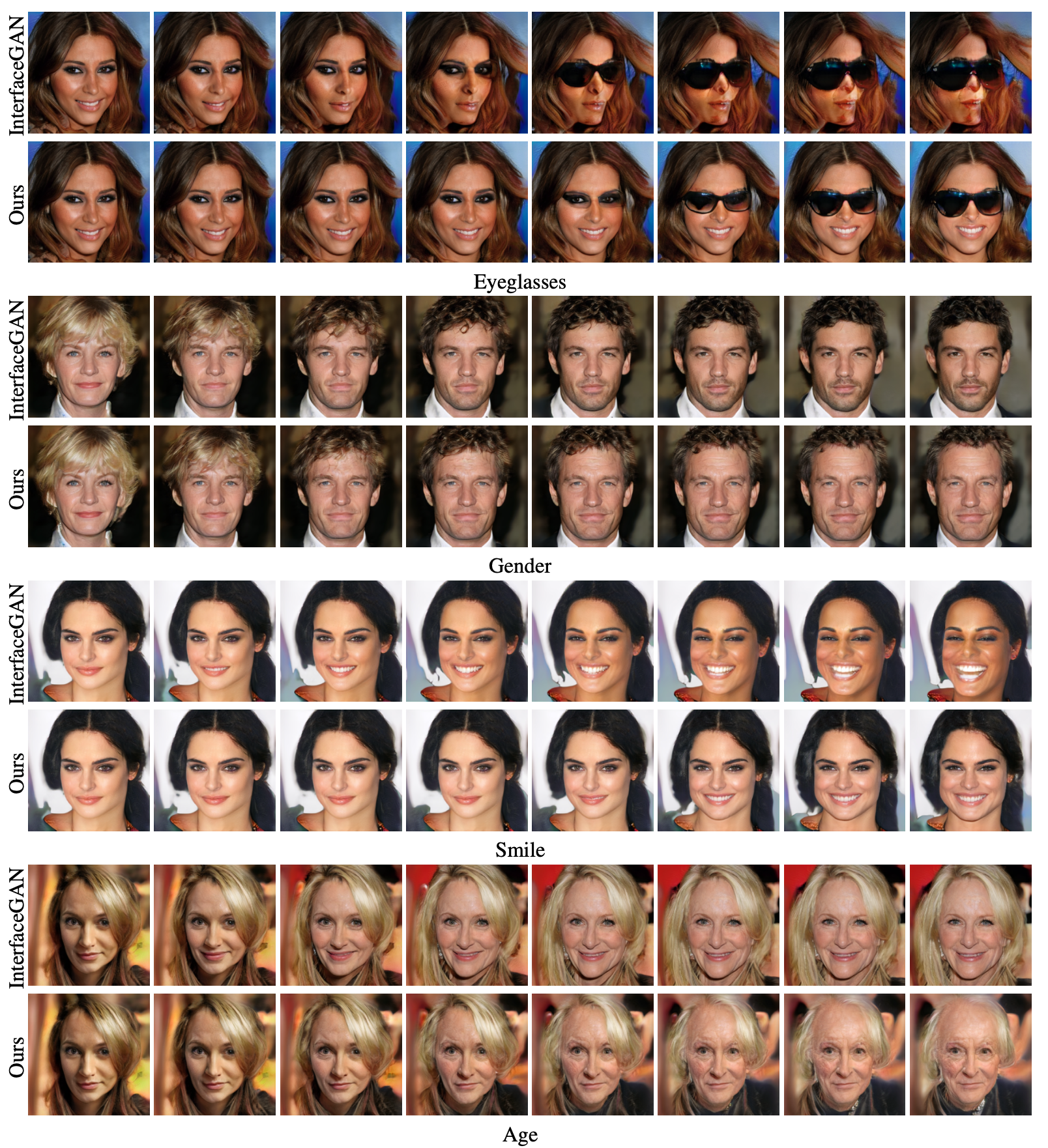}
    \caption{\textbf{(Conditional)} visualization of smoothness on PGGAN. All attributes should remain the same except the target one.}
    \label{fig:pggan_smooth}
\end{figure*}

\begin{figure*}
    \centering
    \includegraphics[width=\linewidth]{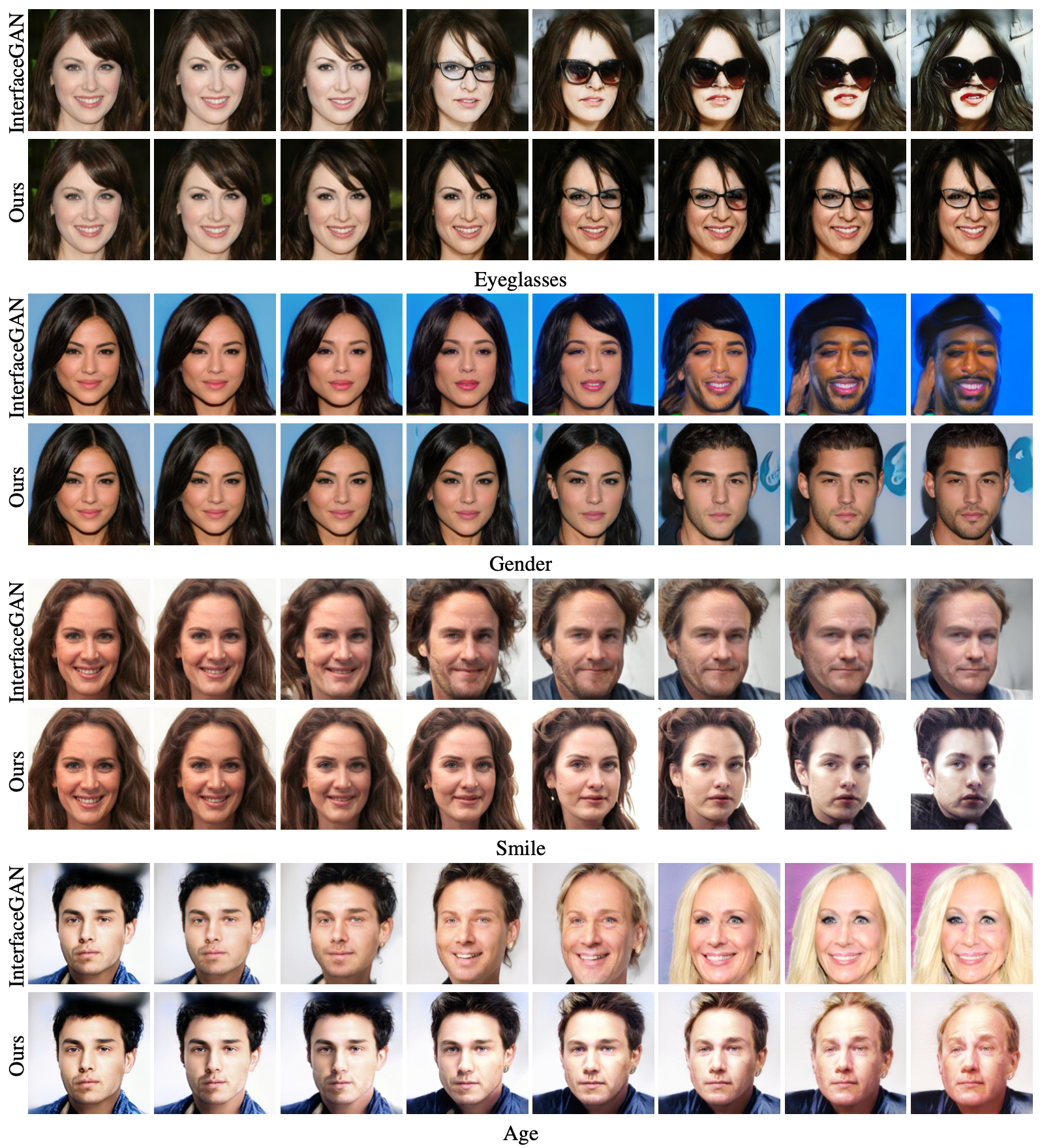}
    \caption{\textbf{(Conditional)} visualization of smoothness on StyleGAN. All attributes should remain the same except the target one.}
    \label{fig:stylegan_smooth}
\end{figure*}

Although we have verified the smoothness of our method in terms of modified Perceptual Path Length in Table 1 of the main paper, we additionally provide qualitative results on PGGAN (Figure~\ref{fig:pggan_smooth}) and StyleGAN (Figure~\ref{fig:stylegan_smooth}), respectively. All experiments are conducted in the conditional setting, meaning that we solely edit one attribute, and others should remain the same. In particular, we take 40 steps with step size 0.2 and show the images from every 5 steps. Note that, the closer to the right-hand side, the farther from the initial point.

From Figure~\ref{fig:pggan_smooth} and Figure~\ref{fig:stylegan_smooth}, we make two observation. First, our method preserves attributes better on both models. For example, our method can preserve smiling when editing eyeglasses, and preserve the smiling when editing age on both models. Second, on StyleGAN, our method can produce smoother transitions. For example, gender and eyeglasses.

\end{document}